\documentclass[10pt]{asme2ej}
\usepackage[utf8]{inputenc}
\usepackage{times}          
\usepackage{helvet} 
\usepackage{courier} 
\usepackage{graphicx, wrapfig, subfigure, gensymb} 
\usepackage{multicol} 
\usepackage[bottom]{footmisc}
\usepackage{amsmath,amssymb,epsfig}
\usepackage{epstopdf}
\usepackage{color}
\usepackage{xspace}
\usepackage{url}
\usepackage{changepage}
\usepackage[toc,page]{appendix}
\usepackage{float}
\usepackage[utf8]{inputenc}
\usepackage[normalem]{ulem}
\usepackage[T1]{fontenc}
\usepackage{placeins}
\usepackage{afterpage}
\usepackage{bigfoot}
\usepackage{lineno}
\usepackage{hyperref}
\usepackage{mathtools}
\usepackage{booktabs} 
\usepackage{multirow}
\usepackage{array}
\usepackage{framed}
\usepackage{nomencl}
\usepackage{caption}
\usepackage{breqn}
\usepackage{longtable}
\usepackage{rotating}
\usepackage{cases}
\newcommand{\vmc}[1]{{\color{black} #1}}

\usepackage{xspace}
\usepackage{amsmath,amssymb,amsfonts}

\graphicspath{{./Figures/}}

\begin{document}

\title{A review of cuspidal serial and parallel manipulators}

\author{
Philippe Wenger and Damien Chablat \\
Nantes Université, École Centrale Nantes, CNRS, LS2N, UMR 6004, F-44000 Nantes, France
}
\maketitle

	%
\begin{abstract}
\emph{Cuspidal} robots can \vmc{move from one inverse or direct kinematic solution to another} without ever passing through a singularity. These robots have remained unknown because almost all industrial robots do not have this feature. However, in fact, industrial robots are the exceptions. Some robots appeared recently in the industrial market can be shown to be cuspidal but, surprisingly, almost nobody knows it and robot users meet difficulties in planning trajectories with these robots. This paper proposes a review on the fundamental and application aspects of cuspidal robots. It addresses the important issues raised by these robots for the design and planning of trajectories. The identification of all cuspidal robots is still an open issue. This paper recalls in details the case of serial robots with three joints but it also addresses robots with more complex architectures such as 6-revolute-jointed robot and parallel robots. We hope that this paper will help disseminate more widely knowledge on cuspidal robots.  
\end{abstract}


\section{Introduction}\label{sc:intro}
    When a new robot is to be implemented in a production site, its kinematic architecture must be chosen initially (serial or parallel, choice of the number and types of axes, etc.). Then, the robot must be programmed and controlled so that the tool it is handling can properly follow the trajectories defined to carry out the required tasks. Most serial robots have the ability to reach a pose in their workspace with several \vmc{inverse kinematic solutions}, e.g., "elbow up" and "elbow down". A change of \vmc{solution} can be made to avoid a joint limit or a collision. It has long been assumed that any robot must necessarily pass through a singular configuration during a change of \vmc{solution}, as is the case for most industrial robots: the singular configuration "extended arm" must always be passed through to move between "elbow up" and "elbow down". This property is indeed observed on robots that have geometrical simplifications, such as, parallel or intersecting axes, which is the case of most industrial robots. Yet, these robots are exceptions. A cuspidal robot is defined as a robot that can \vmc{move from one solution to another} without ever passing through a singularity. 
    The very first mention of a non-singular change of \vmc{solution} was made in 1988 by Innocenti and Parenti-Castelli from the University of Bologna, who demonstrated this phenomenon numerically on two different serial robots with six revolute joints \cite{parenti-castelli_position_1988}. This revelation went unnoticed and was even sometimes considered fanciful. In fact, the community was convinced that any robot must necessarily cross a singularity to change its \vmc{solution}, as was the case for all industrial robots at that time. A mathematical proof - which later proved to be false - had even been produced to conﬁrm this hypothesis \cite{borrel1986study}. Shortly after the Bologna researchers' revelation, Burdick (Stanford University, California) confirmed the Italians' revelations in his Ph.D thesis, this time on several serial robots with 3 revolute joints. It was not until 1992 that a new formalism was proposed for serial robots capable of changing \vmc{solution} without passing through a singularity \cite{wenger1992new}, pointing out an error in the proof produced in \cite{borrel1986study}. It took many years before the scientific community began to accept the existence of the so-called \emph{cuspidal} robots. The term \emph{cuspidal} was coined in 1995 with the demonstration of the existence of a "cusp" point (see Fig.~\ref{fig:FoldCusp}) on the locus of the singularities of cuspidal serial robots with three revolute joints \cite{el1995recognize}. Finally, an exhaustive classification of 3R serial robots with mutually orthogonal joint axes was carried out in 2004 \cite{baili2004classification}.
        Studies on cuspidal parallel robots were initiated in 1998 by Innocenti and Parenti-Castelli \cite{innocenti1998singularity} and have been limited to planar robots \cite{WorkspaceAM98}, \cite{zein2008non}, \cite{Macho2008TransitionsBM},\cite{Bamberger}, \cite{Macho2008TransitionsBM}, \cite{5272470}, \cite{DallaLibera2014NonsingularTB}, \cite{Macho}, \cite{Manfred}, \cite{chablat2011uniqueness}, \cite{Peidro}, with the exception of a 6-degree-of-freedom decoupled parallel robot \cite{caro2012non}.
    
    If cuspidality might appear interesting at first sight, it may, in fact, produce more difficulties than advantages. This raises several crucial questions, both for the user and the designer. How do we know if a newly designed robot is cuspidal? If yes, how do we know if the robot has changed its \vmc{solution} during motion? Which robots are cuspidal, or which robots are noncuspidal? Any robot user should know if its robot is cuspidal or not and, more importantly, any robot manufacturer should know how to select suitable geometric parameters to design a cuspidal or a noncuspidal robot. Unfortunately, this topic has not attracted a lot of researchers. Several decades after the first published work showing cuspidal robots \cite{parenti-castelli_position_1988}, it turns out that still almost nobody is aware of this feature, as confirmed in a recent report \cite{Achille}. However, we do think that this topic is very important and the available knowledge should be disseminated more widely to the robotics community and even taught in robotic courses. This has motivated us to write this review paper which, for the first time, collects all available results on cuspidal robots, both of the serial and parallel type. 
\\This paper proposes a methodology to identify and classify cuspidal robots and describes in depth their \vmc{solution} changing mechanism. Particular attention is given to a family of robots with three mutually orthogonal revolute joint axes. These robots, which can be interesting alternatives to the usual robots, can be classified into cuspidal and noncuspidal robots on the basis of the values of their geometric parameters. Robots with six revolute joints are more difficult to analyze and are still under research at the time of writing this paper. They are discussed through some examples of cuspidal robots, some of which are used in the industry, such as painting or collaborative robots. The case of parallel robots, which is more delicate, is also investigated. Examples of parallel planar, spherical and spatial 6-degree-of-freedom robots are analyzed, capable of changing assembly mode without crossing a singularity. 

\section{Preliminaries}

We recalled in this section some basic definitions and notions that are useful for the understanding of cuspidal robots.

\subsection{Types of robots studied}\label{sec:TypesRobots}
\begin{figure}
\centering
\includegraphics[width=0.45\textwidth]{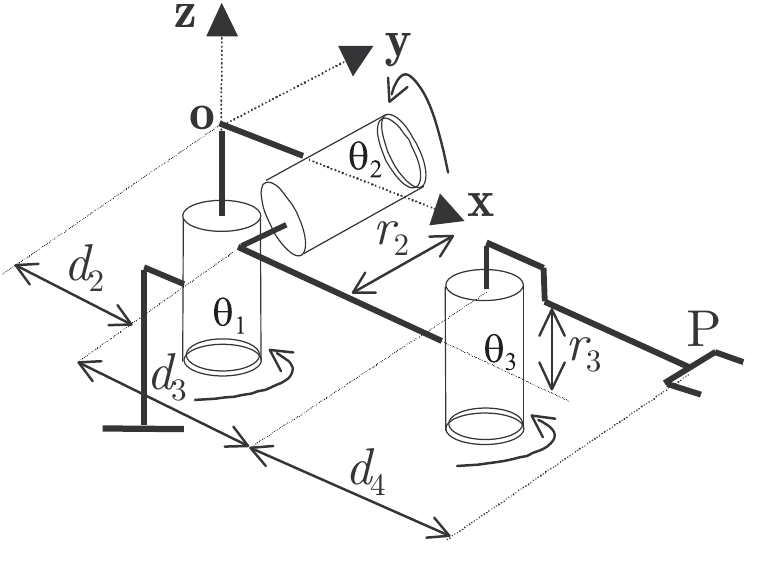}
\includegraphics[width=0.35\textwidth]{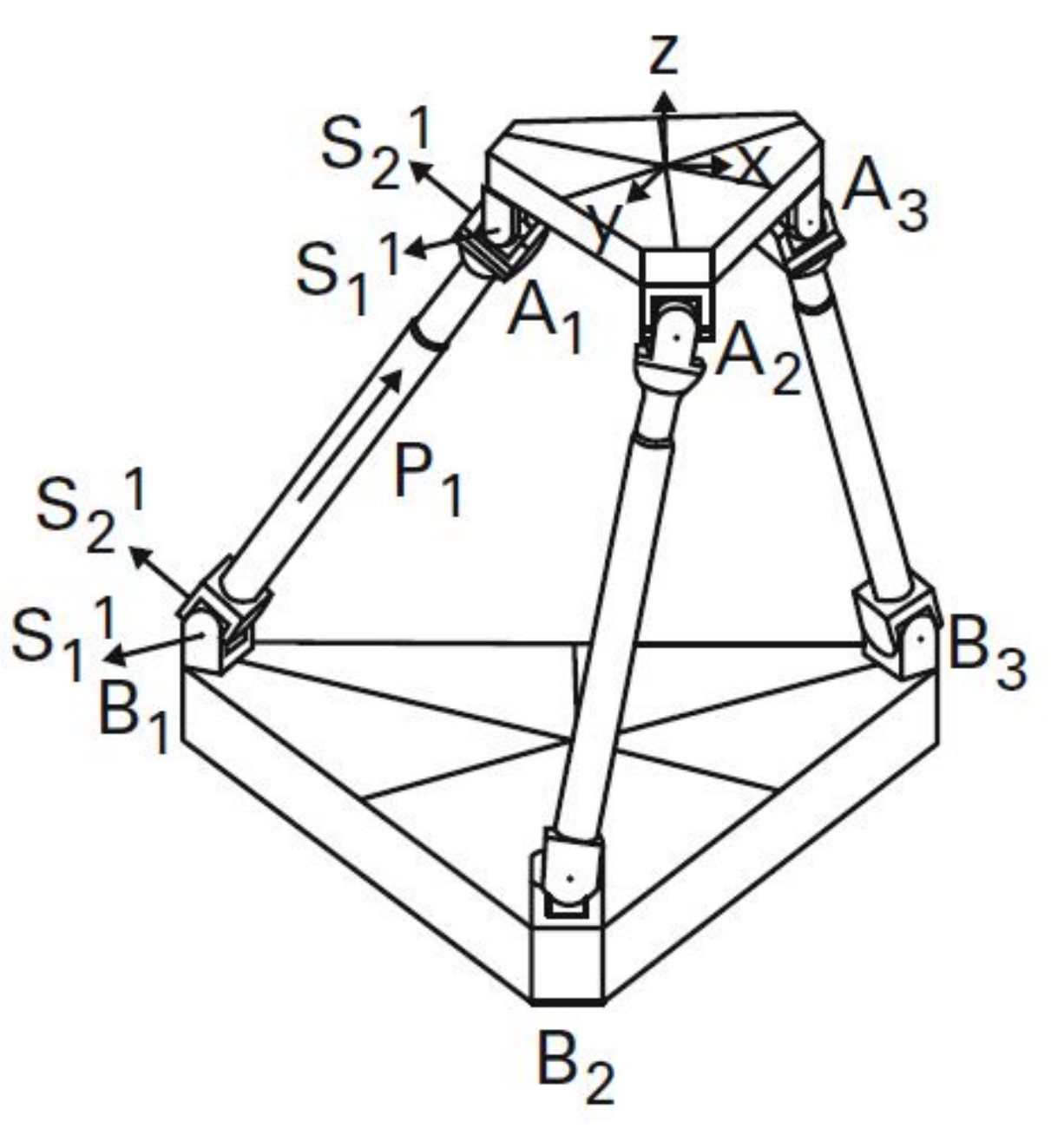}
\caption{A 3R serial robot (left), a parallel robot with 3 telescopic legs (right).}
\label{fig:SerPar}
\end{figure}

Two main categories of robots are studied in this paper: serial robots (Fig. \ref{fig:SerPar}, left) and parallel robots (Fig. \ref{fig:SerPar}, right). A large part of the paper (sections 3, 4, 5 and 6) is devoted to serial robots, and more particularly to 3R type robots, i.e. those whose articulated chain is composed of three revolute joints as shown in Fig. \ref{fig:SerPar}, left. Besides, all the robots studied in this paper are non-redundant, i.e. their number of degrees of freedom is equal to the number of coordinates describing the task. This is the case for the two 3-degree-of-freedom robots of Fig. \ref{fig:SerPar} where the task is defined by three position coordinates in space. 

\subsection{Postures and assembly-modes}\label{sec:Post}
For a serial robot, the word \emph{posture} defines a joint configuration that allows the end-effector to reach a given pose in its workspace. A posture is thus associated with an inverse kinematic solution.  

\begin{figure}
  \begin{minipage}[b]{0.24\linewidth}
   \centering
   \includegraphics[width=0.9\textwidth]{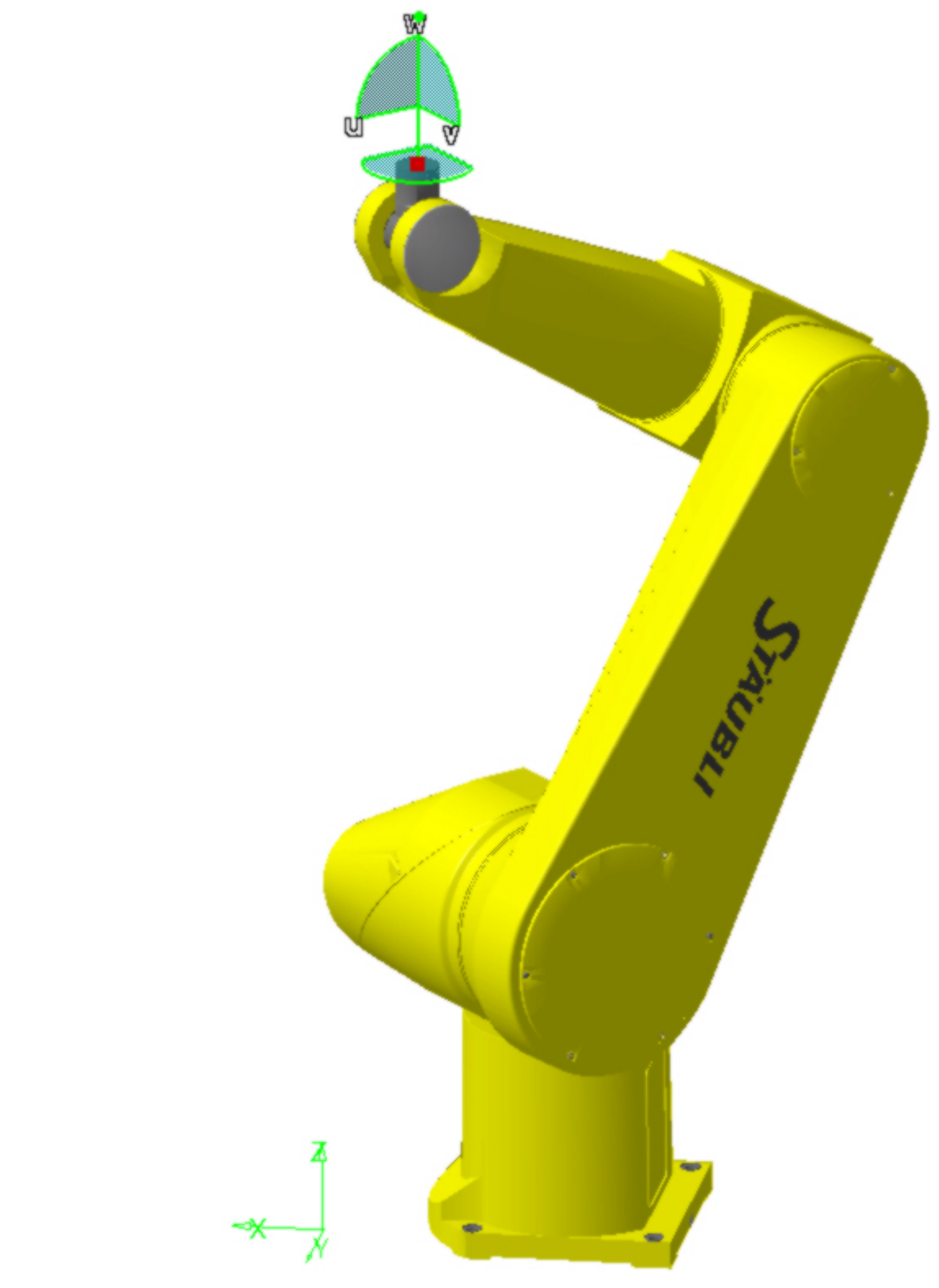}  
  \end{minipage}
  \begin{minipage}[b]{0.24\linewidth}
   \centering
   \includegraphics[width=0.9\textwidth]{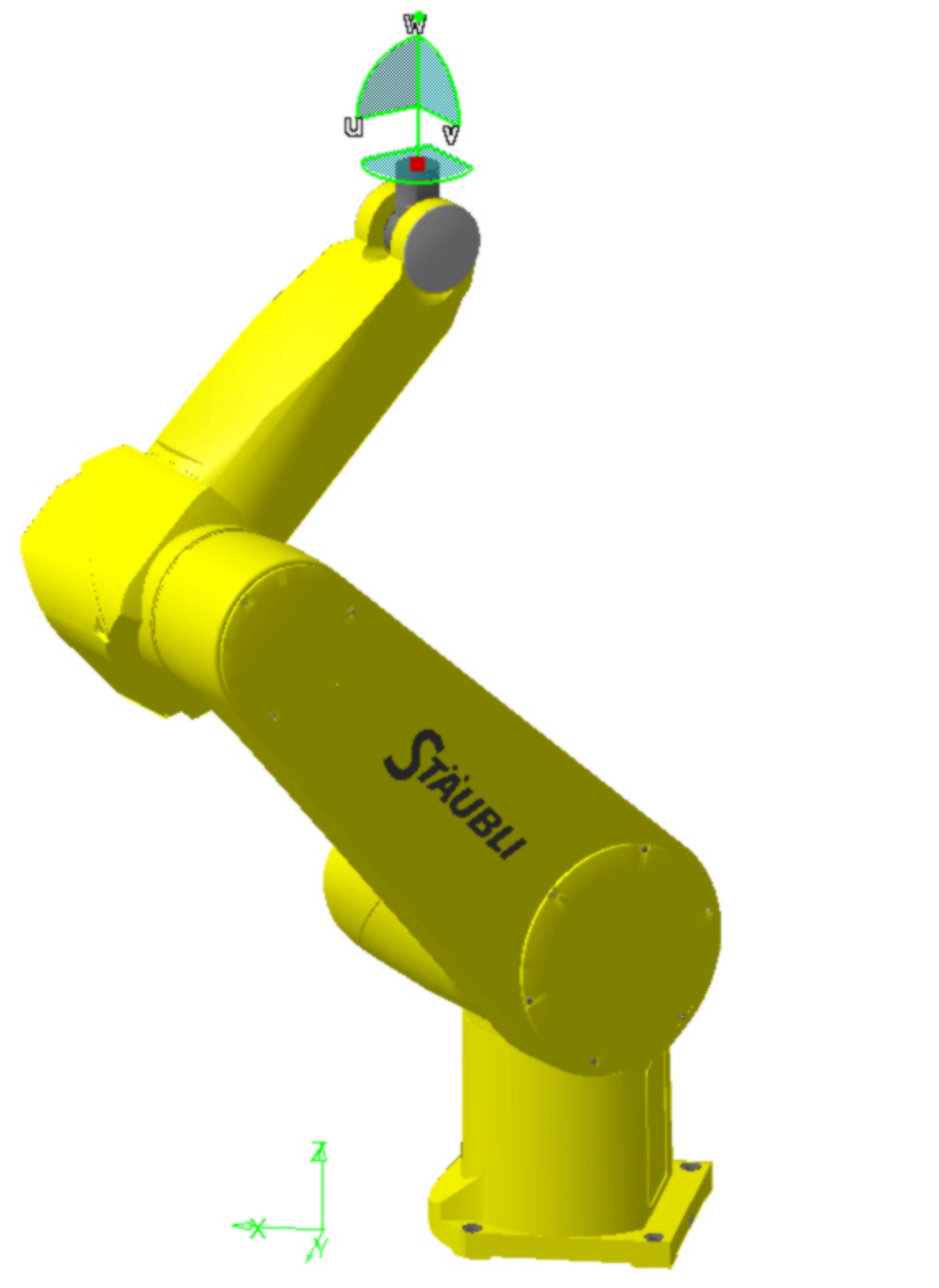}  
  \end{minipage}
  \begin{minipage}[b]{0.24\linewidth}
   \centering
   \includegraphics[width=0.9\textwidth]{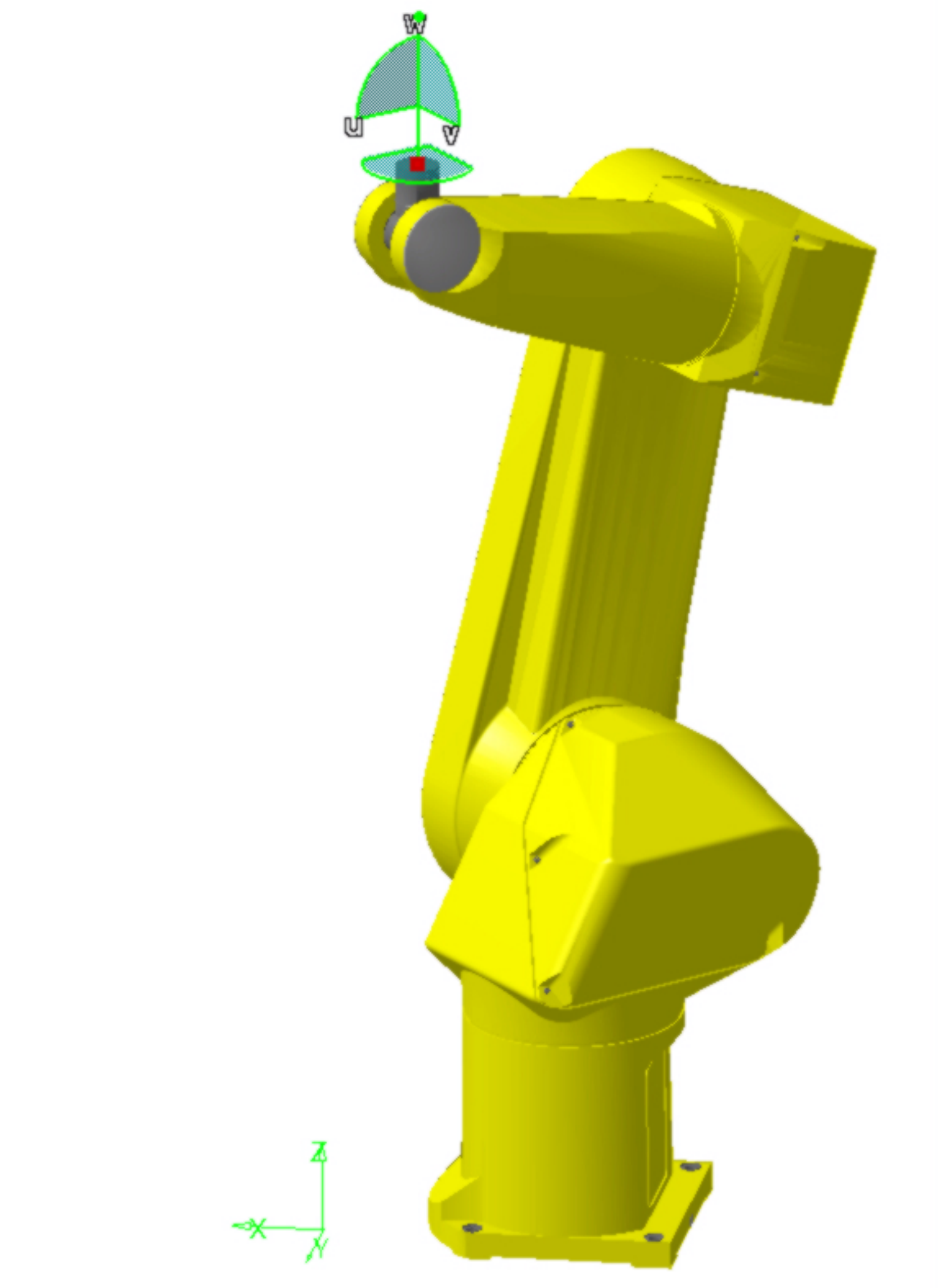}  
  \end{minipage}
  \begin{minipage}[b]{0.24\linewidth}
   \centering
   \includegraphics[width=0.9\textwidth]{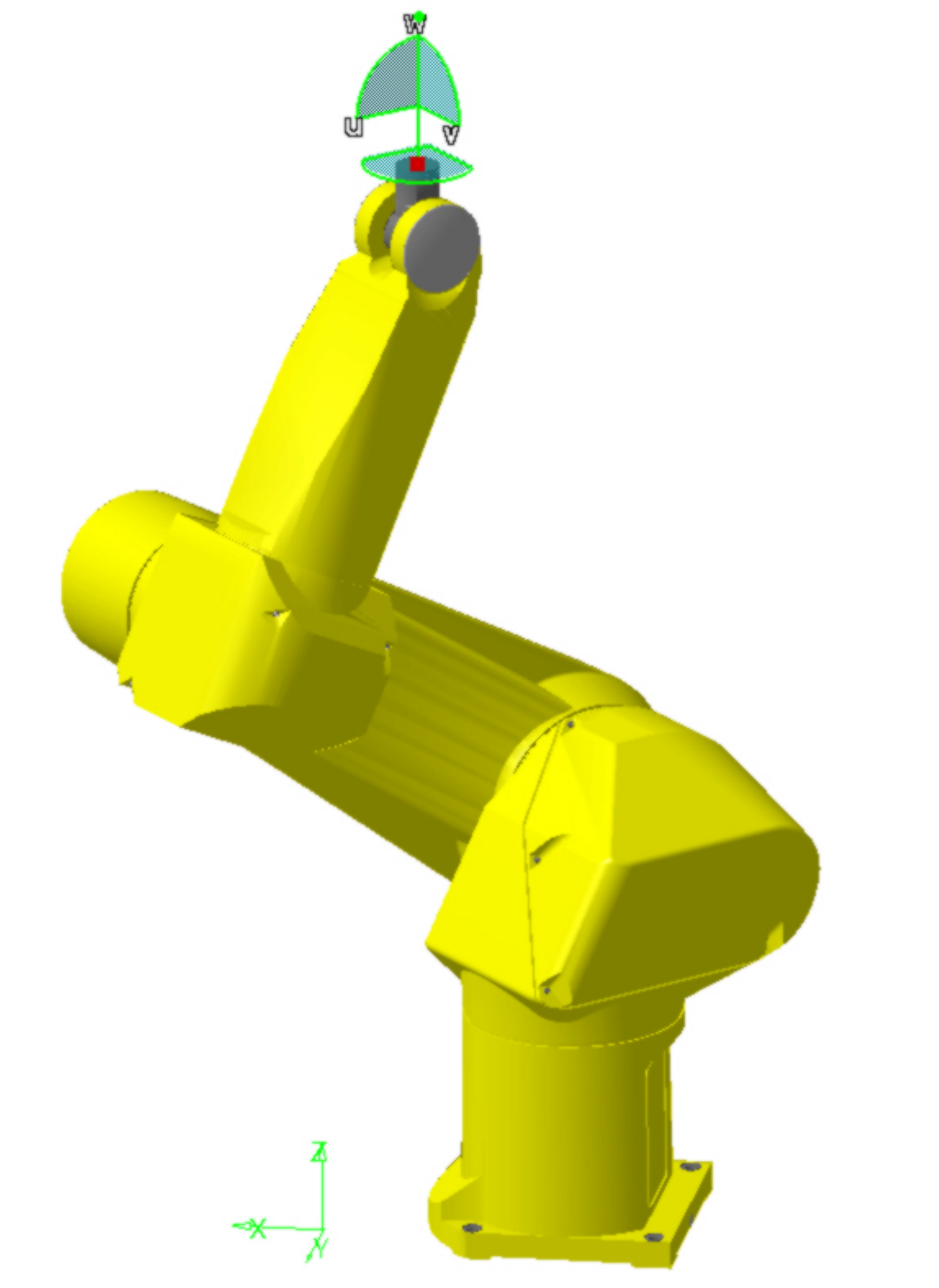}  
  \end{minipage}
\caption{The four \vmc{solutions} of an anthropomorphic robot defined by the elbow and shoulder configurations, while the wrist configuration is fixed to one of its two configuration. The remaining four postures are defined by the other wrist configuration.} 
\label{fig:PosturesNonCusp}
\end{figure}

For common industrial robots, a posture can be easily identiﬁed. For an anthropomorphic robot such as the one in Fig.~\ref{fig:PosturesNonCusp}, for example, these postures can be identiﬁed by the configuration of the elbow (up or down), shoulder (right or left), and wrist (flip or noflip). The total number of combinations amounts to $2^3 = 8$ distinct postures. Figure~\ref{fig:PosturesNonCusp} shows the four postures that can be identified by the elbow and shoulder configurations.

For a parallel robot, the term \emph{assembly mode} defines a solution to the direct kinematics, i.e. a pose of the moving platform of the robot associated to the given values of actuated joint variables. 

\begin{figure*}
\centering
\includegraphics[width=0.8\textwidth]{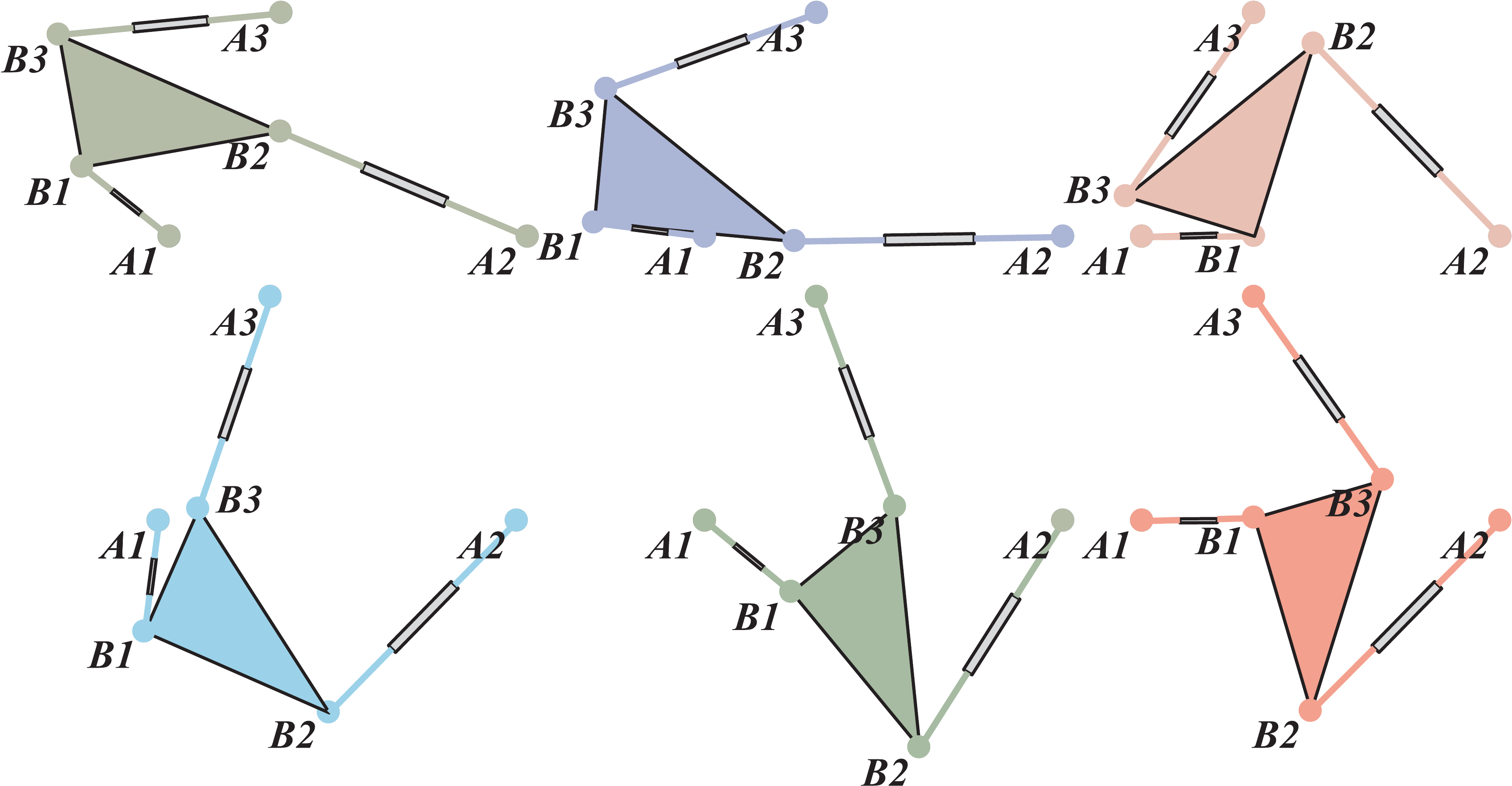}
\caption{The six assembly-modes of a planar parallel robot with three telescoping legs.}
\label{fig:ModeDassemblage}
\end{figure*}

Figure~\ref{fig:ModeDassemblage} shows the six assembly-modes of a planar parallel robot with three telescoping legs. For a given set of leg lengths $(Ai, Bi)$, the triangular platform $(B1, B2, B3)$ can assume up to six different poses in the plane. This robot will be analyzed in details in section \ref{CuspPar}.

\subsection{Singularities}\label{sec:Sing}
\begin{figure}
\centering
\includegraphics[width=0.25\linewidth]{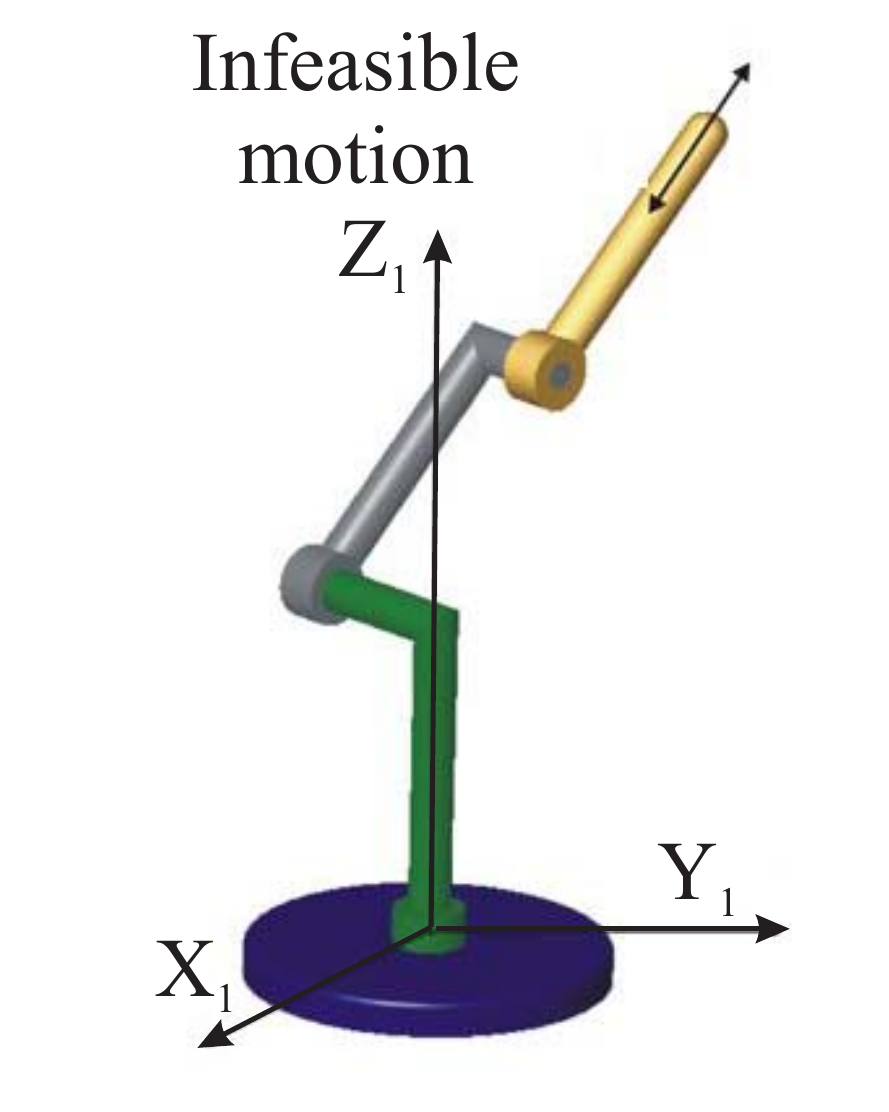}
\includegraphics[width=0.6\linewidth]{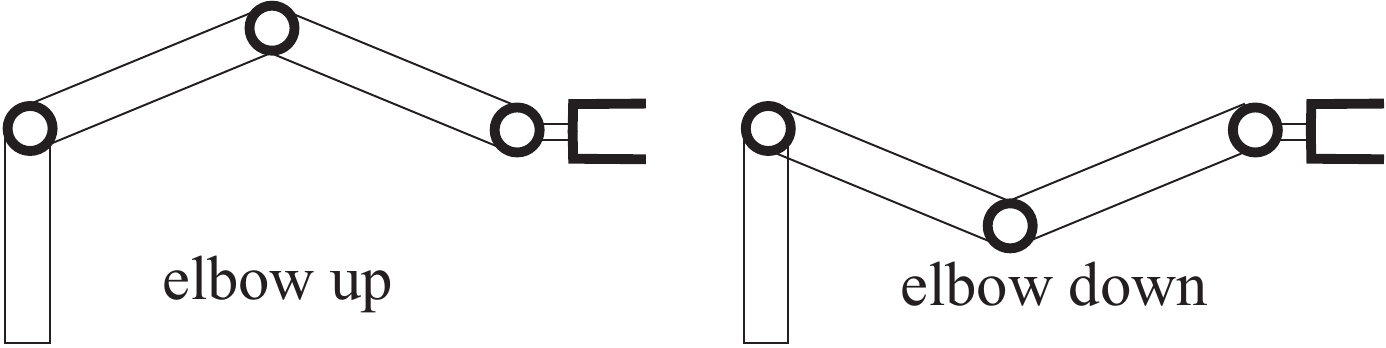}
\caption{Anthropomorphic robot in a fully outstretched arm singularity (left). The two inverse kinematic solutions ``elbow up'' and ``elbow down'' (right) coincide.}
\label{fig:SerielCoude}
\end{figure}

We briefly recall here the concept of singularity, their role being essential for cuspidal robots.
A singularity of a serial robot defines a joint configuration from which the robot cannot produce certain instantaneous motions. Moreover, at least two inverse kinematic solutions coincide when the robot is on a singularity. Figure~\ref{fig:SerielCoude}, left shows an anthropomorphic robot (sketched without its wrist) in a fully outstretched arm singularity: locally, a movement along the direction of the arm is not possible. In this singularity, the two inverse solutions associated with the solutions ``elbow up'' and ``elbow down'' shown at the right of Fig.~\ref{fig:SerielCoude}, coincide. Singularities generate boundaries in the workspace that cannot be crossed, because they are located at the edge of the workspace. The fully extended arm singularity, for example, defines the robot's reachability limit. Singularity may also define internal boundaries that can be crossed or not, depending on the robot posture. This important feature is investigated further in the paper. 

We can see that for the robot in Fig. \ref{fig:SerielCoude}, the transition between the two solutions elbow up and elbow down is necessarily done by crossing the fully outstretched arm singularity.

Parallel robots can have several types of singularities. In this paper, we will only consider parallel robots with "type 2" singularities, also called "parallel singularities". On these singularities, the parallel robot has certain motions that cannot be controlled anymore. Moreover, at least two of its assembly modes coincide.   
\subsection{Aspects}\label{sec:Aspect}
The notion of aspects was initially defined for serial robots by Borrel \cite{borrel1986study}. We recall the following definition for non-redundant serial robots.

Let $\cal D$ be the robot joint space~: 
\begin{equation}
    {\cal D} = \left\{{\bf q} | q_{i min} \leq q_i \leq q_{i max}, \forall i \in [i, n] \right\} 
\end{equation}
where ${\bf q} = [q_1, .., q_n]^T$ is the joint vector, i.e. containing the $n$ actuated joint variables.\\
Let ${\bf X} = [x_1, .., x_n]$ be a choice of output coordinates defining the pose of the end-effector. These coordinates are linked to the joint variables by the kinematic map $\bf{f}$ defined by ${\bf X}=\bf{f}(\bf{q})$, that is, $x_i=f_i(\bf{q})$, $i=1, ..,n$.
\\
The aspects are the largest \vmc{sets} in ${\cal A}_j$ in ${\cal D}$ such that~:
\begin{itemize}
\item	${\cal A}_j$ is path-connected ;
\item $ \forall \bf{q} \in {\cal A}_j$, $det({\bf J})\neq 0$
where ${\bf J} = \left[\frac{\partial f_i}{\partial q_i}({\bf q})\right]$ is the Jacobian matrix of the robot.

\end{itemize}

Aspects are bounded by singularities and joint limits when they exist. For non-redundant serial robots, the aspects defined in this way are the largest domains $\cal D$ free of any singularity. For  any noncuspidal robot, the aspects are the largest uniqueness domains of the kinematic map $\bf{f}$, i.e. there is only one inverse solution in each aspect. \\
Aspects were later generalized to parallel robots \cite{chablat1998domaines}, see section 7.1.

As an illustrative example, let us consider a 3R robot as in Fig. \ref{fig:SerPar} (left), whose geometric parameters are: $d_2=1$, $d_3= 2$, $d_4= 1.5$, $r_2= 1$, $r_3=0$, $\alpha_2 = -90^{\circ}$ and $\alpha_3 = 90^{\circ}$. The dimensions are unit-less, without importance for the understanding of the example. 
This robot is called \emph{orthogonal} because its axes are mututally orthogonal. We also assume that the robot has no joint limits. This robot will be used in several examples throughout the first part of this paper.
We show that the determinant of the Jacobian matrix \textbf{J} of this robot can be written as a product of two factors (see for example \cite{el1996analyse})~:
\begin{eqnarray}
&&\det ({\textbf{J}}) = \big(d_3 + \cos(\theta_3)d_4\big)\big(\cos(\theta_2)(\sin(\theta_3)d_3 - \cos(\theta_3)r_2) + \sin(\theta_3)d_2\big) \nonumber
\end{eqnarray}

Note that the singularities, defined by $\det ({\textbf{J}}) = 0$, do not depend on $\theta_1$, which is always the case when the first joint is revolute. We can therefore represent the joint space in the $(\theta_2, \theta_3)$ plane. 
The first factor of det(\textbf{J}) cannot cancel here because $d_4<d_3$. The second factor defines two curves of identical shape and translated one with respect to the other by $\pi$ along the $\theta_3$ axis. The singularities then divide the joint space into two aspects (in green and blue in Fig.~\ref{fig:robot_cuspidal}, left). Indeed, as there are no joint boundaries, it is necessary to identify the opposite sides of the joint space, which has the topology of a torus. 

In the workspace, the singularities define boundary surfaces. The workspace is 3-dimensional but since the singularities do not depend on $\theta_1$ and there is no joint boundary, the workspace is symmetrical around the robot axis 1. It is therefore sufficient to represent only a half-section in a plane passing through this axis. Such a section, called \emph{cross section}, can be defined in the plane $(\rho, z)$ where $\rho = \sqrt {{x^2} + {y^2}}$. The total workspace can be obtained by rotating this section by 360° around axis 1 of the robot. For the robot under study, the singularities generate two closed curves in the cross section of the workspace. One defines the outer boundary, the other defines the inner boundary. The inner boundary separates a central region accessible with four \vmc{solutions}, from a peripheral region accessible with two \vmc{solutions} (Fig.~\ref{fig:robot_cuspidal}, right).

\section{Cuspidal robot~: definition, identification and properties}
\subsection{Definition and example}

A robot is said \emph{cuspidal} if it can \vmc{move from one solution to another without passing through a singularity}. There are thus several solutions in a single aspect. In other words, this aspect is no longer a uniqueness domain of the kinematic map $\bf{f}$. 

\begin{figure}
\centering
\begin{minipage}[t]{0.5\linewidth}
\includegraphics[width=\textwidth]{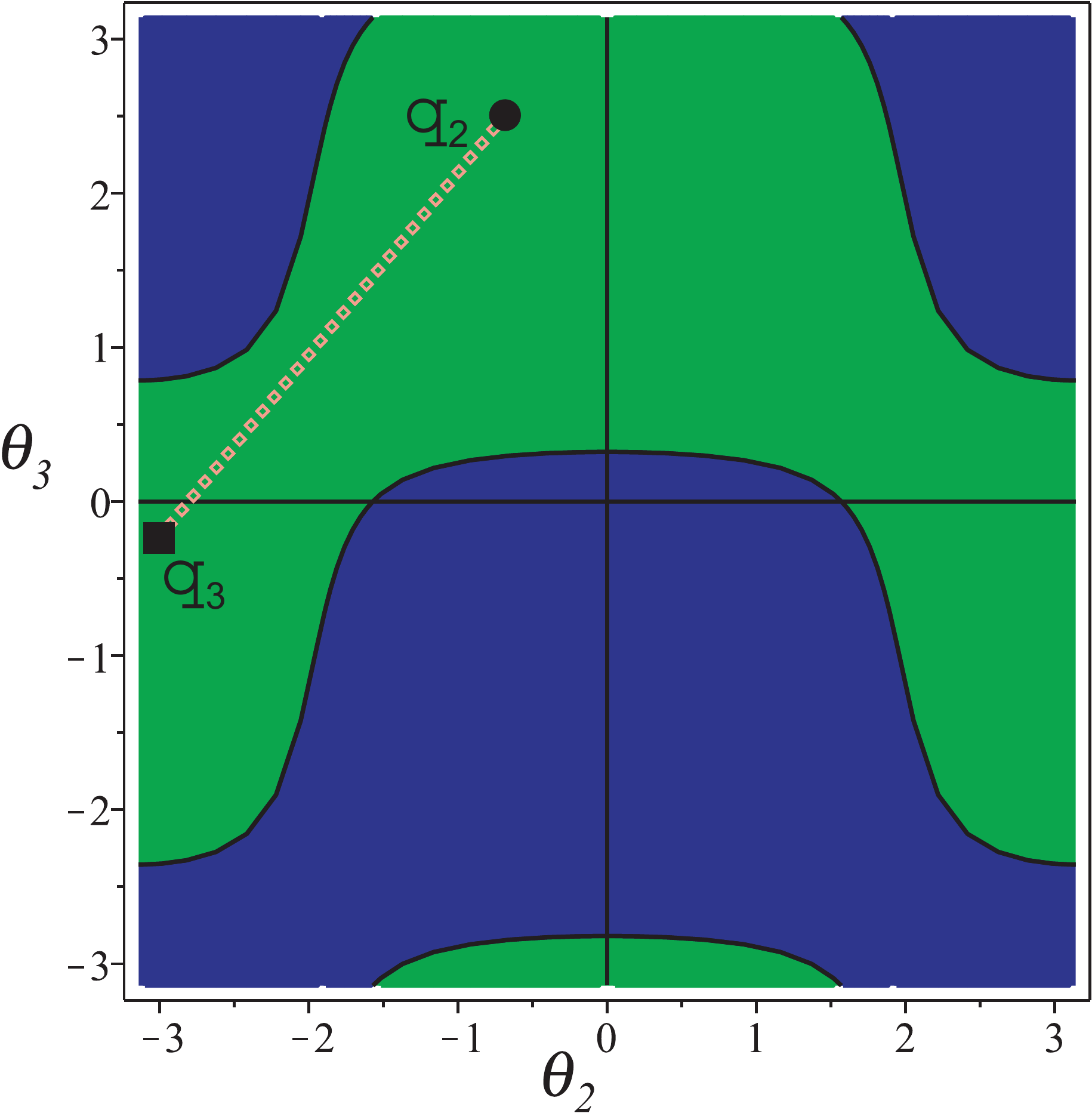}
\end{minipage}
\begin{minipage}[t]{0.4\linewidth}
\includegraphics[width=\textwidth]{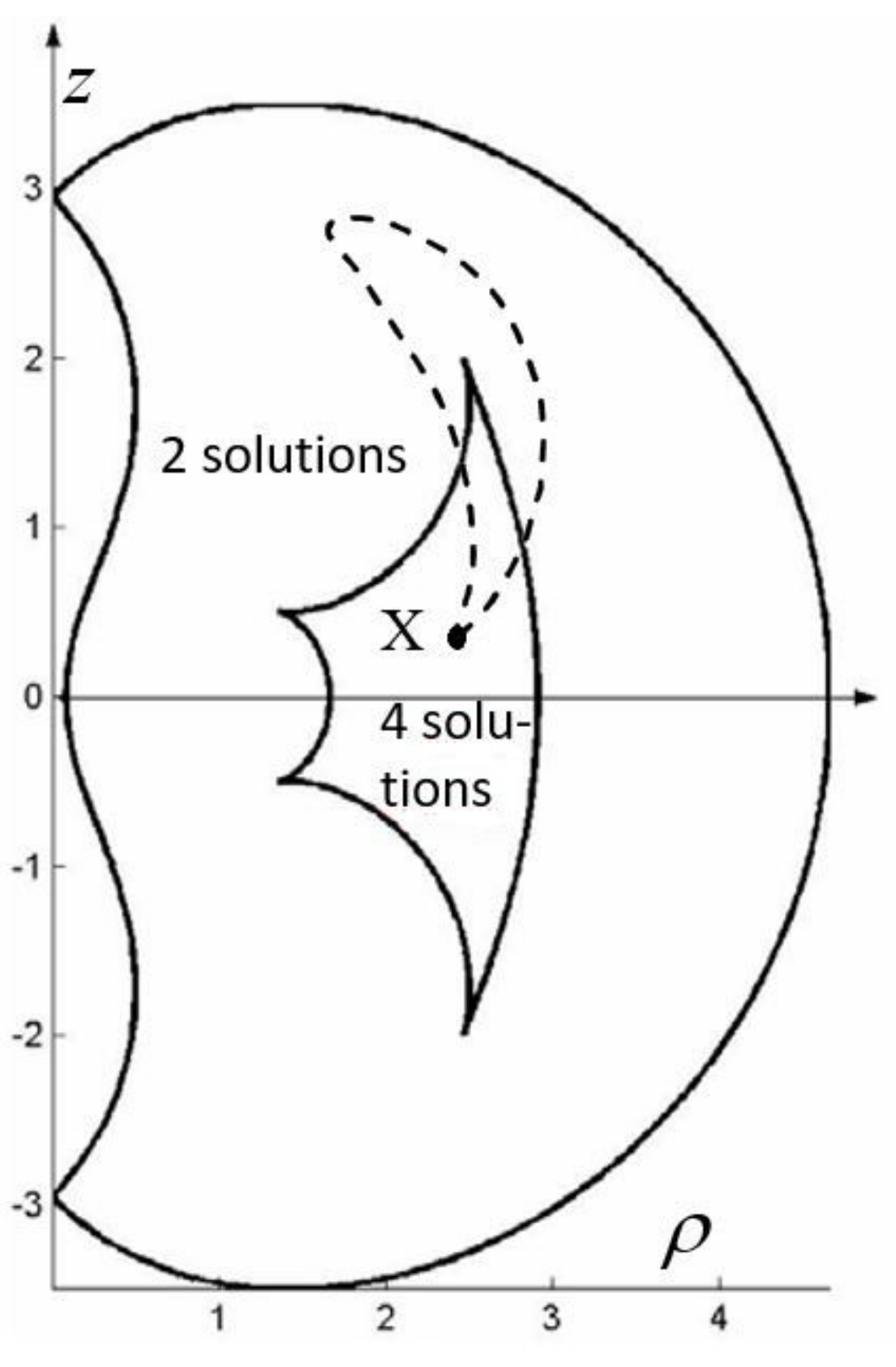}
\end{minipage}
\caption{3R cuspidal robot~: the two aspects in the joint space (left, units in radians) and cross section of the workspace (right). A non-singular solution changing trajectory is shown.}
\label{fig:robot_cuspidal}
\end{figure}

The name ``cuspidal robot'' was coined in connection with the existence condition of a particular singular point of the workspace, called \emph{cusp point}. We can show that if a cusp point exists, then the robot is cuspidal \cite{el1995recognize}. The exhaustive classification of orthogonal 3R robots that was done afterwards, shows that the existence of a cusp point is also a necessary condition for these robots \cite{baili2004classification}, \cite{Baili2004analyse}. In January 2022, a mathematical proof that the same is true for all 3R robots has been established in the framework of a France-Austria project \cite{wenger_generic_2022}. Note that, unfortunately, this necessary condition does not hold for any robot, in particular when the robot is parallel (see §\ref{sec:Robotsanscusp}).

A \emph{cusp} is one of two stable singularities formed when a folded surface is projected onto a plane, the other singularity being a "fold" \cite{Whitney55}. These singularities are called stable because they do not disappear under the effect of a small perturbation of the surface.

Let us take the case of the 3R robot from previous example ($d_2=1$, $d_3= 2$, $d_4= 1.5$, $r_2= 1$, $r_3=0$, $\alpha_2 =  -90^{\circ}$ and $\alpha_3 = 90^{\circ}$).

The point \textbf{X} of coordinates $x = 2.5$, $y = 0$, $z = 0.5$ in the reference frame $(O, x, y, z)$ (see Fig.~\ref{fig:robot_cuspidal}, right) is reachable with four \vmc{solutions} corresponding to the following joint configurations (values in radians)~:
\begin{eqnarray}
\textbf{q_1}=[-1.8,-2.8,1.9]^t,&\textbf{q_2}=[-0.9,-0.7,2.5]^t\nonumber \\
\textbf{q_3}=[-2.9,-3,-0.2]^t,&\textbf{q_4}=[0.2,-0.3,-1.9]^t\nonumber
\end{eqnarray}

\begin{figure}
	\centering
	\includegraphics[width=0.8\linewidth]{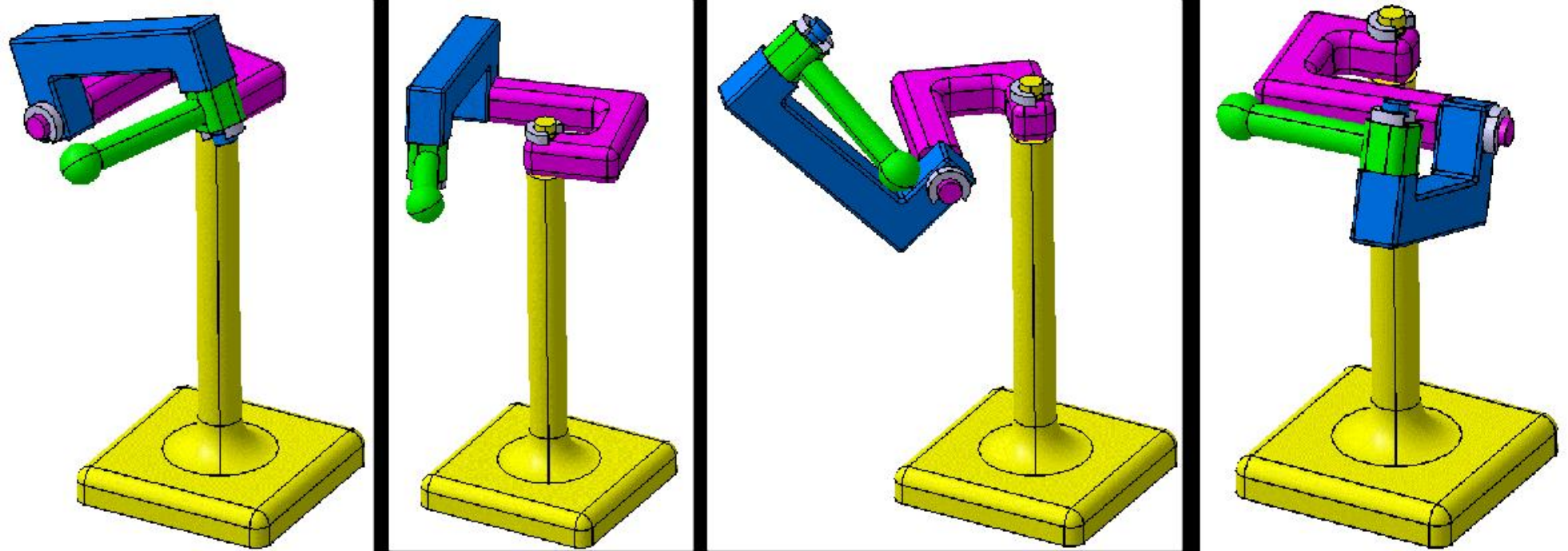}
	\caption{The four \vmc{solutions} of the 3R robot at point $x = 2.5$, $y = 0$, $z = 0,5$.}
	\label{fig:PosturesCusp}
\end{figure}

The four \vmc{solutions} are depicted in Fig.~\ref{fig:PosturesCusp}.

From Fig.~\ref{fig:robot_cuspidal}, we see that the configurations $\bf{q_2}$ and $\bf{q_3}$ belong to the same aspect. Therefore, the passage from one to the other can be realized without crossing singularities, for example through a linear trajectory (left). The trajectory connecting $\bf{q_2}$ and $\bf{q_3}$ then generates a loop in the workspace (right).

We can show that three inverse kinematic solutions coincide when the robot is on a cusp point. This property is very important because it allows us to search for the cusp points as triple roots of the characteristic polynomial of the inverse kinematics. 

\begin{figure}
	\centering
	\includegraphics[width=0.65\linewidth]{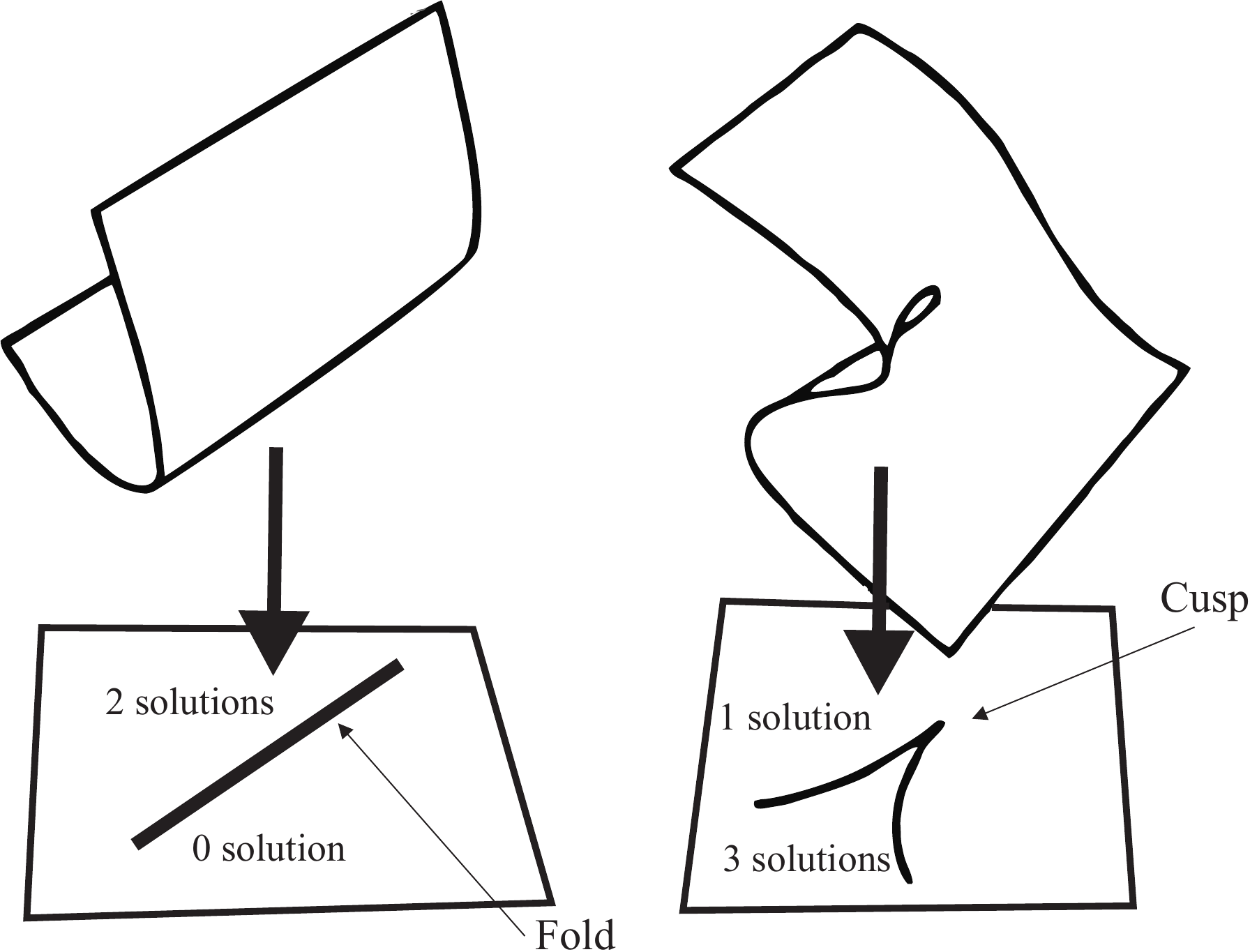}
	\caption{The two stable singularities of type ``fold'' (left) and ``cusp'' (right) obtained upon projection of a folded surface onto a plane. On the left, a simple fold is obtained; on the right, a cusp appears.}
	\label{fig:FoldCusp}
\end{figure}

Under the effect of the kinematic map $\bf{f}$, the joint space is "folded" along the singularities of the robot to form several "layers", then projected onto the workspace, forming fold lines. Each layer corresponds to an inverse kinematic solution of the robot. Figure~\ref{fig:FoldCusp} illustrates this phenomenon using a surface, representing the reachable joint space, projecting onto a plane, representing the workspace. With a single fold (left), the area to the left of the fold line in the projection plane results from the projection of two layers, corresponding to two inverse solutions. When the fold is like in Fig.~\ref{fig:FoldCusp}, right, the projection results in two fold lines that meet at a common point, forming a cusp. In the projection plane, the region inside the two converging fold lines results from the projection of three layers, corresponding to 3 inverse solutions.  

\subsection{Singular and non-singular change of \vmc{solution}}
A noncuspidal robot such as the anthorpomorphic robot in Fig.~\ref{fig:SerielCoude} can only perform singular \vmc{solution} changes. During such a \vmc{solution} change, the robot performs a back-and-forth trajectory towards a boundary of its workspace: in moving from "elbow up" to "elbow down", the anthropomorphic robot passes through the fully extended arm singularity. In doing so, the end-effector moves to the outer boundary of its workspace, reaches it, and then moves back to its initial pose. 

\begin{figure}
	\centering
	\includegraphics[width=0.8\linewidth]{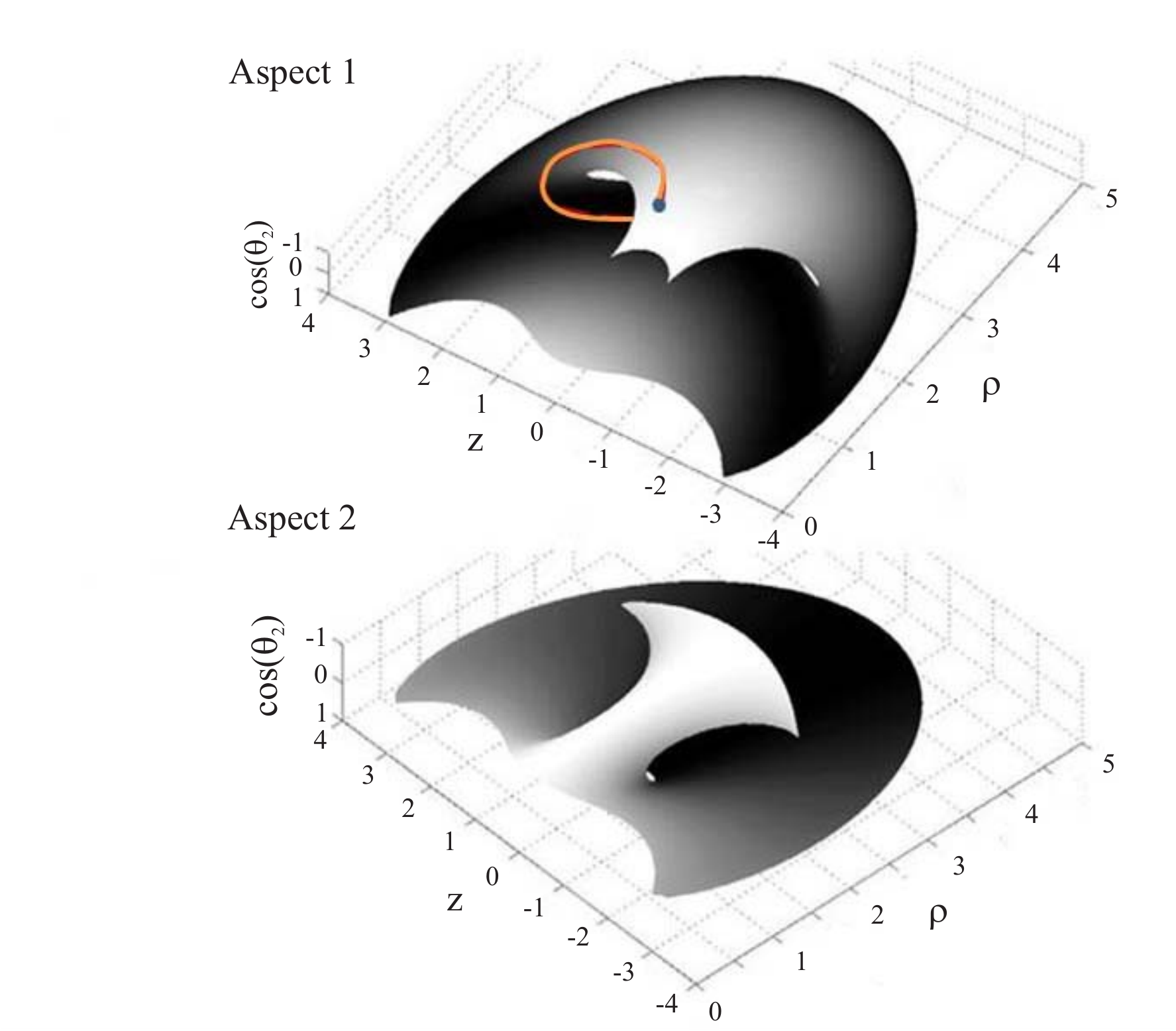}
	\caption{Folding of the two aspects in the workspace for the orthogonal 3R robot and non-singular \vmc{solution} changing trajectory around a cusp.}
	\label{fig:Pliage}
\end{figure}

For a cuspidal robot, Fig. \ref{fig:robot_cuspidal}, right, shows that a non-singular change of \vmc{solution} occurs by encircling a cusp point. To better understand this phenomenon, it is useful to show how, for this orthogonal 3R robot, the reachable joint space is ``folded'' before being projected onto the workspace. To do this, we can show the ``layers'' associated with the different inverse kinematic solutions by drawing the cross section of the workspace as a surface in the space $(\rho, z, \cos(\theta_2)$. The additional coordinate $\cos(\theta_2)$ allows us to distinguish the different inverse solutions that then appear on different layers. The folding pattern of the joint space being more complex than the one schematized in Fig.~\ref{fig:FoldCusp}, it is necessary to separate the representation according to the two aspects (Fig.~\ref{fig:Pliage}). It is necessary to imagine the global folding scheme considering that the two surfaces corresponding to the two aspects are actually part of the same surface that is folded at the outer edges and part of the inner edges. The separate representation shows how one can go from one solution to the other in the same aspect by encircling a cusp. 

\begin{figure}
	\centering
	\includegraphics[width=1\linewidth]{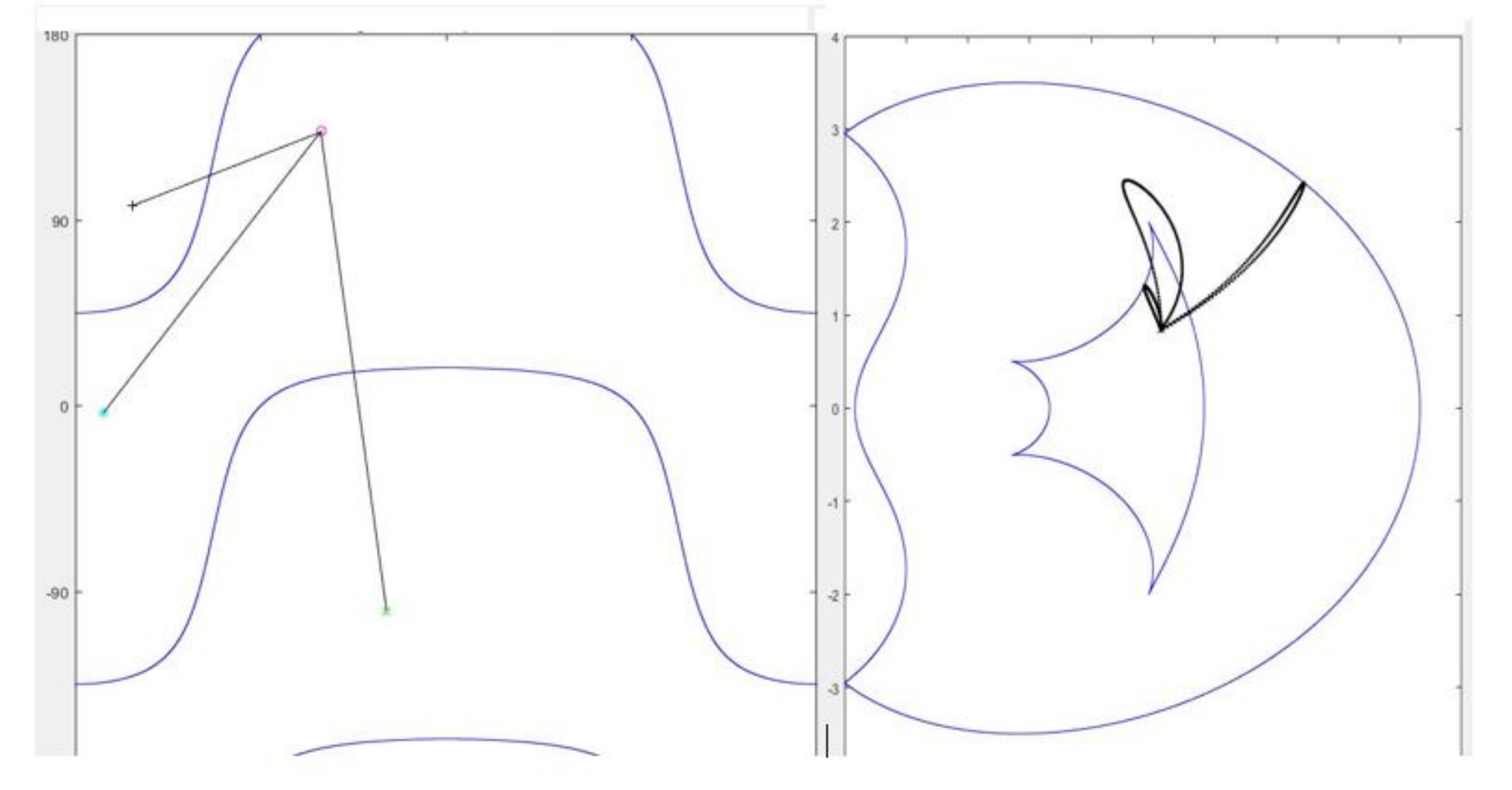}
	\caption{Singular and non-singular change of \vmc{solution} of the cuspidal 3R robot of Fig. \ref{fig:SerPar}, in joint space (left) and workspace (right).}
	\label{fig:SingNonsing}
\end{figure}

A cuspidal robot can also make a singular change of \vmc{solution}. This is the case when the robot of Fig. \ref{fig:SerPar} has to change its \vmc{solution} when it is in the peripheral region of its workspace. In this region, the robot admits only two \vmc{solutions}, one in each aspect. The change of \vmc{solution} will then produce a back and forth trajectory towards the outer boundary of the workspace, as for a noncuspidal robot. When the robot is in the central area, there are four \vmc{solutions}, two in each aspect. If the initial and final \vmc{solutions} are in two different aspects, the \vmc{solution} change will be singular. The robot will then perform a back and forth trajectory, either towards the inner boundary that separates the central region from the peripheral region, or towards the outer boundary. Figure \ref{fig:SingNonsing} shows, starting from the central region, two singular \vmc{solution} changing trajectories (round-trip trajectories to the inner boundary and to the outer boundary) and one non-singular \vmc{solution} changing trajectory encircling the upper right cusp.

\subsection{Identification of cuspidal robots}
We know that if there is a cusp point in the workspace of a robot, then it is cuspidal. We also know that when the robot is an orthogonal 3R robot, the existence of a cusp point is also a necessary condition. In other words, if there is no cusp point in the workspace cross-section of an orthogonal 3R robot, then this robot is noncuspidal. Numerical, graphical or algebraic methods can be used to verify the existence conditions of cusp points, and they are an interesting help for the designer. 

Since the existence of a cusp point proves that the robot is cuspidal, this can be used as a classification criterion.
We know that a cusp point is associated with a triple root of the characteristic polynomial of the inverse kinematics. We can therefore search for these triple roots. If there is at least one, we can say that the robot is cuspidal. The triple roots can be computed upon solving the following system:
\begin{equation}
\left\{ \begin{array}{l}
P(t) = 0\\
\frac{{\partial P(t)}}{{\partial t}} = 0\\
\frac{{{\partial ^2}P(t)}}{{\partial {t^2}}} = 0
\end{array} \right.
\label{eq:condition_cusp}
\end{equation}
where $P(t)$ is the characteristic polynomial of the robot.
System~\eqref{eq:condition_cusp} generates high-degree polynomial equations, which are difficult to solve with existing standard computer algebra tools \cite{el1996analyse}. 

\emph{Example}: 
	Let us consider a 3R robot such that $r_3=0$. Upon setting $d_2=1$ without loss of generality (normalization), only three parameters need be considered:  ${d_3}$,  ${d_4}$ and  ${r_2}$.
	The direct kinematic equations can thus be written as :
	\begin{equation}
\left\{ \begin{array}{c}
x = ({d_3} + {d_4}\cos {\theta _3})(\cos {\theta _1}\cos {\theta _2})
 - ({r_2} + {d_4}\sin {\theta _3})\sin {\theta _1} + \cos {\theta _1}\\
y = ({d_3} + {d_4}\cos {\theta _3})(\sin {\theta _1}\cos {\theta _2})
 + ({r_2} + {d_4}\sin {\theta _3})\cos {\theta _1} + \sin {\theta _1}\\
z =  - ({d_3} + {d_4}\cos {\theta _3})\sin {\theta _2}
\end{array} \right.
\label{MGD}
	\end{equation}
	The inverse kinematics is obtained by eliminating two of the three joint variables, for example $\theta_1$ and $\theta_2$. We then obtain the following equation \cite{pieper1968kinematics} :

\begin{equation}
    \begin{array}{l}
    {m_5}\cos {\theta _3}^2 + {m_4}\sin {\theta _3}^2 + {m_3}\cos {\theta _3}\sin {\theta _3}\nonumber
    + {m_2}\cos {\theta _3} + {m_1}\sin {\theta _3} + {m_0} = 0
    \end{array}
\end{equation}
where	
\begin{equation}
\begin{array}{l}
\left\{ {\begin{array}{*{20}{c}}
{{m_0} =  - {x^2} - {y^2} + {r_2}^2 + \frac{{{{(R + 1 - L)}^2}}}{4}}\\
{{m_1} = 2{r_2}{d_4} + (L - R - 1){d_4}{r_2}}\\
{{m_2} = (L - R - 1){d_4}{d_3}}
\end{array}} \right. \\
L = {d_3}^2+{r_3}^2+{r_2}^2+{d_4}^2\\
R = \rho^2+z^2\\
{\rm{and}}\quad  \left\{ {\begin{array}{*{20}{c}}
{{m_3} = 2{r_2}{d_3}{d_4}^2}\\
{{m_4} = {d_4}^2({r_2}^2 + 1)}\\
{{m_5} = {d_3}^2{d_4}^2}
\end{array}} \right.
\end{array}
\end{equation}

The characteristic polynomial is obtained by resorting to the half-angle tangent substitution $t=\tan(\theta_3/2)$. We then obtain a polynomial of degree four in $t$, whose coefficients depend on $R$, $z$, $d_3$, $d_4$, $r_2$ \cite{Baili2004analyse}. To know if the robot is cuspidal or not, we must know if the polynomial admits at least one real triple root or not. This is equivalent to looking for the existence of real solutions to the following system:
\begin{equation}
\begin{array}{l}
\left\{ \begin{array}{l}
P(t,{d_3},{d_4},{r_2},R,z) = 0\\
\frac{{\partial P}}{{\partial t}}(t,{d_3},{d_4},{r_2},R,z) = 0\\
\frac{{{\partial ^2}P}}{{\partial {t^2}}}(t,{d_3},{d_4},{r_2},R,z) = 0\\
\frac{{{\partial ^3}P}}{{\partial {t^3}}}(t,{d_3},{d_4},{r_2},R,z) \ne 0
\end{array} \right.\\
\end{array}
\label{eq:condition_3R}
\end{equation}
where
\begin{itemize}
\item $t$, $R = \rho^2+z^2$, $z$ are the variables;
\item $d_3$, $d_4$, $r_2$ are the geometric parameters.
\end{itemize}

\begin{figure}
	\centering
	\includegraphics[width=0.35\linewidth]{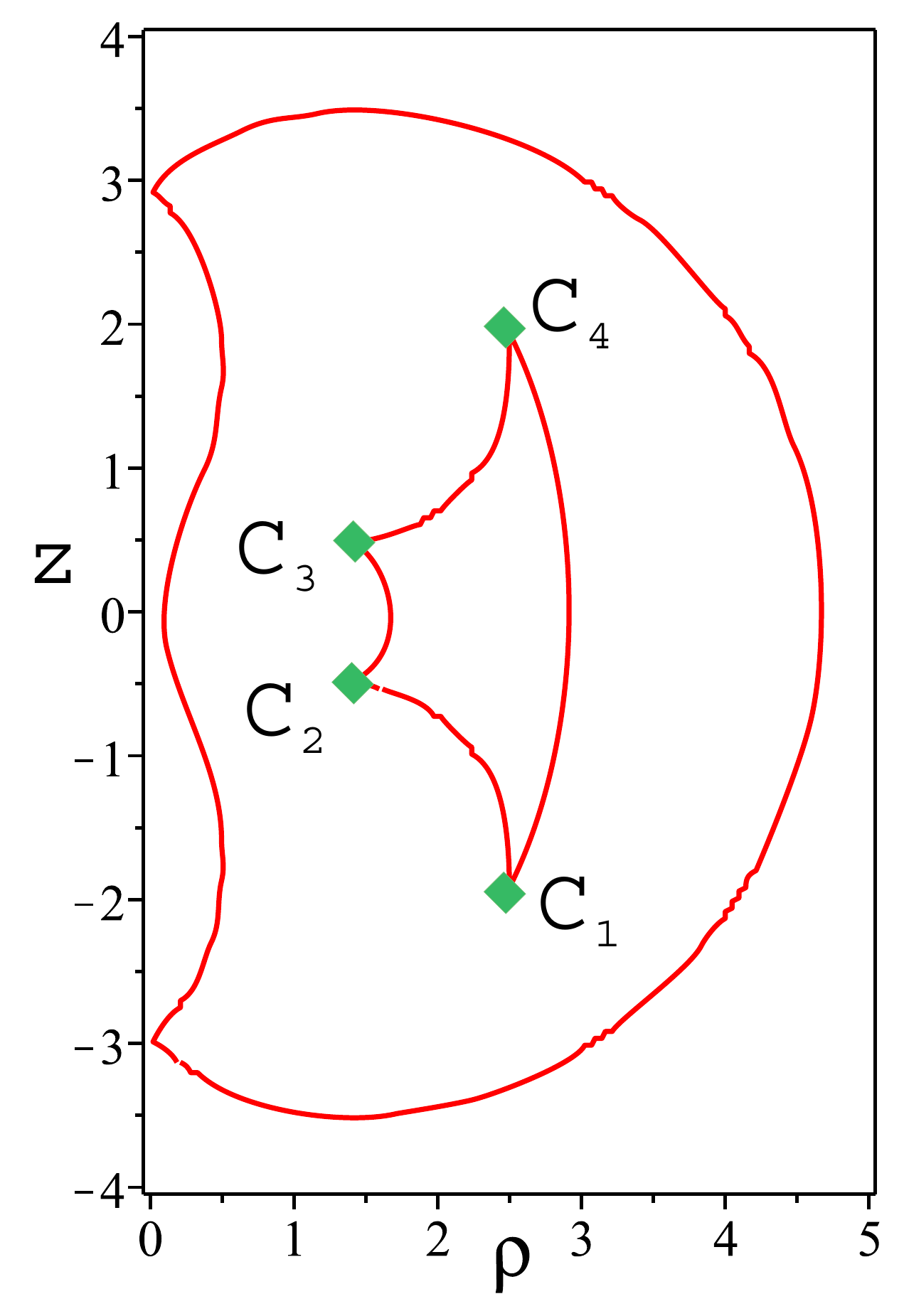}
	\caption{The four cusp points of the 3R robot defined by ($d_2=1$, $d_3= 2$, $d_4= 1.5$, $r_2= 1$, $r_3= 0$, $\alpha_2 =  -90^{\circ}$ and $\alpha_3 = 90^{\circ}$).}
	\label{fig:point_cusp_3r}
\end{figure}

For the robot of Fig. \ref{fig:SerPar} ($d_2=1$, $d_3= 2$, $d_4= 1.5$, $r_2= 1$, $r_3= 0$, $\alpha_2 = -90{\circ}$ and $\alpha_3 = 90^{\circ}$), the previous system admits four real solutions. In the plane $(\rho, z)$ corresponding to the workspace cross section, these values are~:
\begin{eqnarray}
C_1= [2.4655,-1.9987] \quad C_2= [1.3555,-0.5047] \\
C_3= [1.3555, 0.5047] \quad C_4= [2.4655,1.99872] 
\end{eqnarray}
Figure \ref{fig:point_cusp_3r} shows these four cusp points marked in green.

\section{Feasibility of trajectories and t-connected regions}
Planning trajectories for a cuspidal robot is more difficult than for a noncuspidal robot, even when there are no joint limits and no obstacles around the robot. The problem might become challenging when the robot is supposed to follow a predefined path in the workspace, such as a welding bead in arc-welding. The notion of \emph{t-connected regions} provides a global solution to the above-mentioned problem. 
A region of the workspace is said to be t-connected if any continuous path in that region can be followed by the end-effector without ever leaving the path during motion \cite{wenger1991ability}. 

\subsection{T-connected regions for a noncuspidal robot}

For a noncuspidal robot, it has been shown that aspects define the t-connected regions \cite{borrel1986study},
\cite{wenger1991ability}. This property is very interesting for the user or the designer of a robotic site because it gives him a global information on the performance of the robot in its workspace. In the absence of obstacles that could hinder the robot's movements, we can obtain the t-connected regions from their boundaries easily. The image of the singularities and joint limits in the workspace is drawn using the direct kinematic map \textbf{f}. 

\begin{figure}
	\centering
	\includegraphics[width=0.65\linewidth]{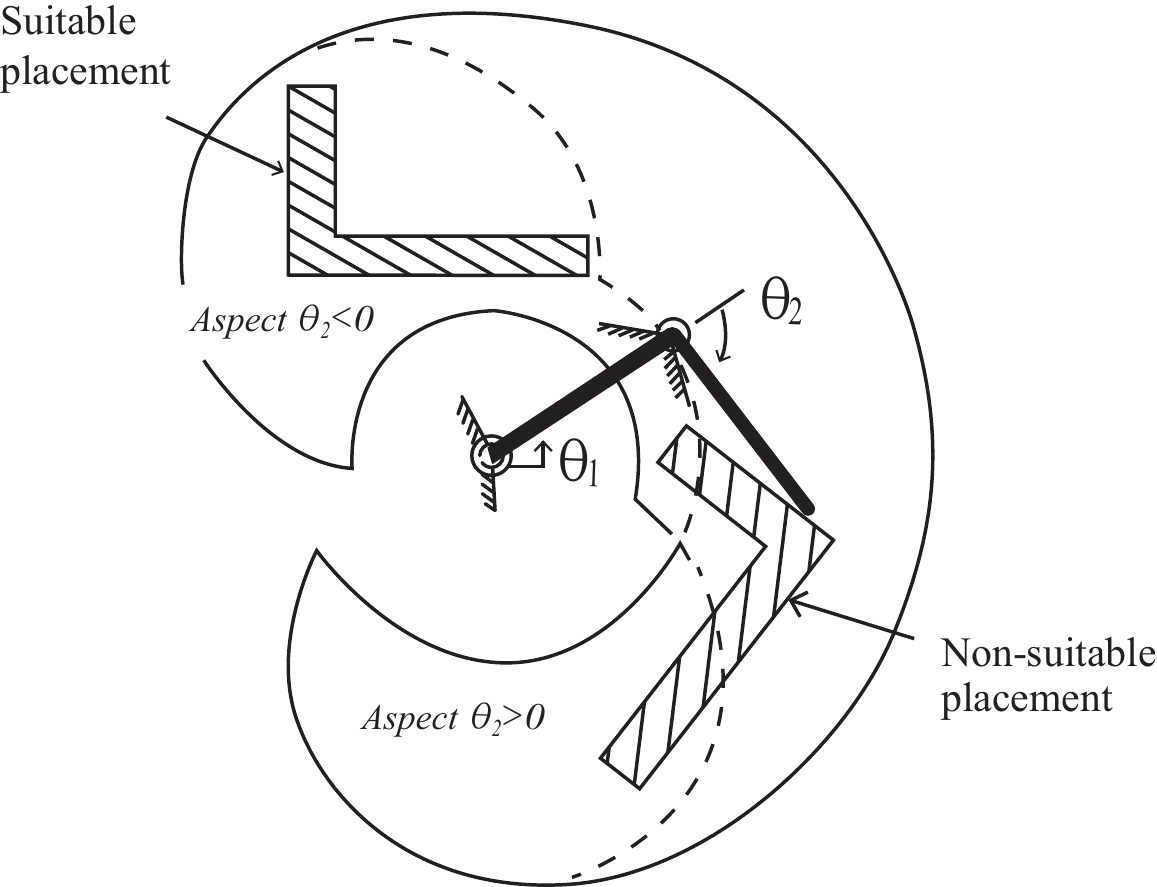}
	\caption{Using the t-connected regions of a noncuspidal robot for part placement.}
	\label{fig:Tparcourable}
\end{figure}

Figure~\ref{fig:Tparcourable} shows the two t-connected regions of a planar 2-revolute-jointed robot that correspond to aspects $\theta_2<0$ and $\theta_2>0$. In this example, the robot has joint limits, schematized by two hatched half-segments on each joint. The two t-connected regions overlap, their common outer boundary being a circular arc whose radius is the sum of the two arm lengths and which corresponds to the "extended arm" singularity defined by $\theta_2=0$. The other region boundaries are arcs of smaller radius defined by the joint limits. The dashed boundaries show the overlap area. For an "L" shaped part cutting task, the designer will place the part so that the entire contour is in one t-connected region, which is the case for the top placement, but not for the bottom since the ends of the contour are then in two different t-connected regions.

\subsection{T-connected regions for a cuspidal robot}
\begin{figure}
	\centering
	\includegraphics[width=0.45\linewidth]{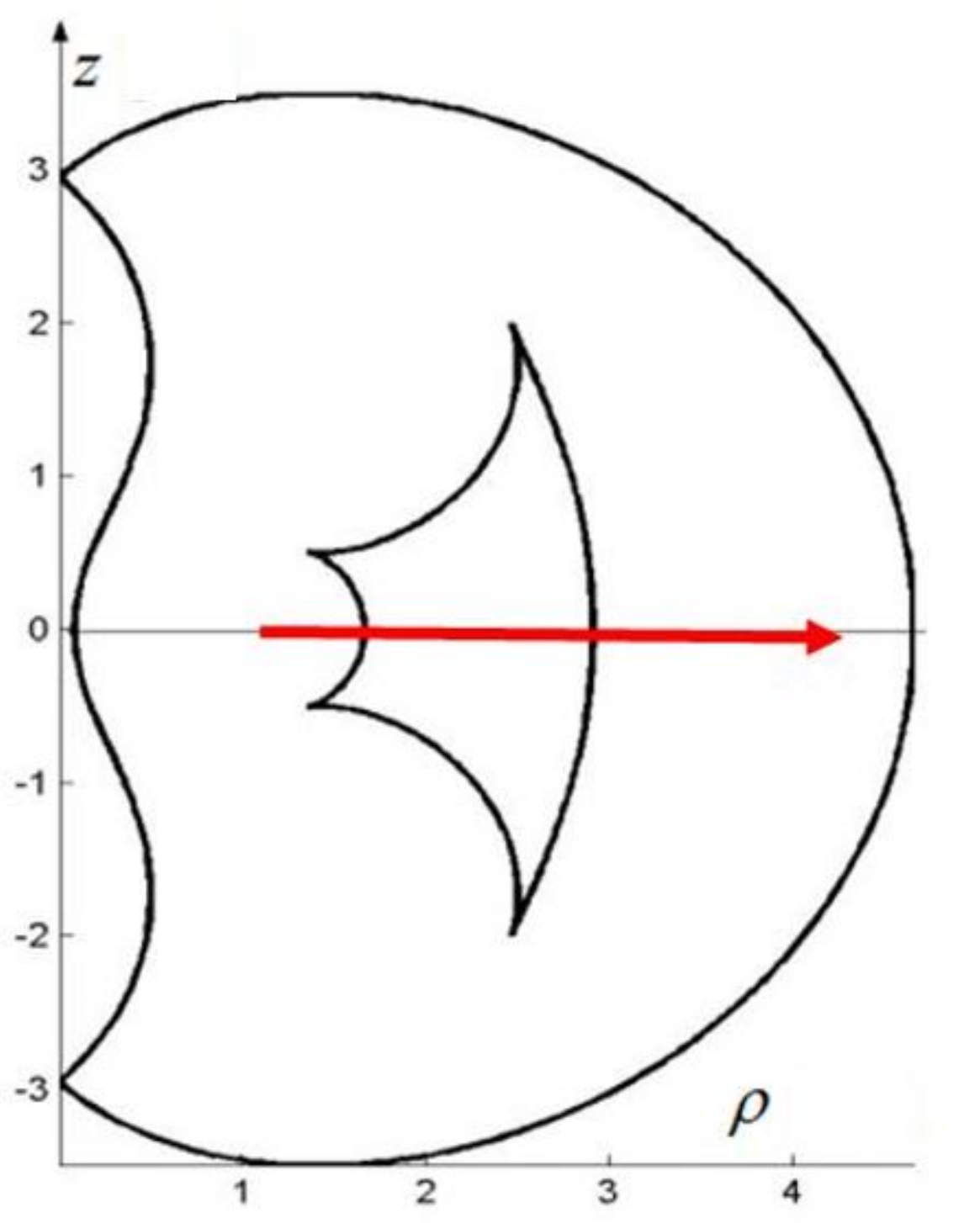}
	\caption{Infeasible trajectory in one aspect of a cuspidal 3R robot.}
	\label{fig:Infeasible}
\end{figure}

The existence of cuspidal robots has major consequences on the feasibility study of trajectories. Indeed, when there is more than one solution in an aspect, this aspect does not guarantee the existence of continuous trajectories anymore. Figure~\ref{fig:Infeasible} shows a trajectory which is totally accessible in the same aspect of the orthogonal 3R robot but which is not feasible. To convince oneself of this, one can refer to Fig.~\ref{fig:Pliage}: whatever the aspect chosen, the robot will be blocked at an inner edge of the aspect. In fact, it is not possible to successively cross the two inner left and right boundaries which delimit the 4-solution area.
 
For a cuspidal robot, the t-connected regions can therefore no longer be defined from the aspects. It is necessary to define new uniqueness domains. 

To do so, we have to introduce new boundaries that separate the multiple solutions within an aspect. These boundaries are called \emph{characteristic surfaces} \cite{Wenger04} and are defined for each aspect.

The characteristic surfaces of an aspect ${\cal A}_j$, denoted ${\cal S}_C({\cal A}_j)$, are defined as the reciprocal image in ${\cal A}_j$ of the image $\bf{f}({\cal A}_j^*)$ of the boundaries ${\cal A}_j^*$ of ${\cal A}_j$ :  
   
\begin{equation}
{\cal S}_C({\cal A}_j) = \bf{f}^{-1} (\bf{f}({\cal A}_j^*)) \cap {\cal A}_j
\label{SC}
\end{equation}
   where:
   \begin{itemize}
        \item $\bf{f}$ is the kinematic map,
        \item ${\cal A}_j^*$ is the set of boundaries of aspect ${\cal A}_j$,
         \item $\bf{f}^{-1} (\bf{f}({\cal A}_j^*)) = \left\{ \bf{q} / \bf{f}(\bf{q}) \in \bf{f}({\cal A}_j^*)\right\}$.
   \end{itemize}

To better understand this definition, let us take the case of the orthogonal 3R robot. Figure~\ref{fig:Pliage} shows us that, in the same aspect, each point of an inner boundary of the workspace is accessible by two inverse kinematic solutions. One is a singular configuration and corresponds to an edge, for example for a position located on one of the edges of the upper "layer" of the aspect. The other is not singular but it is located on the lower "layer". The characteristic surfaces thus correspond to the non-singular joint configurations of an aspect associated with points located on the interior boundaries of the workspace.

\begin{figure}
	\centering
    \includegraphics[width=0.8\linewidth]{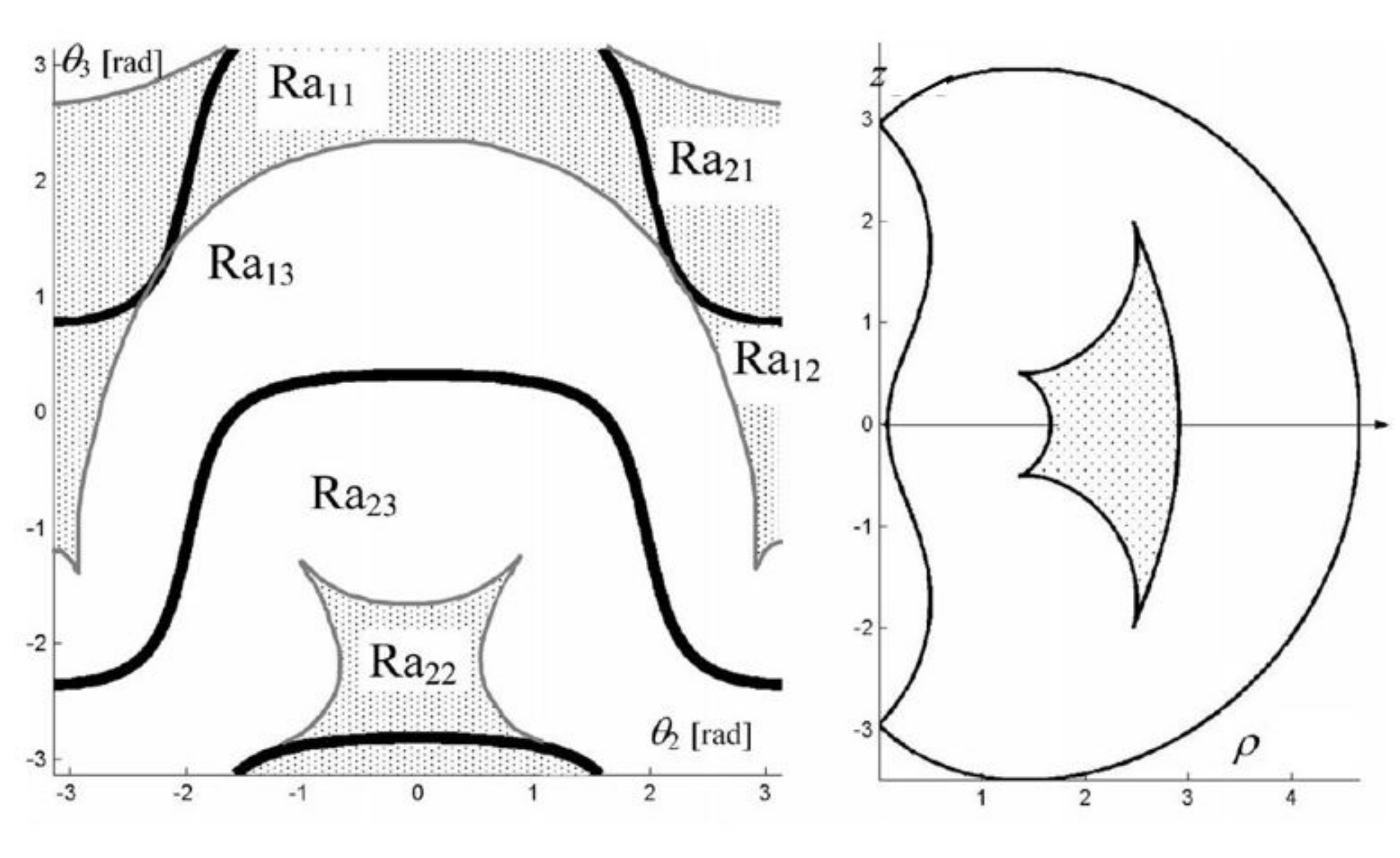}
	\caption{Left: singularities (in bold lines) and characteristic surfaces (in thin lines) for the orthogonal 3R robot ($d_2=1$, $d_3= 2$, $d_4= 1.5$, $r_2= 1$, $r_3= 0$). The shaded regions correspond to the central shaded region of the workspace (right).}
	\label{fig:surface_caracteristique_3r}
\end{figure}

Figure \ref{fig:surface_caracteristique_3r}, left, shows the characteristic surfaces of the orthogonal 3R robot. Recall that although the figure shows curves, they are indeed surfaces, as are the singularities, because the joint space is actually 3-dimensional. Referring to the figure on the left, we can see that the characteristic surfaces separate the two inverse solutions in the aspect. 

\begin{figure}
	\centering
	\includegraphics[width=0.8\linewidth]{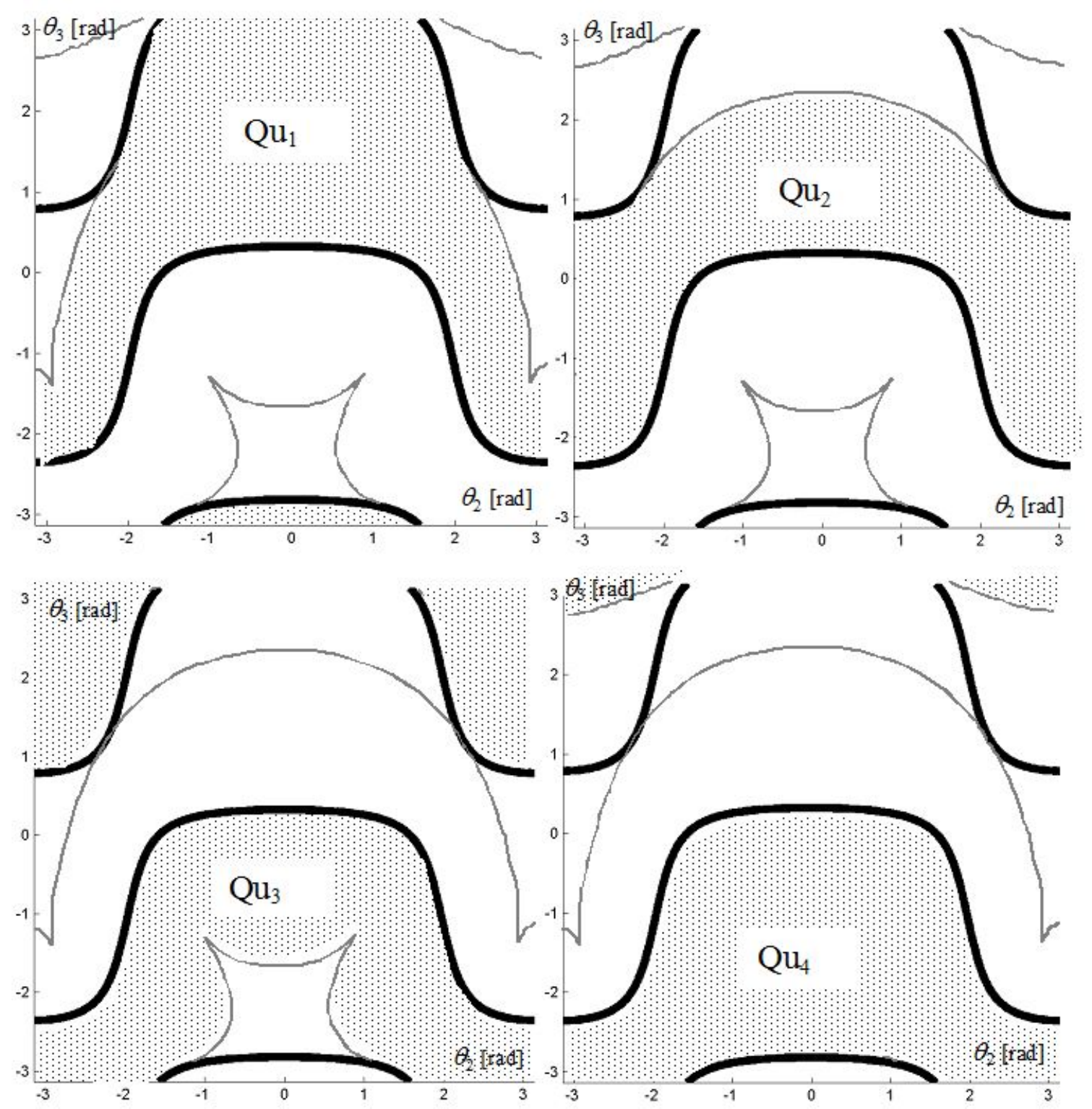}
	\caption{The four uniqueness domains for the orthogonal 3R robot ($d_2=1$, $d_3= 2$, $d_4= 1.5$, $r_2= 1$, $r_3= 0$).}
	\label{fig:domaine_unicite}
\end{figure}

\emph{Note:} For a noncuspidal robot, formula \eqref{SC} gives the empty set. In other words, a noncuspidal robot has no characteristic surfaces, which is logical since it has only one solution per aspect.

The characteristic surfaces allow us to define new uniqueness domains. Indeed, we can show that the characteristic surfaces realize a partition of the aspects into domains where the robot admits only one inverse solution \cite{Wenger04}. In Fig. \ref{fig:surface_caracteristique_3r}, these domains are identified by $Ra_{ij}$. The figure shows in gray the domains that correspond to the central region of the workspace. As there are four solutions in this central region, there are four associated domains in the joint space. Nevertheless, these domains are not the largest uniqueness domains. To obtain the latter, one must take the union of several adjacent domains, as explained in \cite{Wenger04}, where a numerical method is proposed to calculate them. Figure \ref{fig:domaine_unicite} shows the four uniqueness domains $Qu_i$ obtained for the orthogonal 3R robot: the uniqueness domain $Qu_1$, for example, is obtained as the union of the domains $Ra_{13}$ and $Ra_{11}$ of Fig. \ref{fig:surface_caracteristique_3r}.

\begin{figure}
	\centering
	\includegraphics[width=0.6\linewidth]{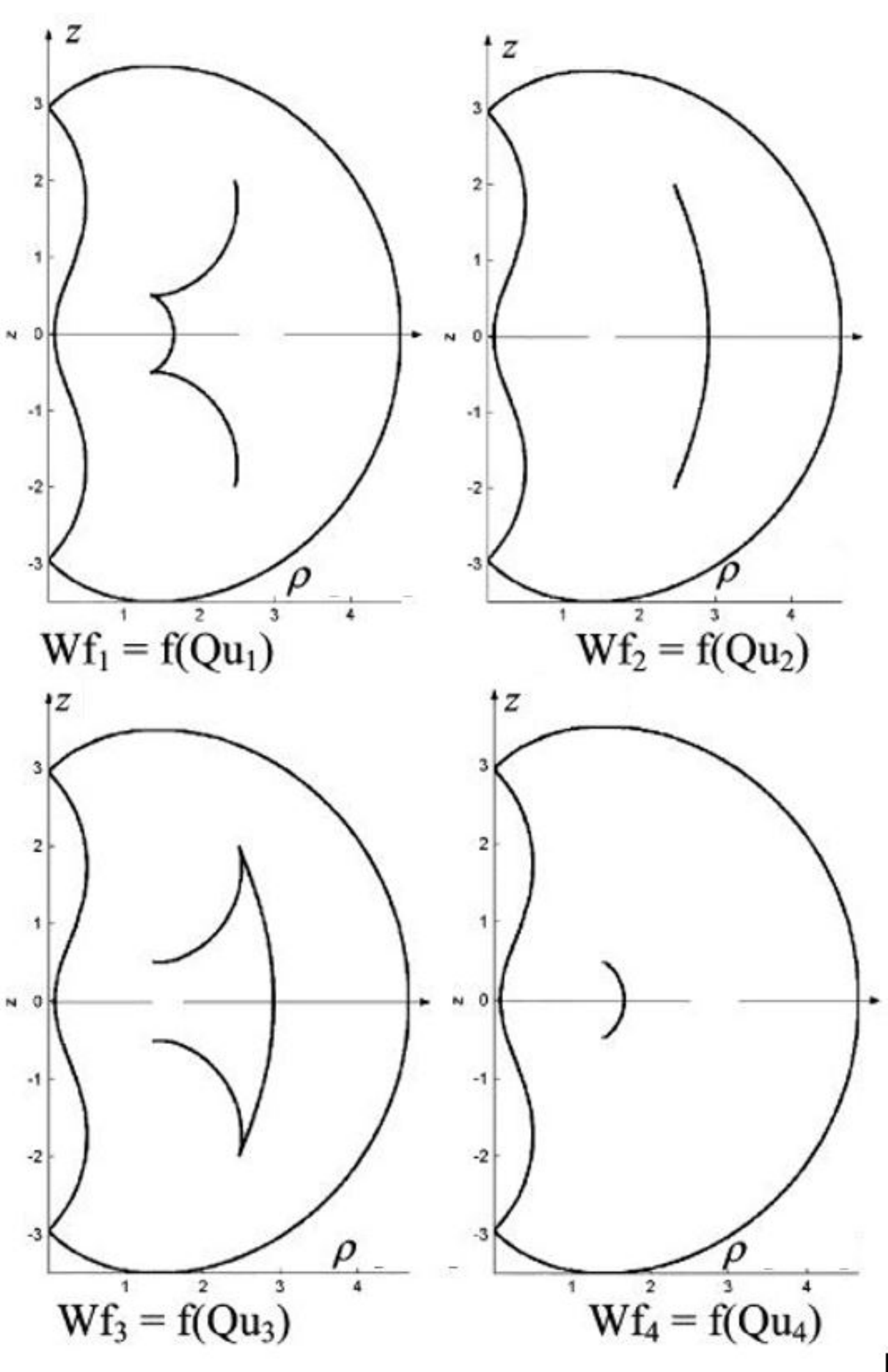}
	\caption{The four t-connected regions for the orthogonal 3R robot ($d_2=1$, $d_3= 2$, $d_4= 1.5$, $r_2= 1$, $r_3= 0$).}
	\label{fig:TparCusp}
\end{figure}

The t-connected regions are then obtained as the images of the largest uniqueness domains in the workspace. Figure \ref{fig:TparCusp} shows the four t-connected regions $Wf_i, i=1, ... 4$ of the orthogonal 3R robot. Regions $Wf_1$ and $Wf_2$ are associated with aspect 1, while regions $Wf_3$ and $Wf_4$ are associated with aspect 2. The inner lines are in fact empty line, i.e. no point is reachable on each of them. It is therefore not possible to cross them. We verify that the horizontal trajectory of Fig. \ref{fig:Infeasible} is not feasible since it encounters an empty line in each of the t-connected regions. A solution is to place the prescribed trajectory vertically: one can use region $Wf_2$ if the robot is initially in aspect 1, or $Wf_4$ if it is in aspect 2. 

\begin{figure}
\centering
\includegraphics[width=0.35\linewidth]{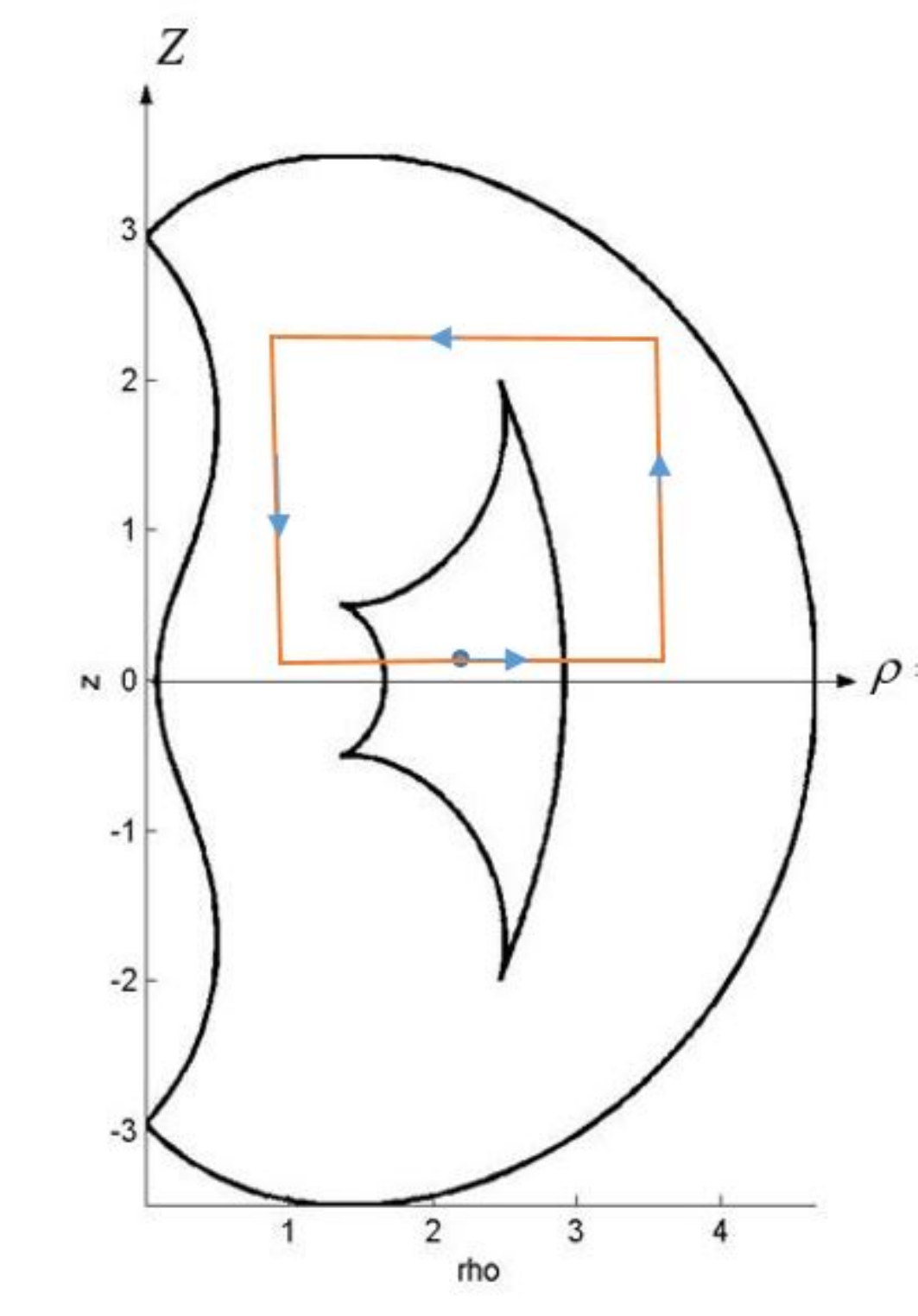}
\includegraphics[width=0.6\linewidth]{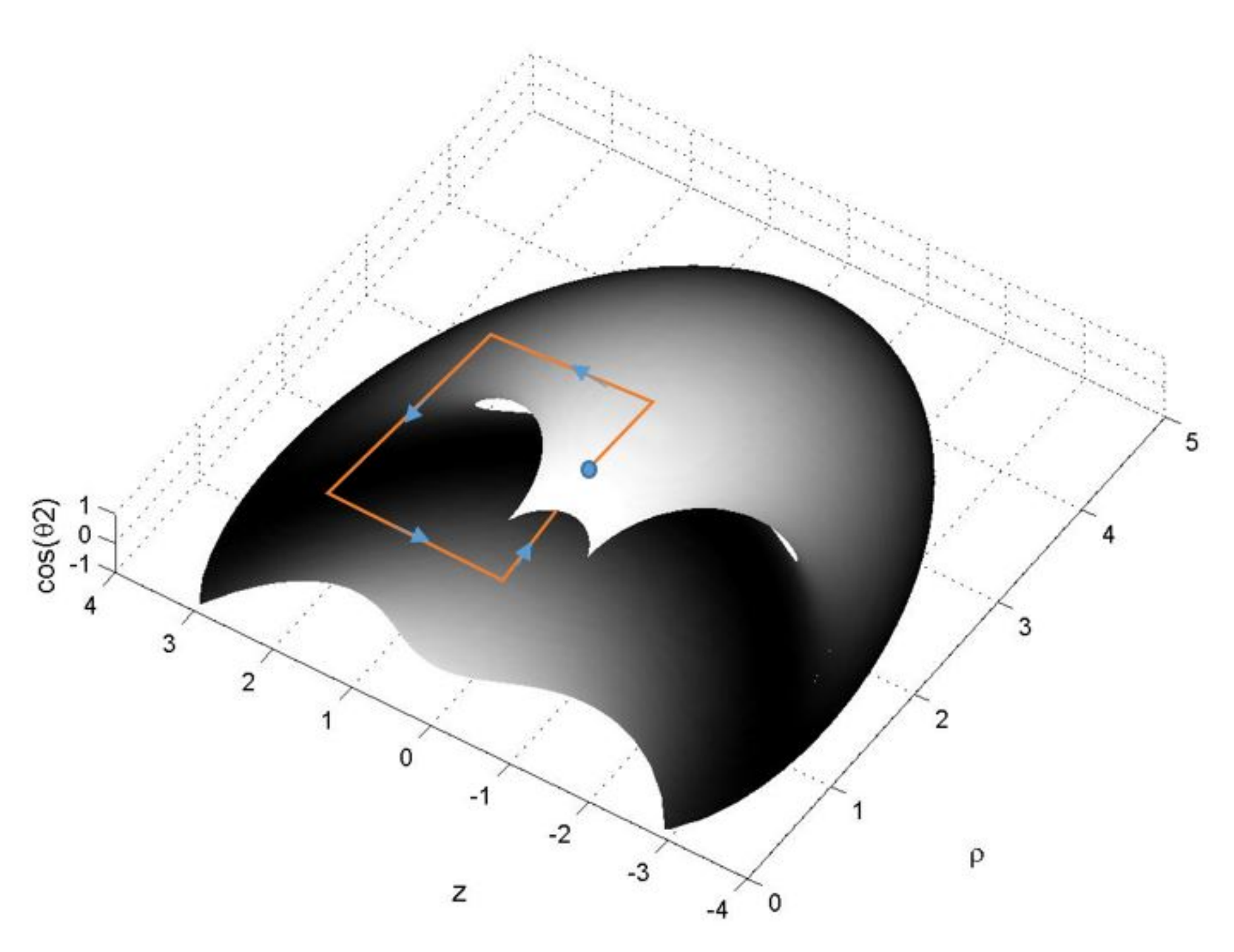}
\caption{This path can be tracked continuously but only once.}
\label{fig:OneTime}
\end{figure}

\vmc{\emph{Note}: By definition, any path enclosed in a t-connected region can be tracked continuously. In addition, \emph{it can be repeated as many times as desired}. This is important as many industrial robotic tasks are defined as a single loop trajectory that must be executed several times. In Fig. \ref{fig:OneTime}, it appears that the rectangular path depicted in orange is feasible, although it does not belong to any t-connected region. However, this closed path can be followed only once. To understand this behavior, one must realize that the robot has to perform a non-singular change of solution while tracking the path. To be feasible, indeed, the robot needs to start in a certain solution in aspect 1 (the one associated with the upper layer shown on the right) and to finish the path in the other solution in the same aspect (associated with the bottom layer). This means that the joint trajectory is not closed. Attempting to repeat the joint trajectory would block the robot against a singularity. In the workspace, it means that after the robot enters the bottom layer, it has to stop at the right edge before completing the closed path. This edge exists only in the bottom layer. Hence, the path cannot be repeated once more. Such a behavior is typical of cuspidal robots.}

\subsection{Effect of joint limits}
The characteristic surfaces and t-connected regions of cuspidal robots were illustrated with an orthogonal 3R robot without joint limits, but the theory and definitions also apply when the joints are limited. In practice, joint limits will remove singularities in the joint space (those beyond the joint space boundaries) but they will produce new boundaries for the aspects, defined by the joint space boundaries. In the workspace, some of the boundaries will disappear, while new ones will appear. In case of tight joint limits, it is possible that all cusps disappear: a cuspidal robot may then become noncuspidal.
The interested reader can find examples in \cite{el1996analyse} and \cite{Wenger04}.   

\section{Enumeration and classification of cuspidal and noncuspidal robots}
Identifying cuspidal/noncuspidal robots is of primary importance. Most industrial robots are noncuspidal and the reason if that industrial robots have many geometric simplifications in their design, such as intersecting and parallel joint axes. A first list of noncuspidal robots can be established based on geometric simplifications. Then, we will see that it is possible to enumerate all cuspidal/noncuspidal designs for a family of robots.

\subsection{Simple geometric conditions for a robot to be noncuspidal}
\label{sec:CondSimples}
Several simple geometric conditions leading to noncuspidal 3R robots have been established. Some conditions correspond to "solvable" robots in the sense of Pieper \cite{pieper1968kinematics}, i.e. whose inverse kinematics can be solved with polynomials of degree 2 or less. Such robots cannot have a cusp point since no triple root can exist with polynomials of degree 2, so they are noncuspidal. Thus, the following six conditions have been established for 3R robots \cite{Wenger97}:

\begin{enumerate}
\item first two axes are parallel; 
\item last two axes are parallel; 
\item first two axes intersect; 
\item last two axes intersect; 
\item first two axes are orthogonal and $r_2=r_3=0$ (i.e. no joint offset along the second and third axes); 
\item axes are orthogonal two by two and $r_2=0$;
\end{enumerate}

The first four conditions correspond to Pieper's solvable robots and are common on industrial robots. On the other hand, the last two conditions are less common. 

It is interesting to note that when the axes are orthogonal two by two (i.e. in orthogonal robots), the robot is cuspidal when $r_2 \neq 0$ and $r_3=0$ but it noncuspidal when $r_2=0$ and $r_3 \neq 0$.

Moreover, it can be shown that when two of its joints are prismatic, a serial robot with three degrees of freedom is necessarily noncuspidal because it is solvable \cite{pieper1968kinematics}. With only one prismatic joint, geometric conditions equivalent to conditions 1 to 4 above also exist \cite{el1996analyse}. 
\subsection{Classification of orthogonal 3R robots}
We have seen that the existence of a cusp point, corresponding to a triple point of the characteristic polynomial, indicates that the robot is cuspidal. Accordingly, we can find all cuspidal robots upon searching for all those robots whose characteristic polynomial has at least one triple point. 

To do this, starting from system \eqref{eq:condition_cusp}, we can look for the conditions on the geometric parameters so that this system admits real solutions. The task is rather difficult and of great algebraic complexity in general. However, it is possible to solve this problem for the family of orthogonal 3R robots from system \eqref{eq:condition_3R}. This requires the use of sophisticated algebraic tools, though. These tools, which are based on Groebner's bases and \emph{Cylindrical Algebraic Decomposition} \cite{Corvez04}, are available in the formal calculation software Maple, through the Siropa library \cite{chablat2019using}. In order to limit the number of parameters to three and to simplify calculations, we start by studying a subfamily of robots for which $r_3=0$, then we normalize the problem by posing, without loss of generality, $d_2=1$. We then look for the conditions for which the number of solutions to the system \eqref{eq:condition_3R} changes. We then obtain four surfaces $C_1, C_2, C_3, C_4$ in the space of parameters $(d_2, d_3, r_2)$. These surfaces are defined by the following equations \cite{baili2004classification}:

\begin{eqnarray}
    C_1:~~{d_4} = \sqrt {\frac{1}{2}({d_3}^2 + {r_2}^2 - \frac{{{{({d_3}^2 + {r_2}^2)}^2} - ({d_3}^2 - {r_2}^2)}}{{AB}})} \label{eq:4} \\
	C_2:~~{d_4} = \frac{{{d_3}}}{{1 - {d_3}}}B\,\,\,{\text{and}}\,\,\,\,{d_3} < 1 \label{eq:5}\\
    C_3:~~{d_4} = \frac{{{d_3}}}{{{d_3} - 1}}B\,\,\,\text{and}\,\,\,\,{d_3} > 1\label{eq:6}\\
    C_4:~~{d_4} = \frac{{{d_3}}}{{1 - {d_3}}}B\,\,\,\text{and}\,\,\,\,{d_3} < 1 \label{eq:7}
\end{eqnarray}

where~:
\begin{equation}\label{eq:8}
    A = \sqrt {{{({d_3} + 1)}^2} + {r_2}^2} \,\,\,\text{and}\,\,B = \sqrt {{{({d_3} - 1)}^2} + {r_2}^2} 
\end{equation}

\begin{figure}
	\centering
	\includegraphics[height=200pt, width=193pt]{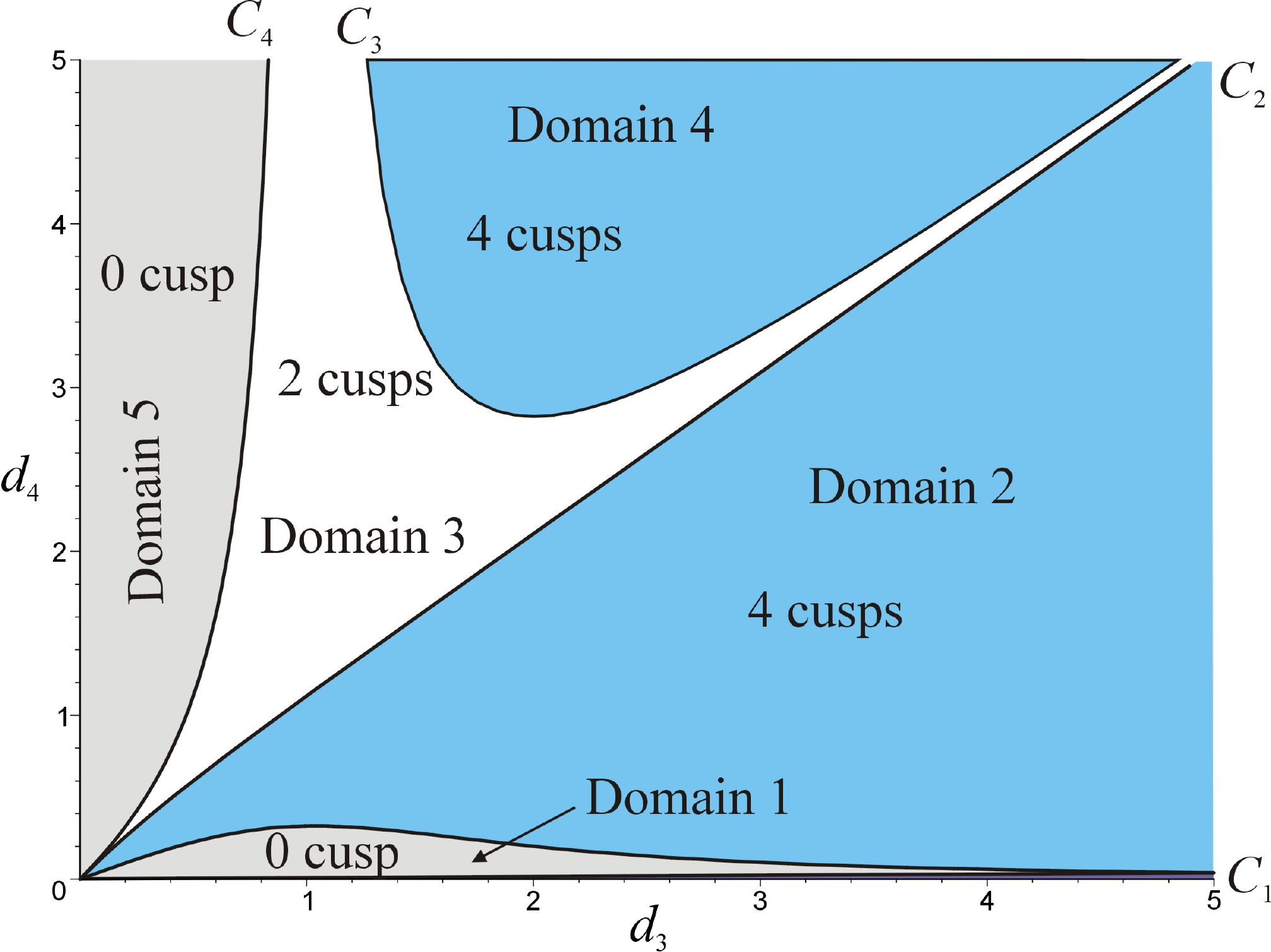}
	\caption{Partition of the parameter space as a function of the number of cusps for orthogonal 3R robots normalized by $d_2=1$ and such that $r_3= 0$ (section at $r_2=1$).}
	\label{fig:classification_3r}
\end{figure}

\begin{figure*}
	\centering
	\includegraphics[width=1\linewidth]{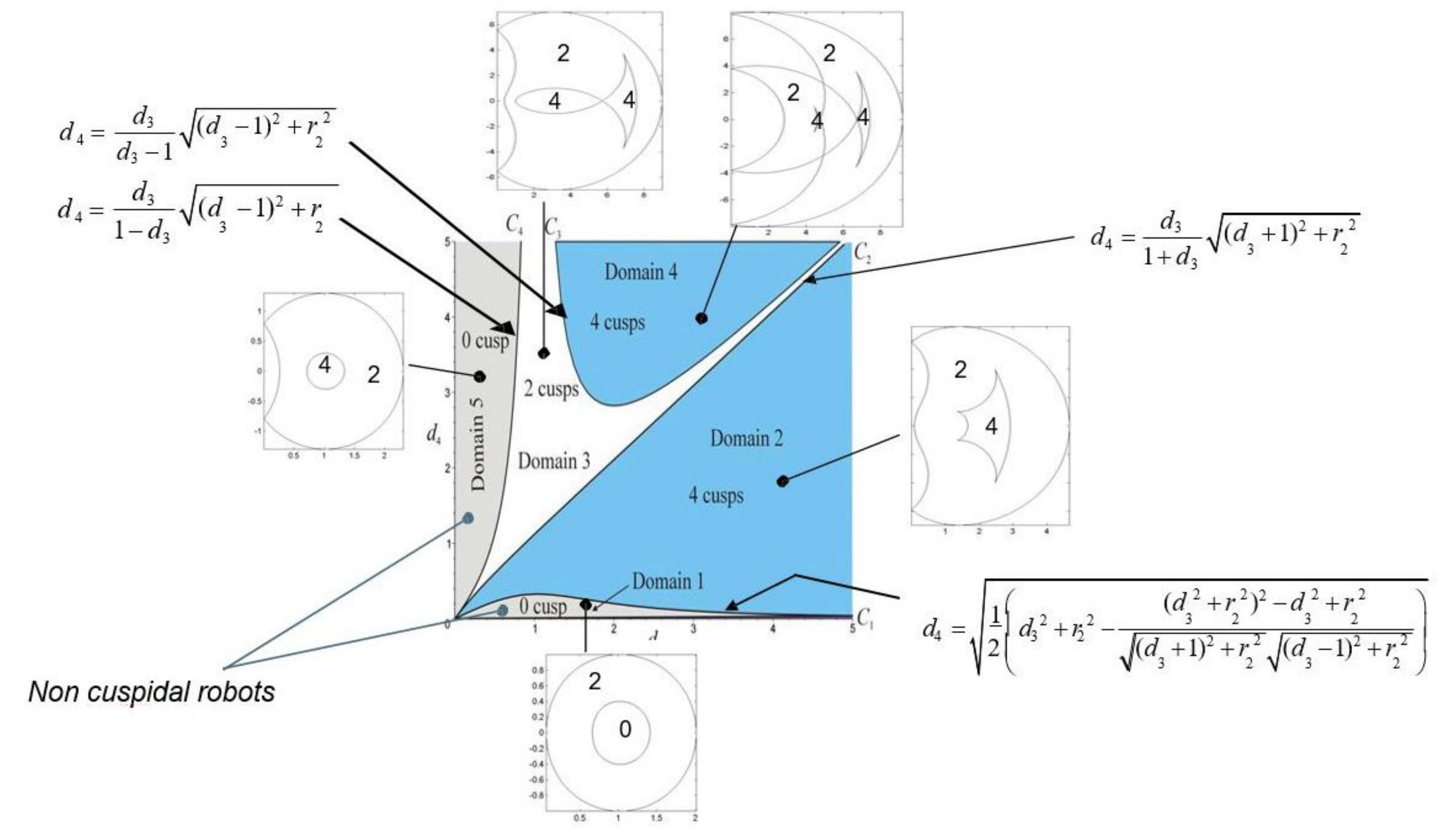}
	\caption{Examples of workspaces in each domain of the parameter space for robots normalized by $d_2=1$ (section at $r_2=1$). The equation for each discriminant surface has been displayed. The number of accessible solutions in each region of the workspace is shown.}
	\label{fig:Robot_par_domaine}
\end{figure*}

These surfaces divide the parameter space into domains corresponding to robots with a constant number of cusps.
To know the number of cusps in each domain, we just have to choose any robot and to count its cusps. Figure \ref{fig:classification_3r} shows a section of the parameter space in $r_2=1$. We show that all the robots in the blue domains have four cusps, those in the white domain have two cusps and those in the grey domains have no cusps, so they are noncuspidal. Figure \ref{fig:Robot_par_domaine} shows the workspace of a robot in each domain. The number of solutions in each region of the workspace is shown. We can clearly see that those robots from domains 1 and 5 are noncuspidal robots and have an inner boundary without cusps. In domain 1, the inner boundary defines a "hole" in the workspace, while in domain 5, the inner boundary encircles a region with four inverse solutions. In both cases, the number of solutions around the central region is two. In fact, domain 1 collects the set of robots having only two solutions at most, while in all other domains, robots have up to four solutions. 
\subsection{Necessary and sufficient cuspidality condition for orthogonal 3R robots}
The classification of the parameter space allows us to identify all cuspidal and noncuspidal 3R robots. Using the equations of the discriminant surfaces bounding domains 1 and 5 and relaxing the $d_2=1$ normalization, we can write that a orthogonal 3R robot such that $r_3=0$ is \emph{noncuspidal} if and only if:

\begin{equation}
\left\lbrace\begin{array}{l}
{d_4} < \sqrt {\frac{1}{2}\left( {d_3^2 + r_2^2 - \frac{{{{\left( {d_3^2 + r_2^2} \right)}^2} - d_2^2\left( {d_3^2 - r_2^2} \right)}}{{\sqrt {{{\left( {{d_3} + {d_2}} \right)}^2} + r_2^2} \sqrt {{{\left( {{d_3} - {d_2}} \right)}^2} + r_2^2} }}} \right)}\\
\text{or}\\
{d_3} < {d_2}~\text{and}~{{d_4} > \frac{{{d_3}}}{{{d_2} - {d_3}}} \sqrt {{{\left( {{d_3} - {d_2}} \right)}^2} + r_2^2} } 
\end{array} \right.
\label{CNS}
\end{equation}

\begin{figure}
	\centering
	\includegraphics[width=0.8\linewidth]{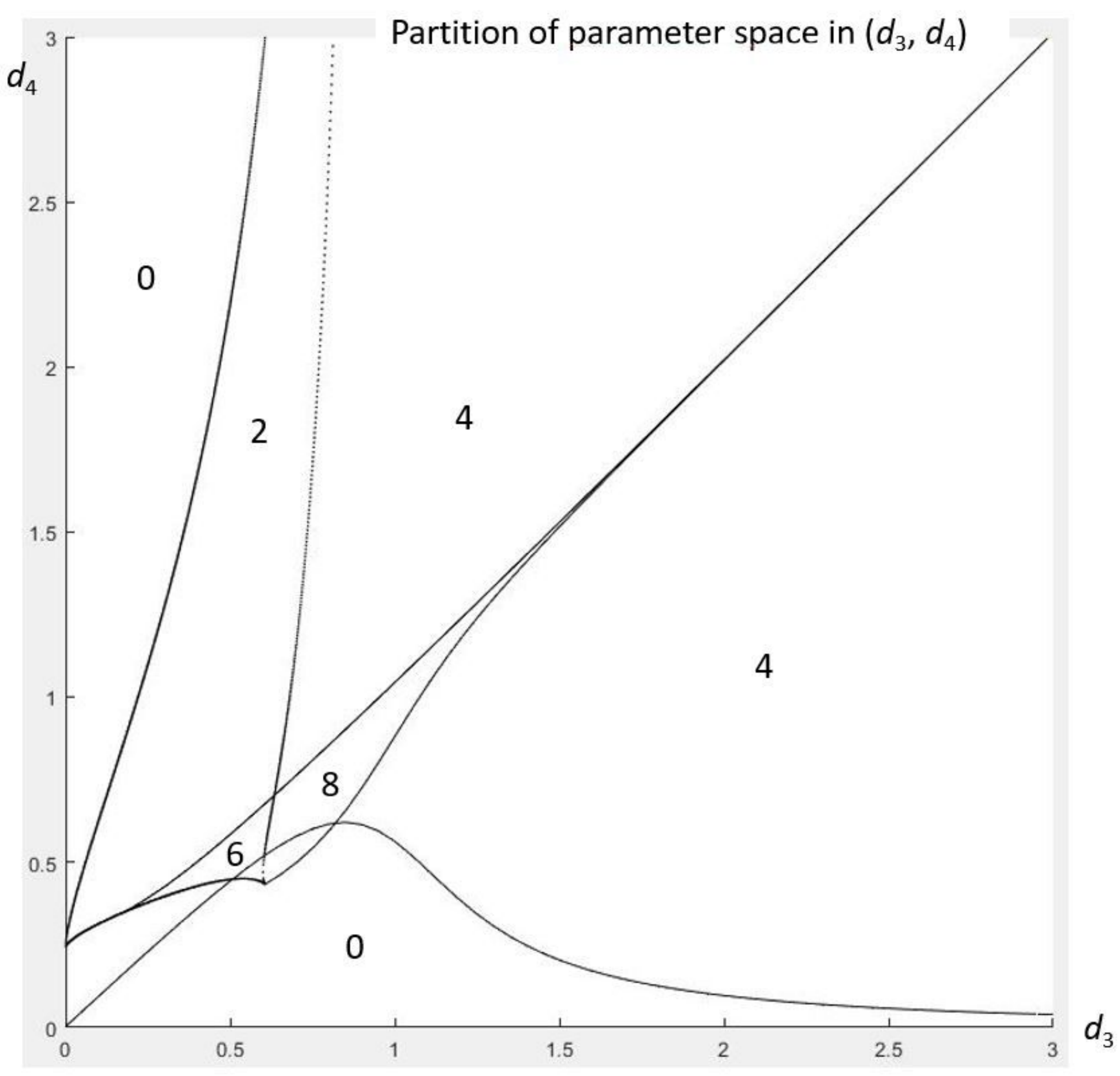}
	\caption{Partition of the parameter space as a function of the number of cusps for orthogonal 3R robots when $r_3\neq 0$ (section at $r_2=0.3, r_3=0.8$). Number of cusps is indicated in each domain.}
	\label{fig:classification_3r_r3}
\end{figure}

This condition is very useful for the designer who wants to implement an orthogonal robot that is not cuspidal. Note that the noncuspidal robots in domain 5, i.e. satisfying the second condition of \eqref{CNS}, are more interesting because they have a 4-solution region in their workspace, while those in domain 1 have a hole and no 4-solution region. Some rules for designing noncuspidal robots are proposed in section \ref{Rules}.
\\
When $r_3\neq0$, the parameter space is of dimension 4. Proceeding in the same way as before, we obtain discriminant hyper-surfaces of very complex equation \cite{Baili2004analyse}. One of them is of degree 12 and contains not less than 536 terms. Moreover, while the robots have only four cusps at most when $r_3=0$, there can be up to eight cusps when $r_3\neq0$. Figure \ref{fig:classification_3r_r3} shows a section of the parameter space in $r_2=0.3, r_3=0.8$. The number of cusp points is shown in each domain.

\subsection{Case of robots with 6 joints}
\subsubsection{Wrist-partitioned robots}
\begin{figure}
	\centering
	\includegraphics[width=0.7\linewidth]{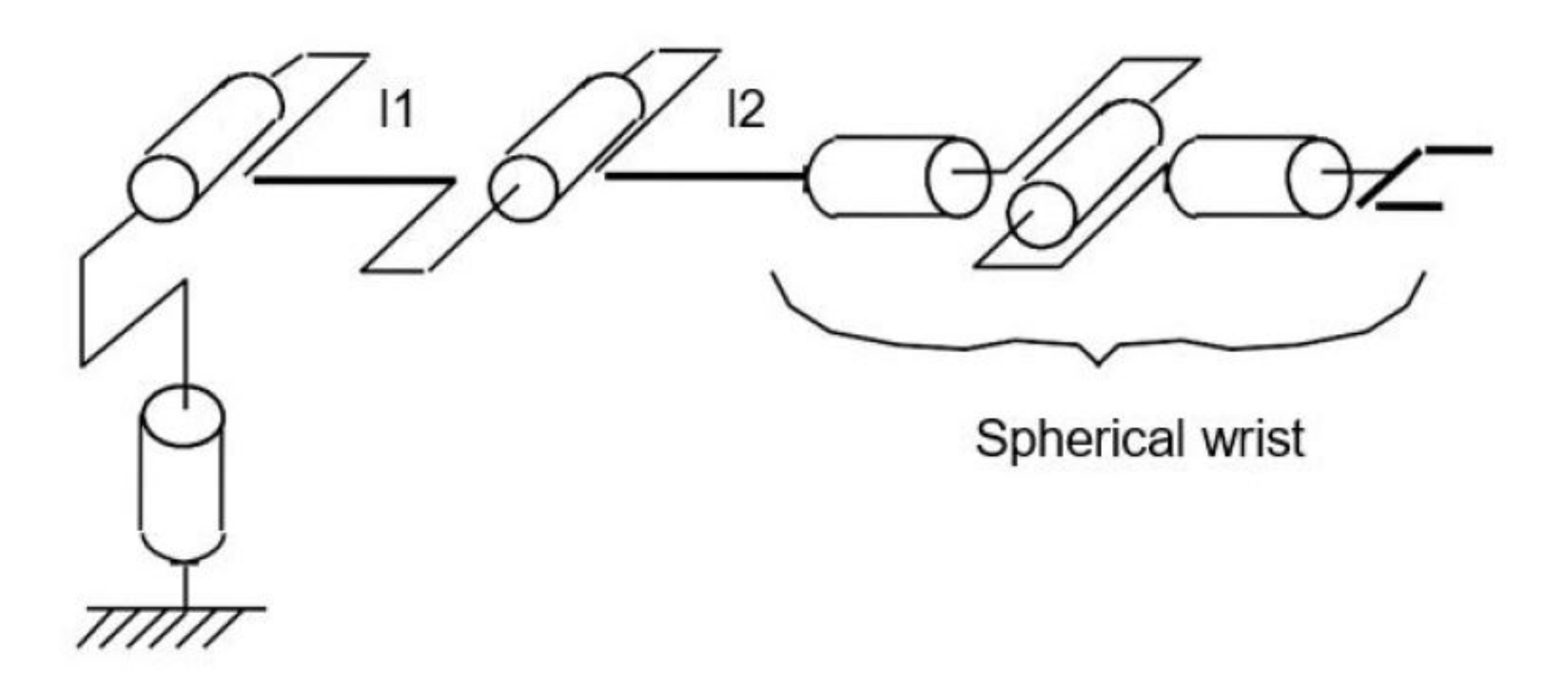}
	\caption{Scheme of a 6R anthropomorphic manipulator with a spherical wrist.}
	\label{fig:6R}
\end{figure}

Most industrial robots are designed by adding a spherical wrist in serie to a 3-axis robot. A spherical wrist is made by putting in series three revolute joints with concurrent and usually orthogonal axes, two by two, which allows making an actuated spherical joint. Such robots are called "wrist-partitioned" because their inverse kinematics and singularities are decoupled in terms of position/orientation. Figure \ref{fig:6R} shows an anthropomorphic robot similar to the one in Fig.  \ref{fig:PosturesNonCusp}, consisting of a 3R robot with a spherical wrist in series.

This design strategy allows simplifying the kinematic models of the robot as well as its singularities by decoupling the position equations from the orientation equations \cite{pieper1968kinematics}. This decoupling also allows decoupling the cuspidality analysis of the robot. Knowing that the spherical wrist is a simple mechanism which is noncuspidal (all its axes intersect, it thus enters the conditions of noncuspidality stated in section \ref{sec:CondSimples}), a 6R robot with a spherical wrist will be cuspidal if and only if its regional part made of the first three axes is cuspidal. All the classification and analysis results presented previously remain valid for 6R robots with a spherical wrist. 

\begin{figure}[ht]
	\centering
	\includegraphics[width=0.6\linewidth]{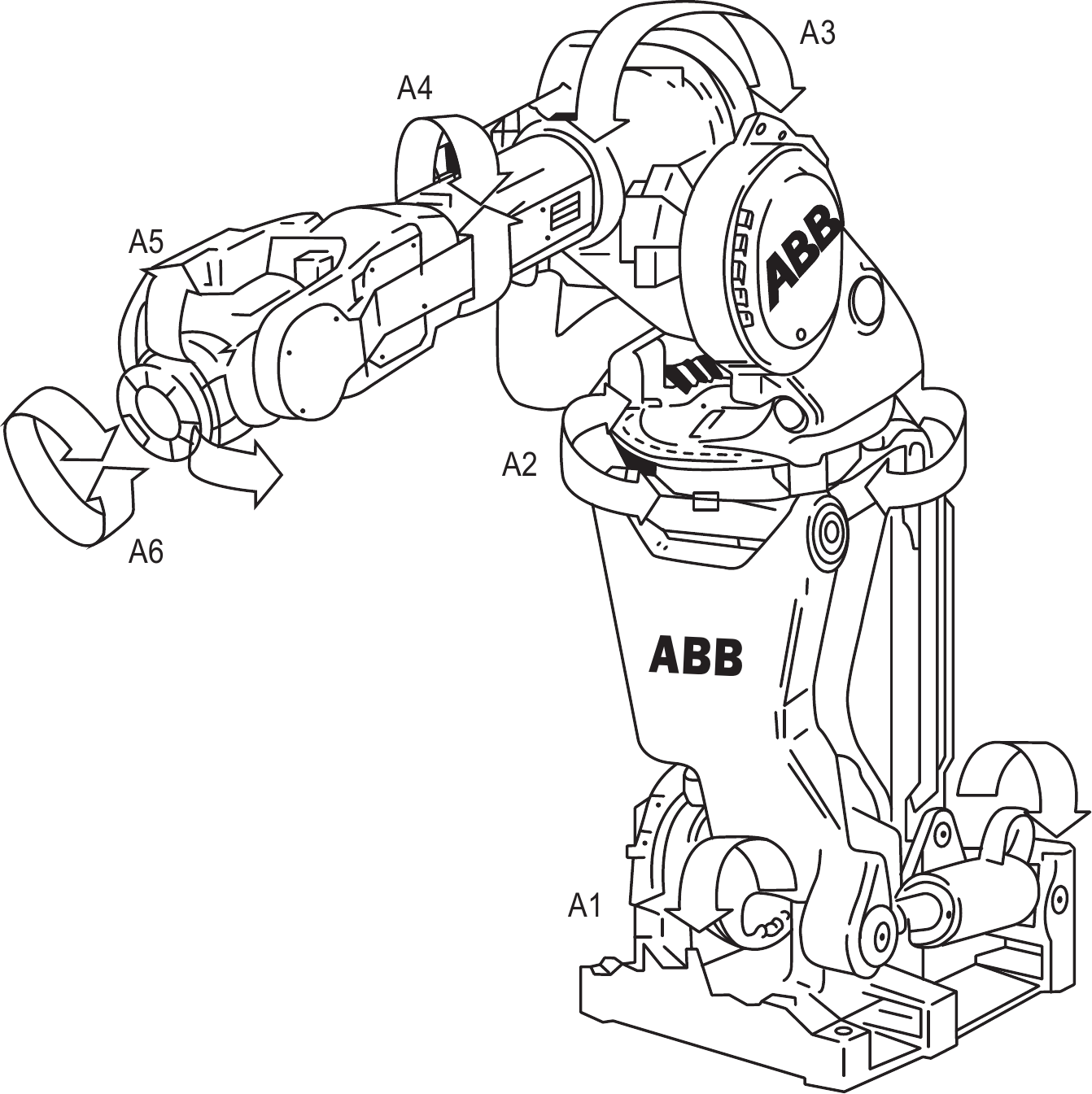}
	\caption{A 6R robot with spherical wrist that is cuspidal (ABB IRB 6400C).}
	\label{fig:IRB6400C}
\end{figure}

To our knowledge and at the time of writing this paper, there is no existing industrial robot with a spherical wrist that is cuspidal. However, a major robot manufacturer, ABB, marketed in 1996 a 6R robot with a spherical wrist that proved to be cuspidal. It was the IRB 6400C robot (Fig. \ref{fig:IRB6400C}). 
Its particularity lies in the permutation of the first two axes with respect to an anthropomorphic robot, making all its axes mutually orthogonal: we were thus dealing with an orthogonal 3R robot such as those studied in this paper. The aim of the designer was to limit the swept volume in order to be able to place robots closer together along the vehicle assembly lines. At the time of its release, the results on the classification of orthogonal robots were not yet published. This robot was removed from the catalog because users were facing difficulty when planning trajectories. In fact, this robot turned out to be cuspidal. The designer, had he known about the classification, could have easily set the geometric parameters so that the 3R part of this robot would fall into domain 5 of the 4-solution noncuspidal orthogonal 3R robots \cite{wenger_2007}.  
\subsubsection{Non-wrist-partitioned robots}
\begin{figure}
\centering
\begin{minipage}[t]{0.6\linewidth}
\includegraphics[width=\textwidth]{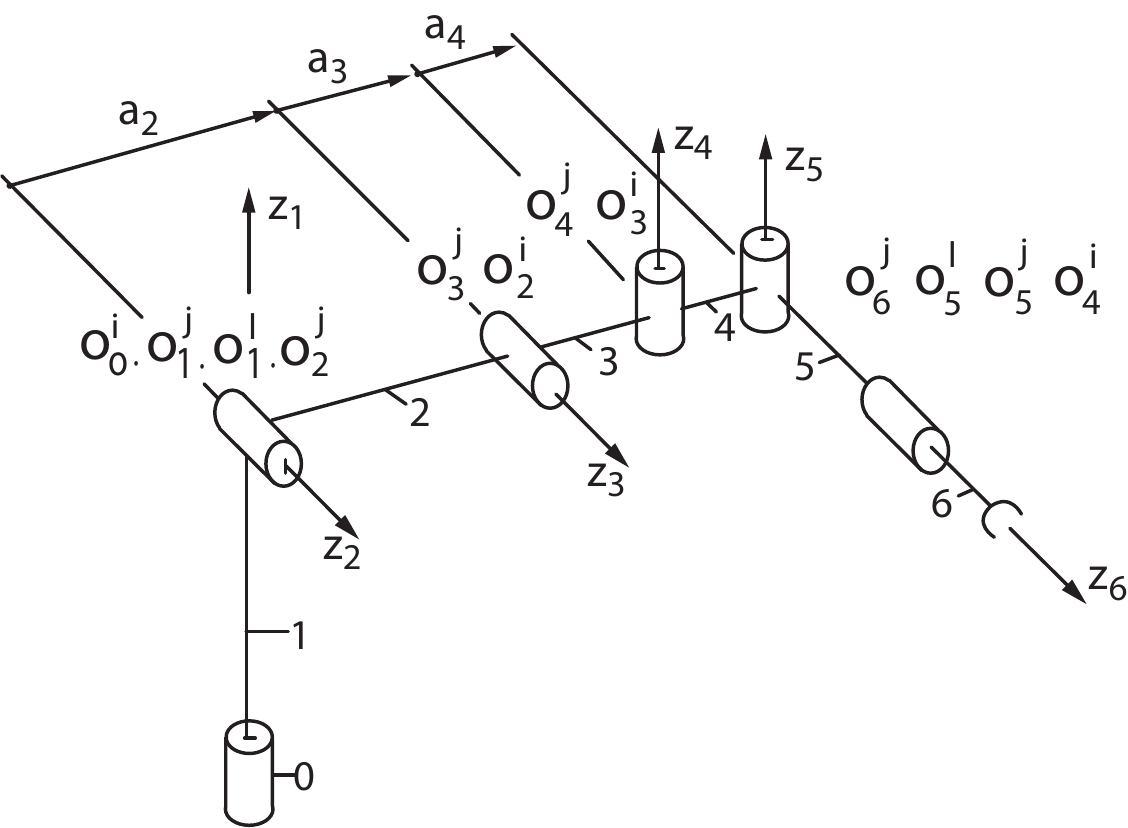}
\end{minipage}
\begin{minipage}[t]{0.6\linewidth}
\includegraphics[width=\textwidth]{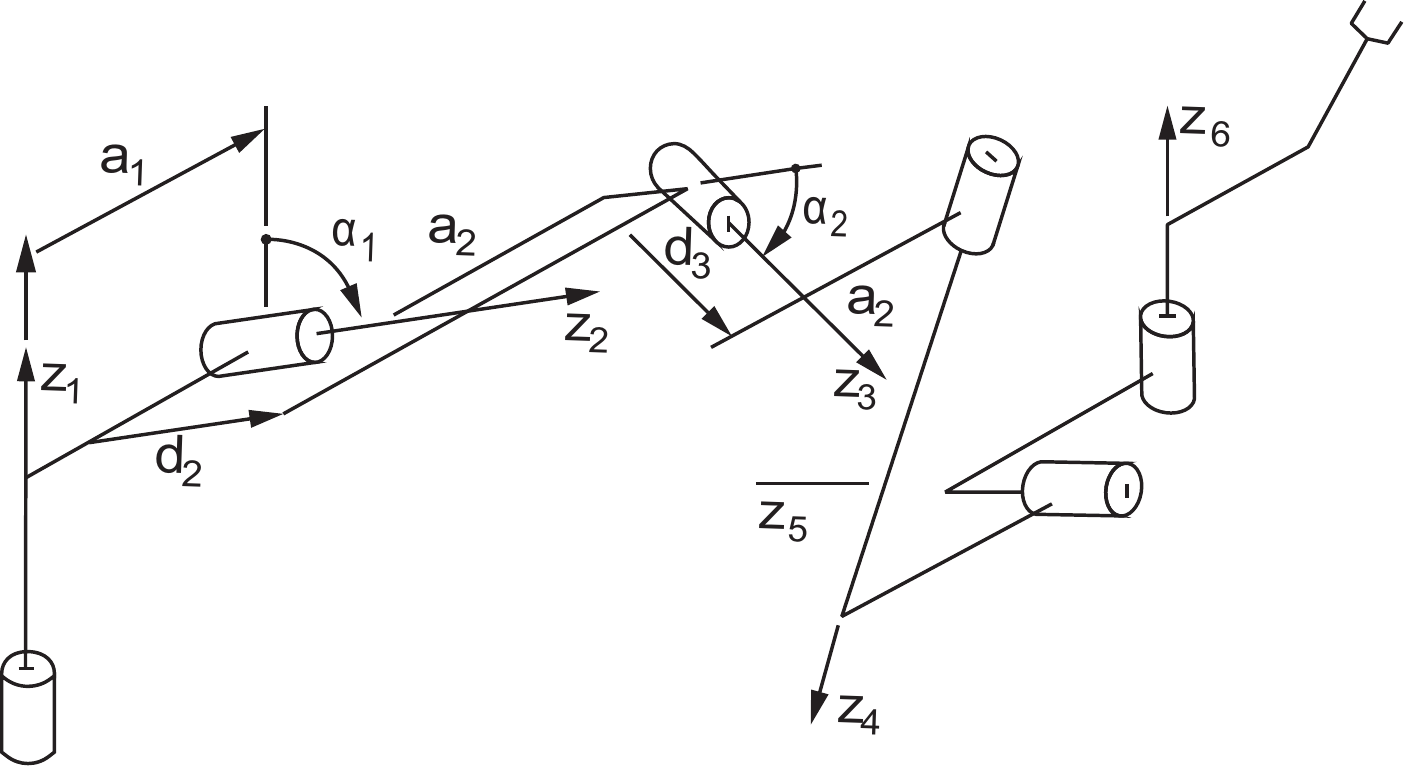}
\end{minipage}
\caption{The two cuspidal 6R robots studied by Innocenti (from \cite{innocenti1988singularity}).}
\label{fig:robotInnocenti}
\end{figure}

There are very few results on 6R robots that are not wrist-partitioned, because they are much more difficult to analyze. Indeed, the lack of decoupling between positions and orientations makes the resolution of the inverse kinematics complex (see for instance \cite{Pfurner} for the proposal of an algorithm based on Study's formalism). Moreover, the determinant of the Jacobian matrix usually does not factor and depends on more than three joint variables, which makes it difficult to identify and count aspects. There are, however, a few examples of 6R robots without a spherical wrist that could be analyzed:

\begin{itemize}
\item{-} 6R robots with three parallel axes: this feature allows a decoupling similar to the robots with a spherical wrist \cite{pieper1968kinematics}, \cite{pfurner2009explicit}. Like anthropomorphic robots, these robots are noncuspidal, have eight aspects and admit up to eight \vmc{solutions} \cite{Capco}. Examples are the UR5 and UR10 industrial robots from the company ``Universal Robots'' (Fig. \ref{fig:UR10Titan}, left) or, much older, the Schilling Titan II robot with hydraulic actuation and generally used as a remote manipulator arm (Fig. \ref{fig:UR10Titan}, right); 
\item{-} Anthropomorphic robots with wrist offset: these are 6R robots derived from the family of anthropomorphic robots, for which an offset has been introduced along axis 5, i.e. the second axis of the wrist. This offset makes the three axes of the wrist no longer intersect. It aims to eliminate the wrist singularity, which is particularly annoying for process applications. Some industrial painting robots have this feature, such as the GMF P150 robot (Fig. \ref{fig:P250}, top left) or, more recently, the Fanuc P250iB robot (Fig. \ref{fig:P250}, top right). Light-weight robots have then been designed with a wrist offset. Examples are the Kinova Jaco robot launched in 2009 \cite{Jaco} and, more recently, the Fanuc CRX-10iA robot (Fig. \ref{fig:P250}, bottom left) and the Yaskawa Motoman HC10 (Fig. \ref{fig:P250}, bottom right) . 

\begin{figure}
    \centering
    \begin{minipage}[t]{0.5\linewidth}
    \includegraphics[width=\textwidth]{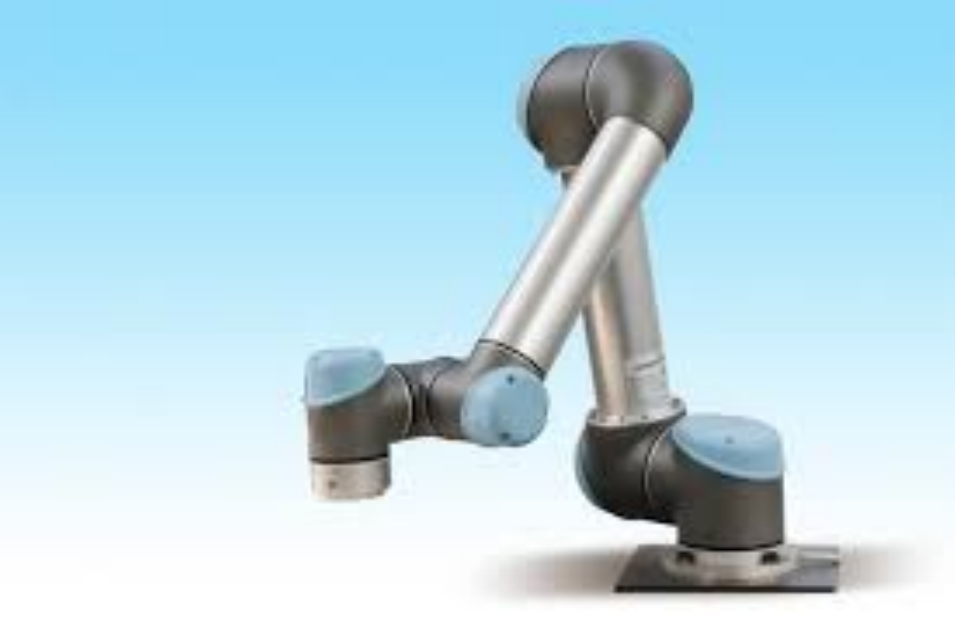}
    \end{minipage}
    \begin{minipage}[t]{0.3\linewidth}
    \includegraphics[width=\textwidth]{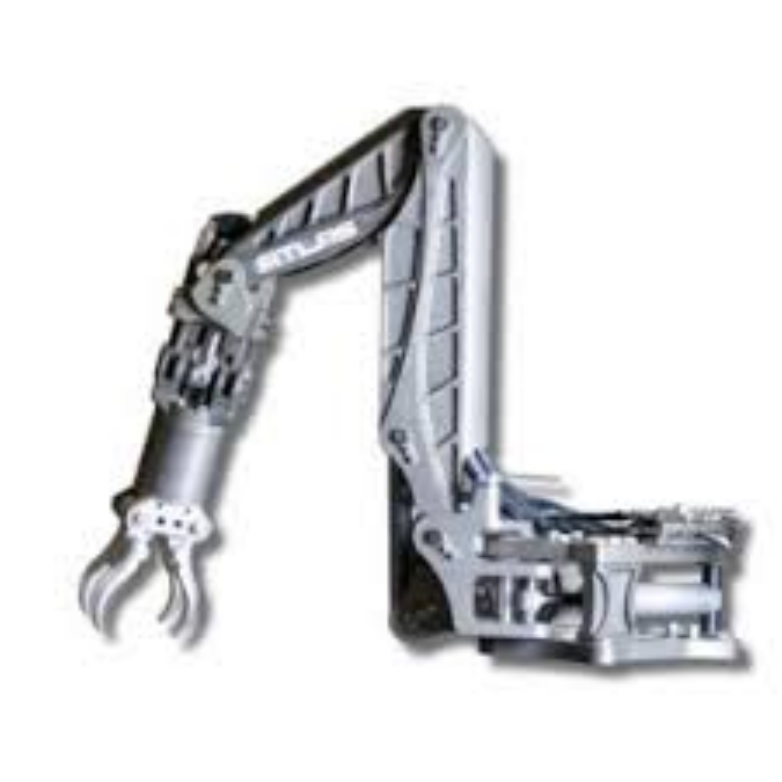}
    \end{minipage}
    \caption{Two 6R industrial robots with 3 parallel axes: the UR10 robot from Universal Robots (left) and the Schilling Titan II robot (right).}
    \label{fig:UR10Titan}
\end{figure}

\begin{figure}
\centering
\begin{minipage}[t]{0.5\linewidth}
\includegraphics[width=0.6\textwidth]{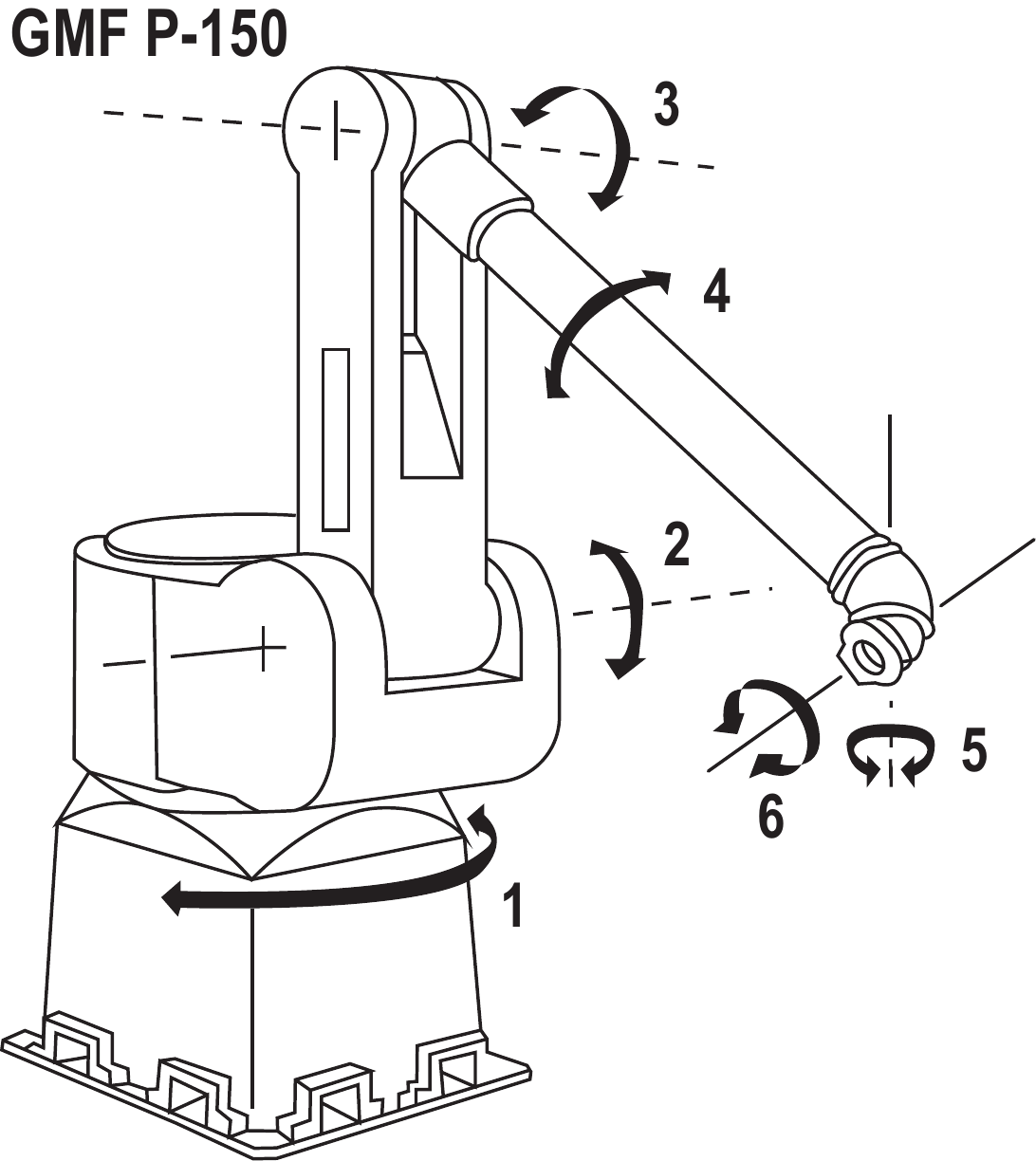}
\end{minipage}
\begin{minipage}[t]{0.4\linewidth}
\includegraphics[width=0.8\textwidth]{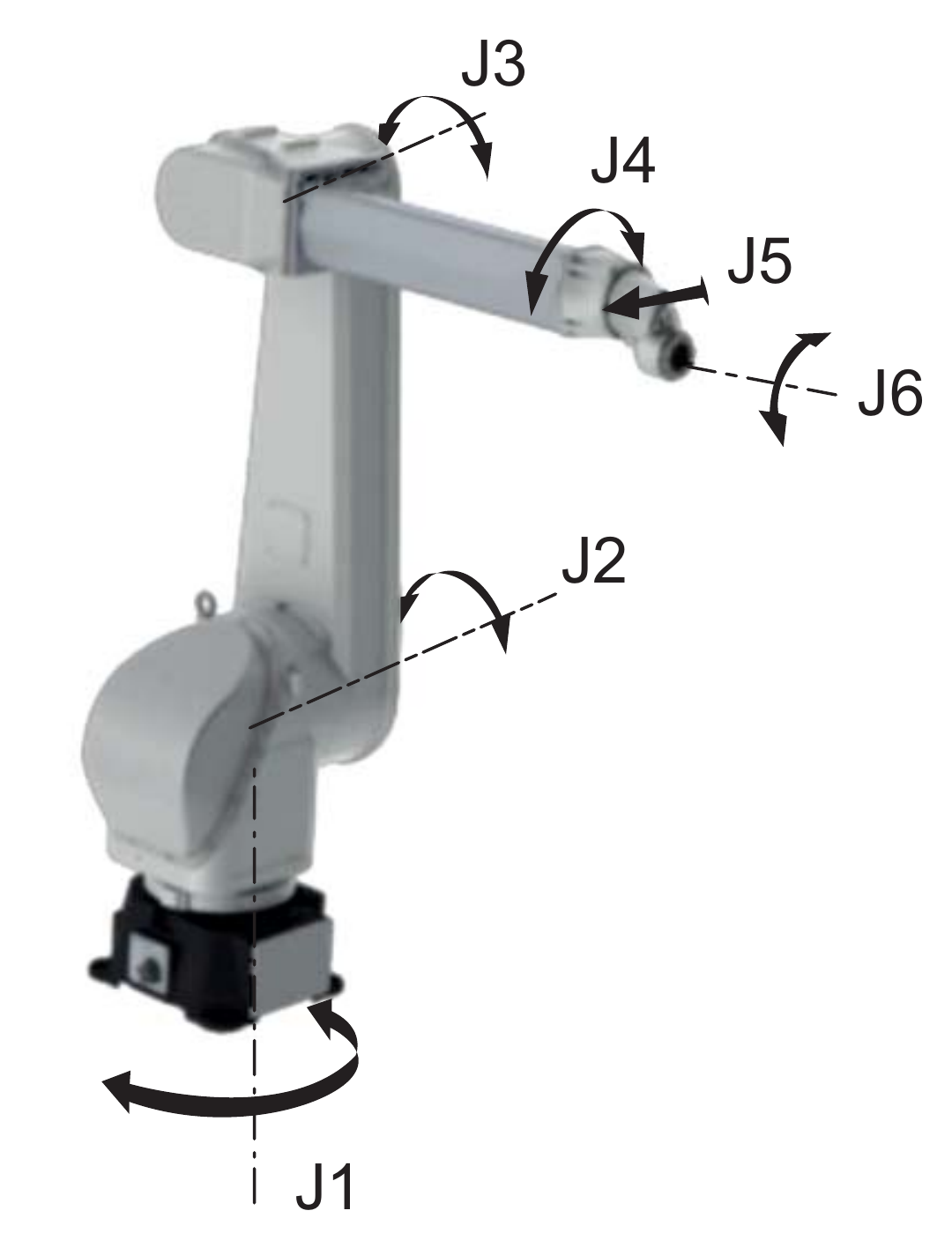}
\end{minipage}
\begin{minipage}[t]{0.5\linewidth}
\includegraphics[width=0.9\textwidth]{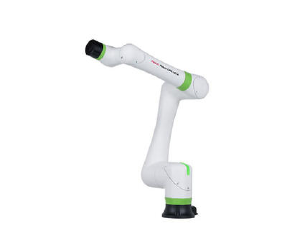}
\end{minipage}
\begin{minipage}[t]{0.4\linewidth}
\includegraphics[width=0.5\textwidth]{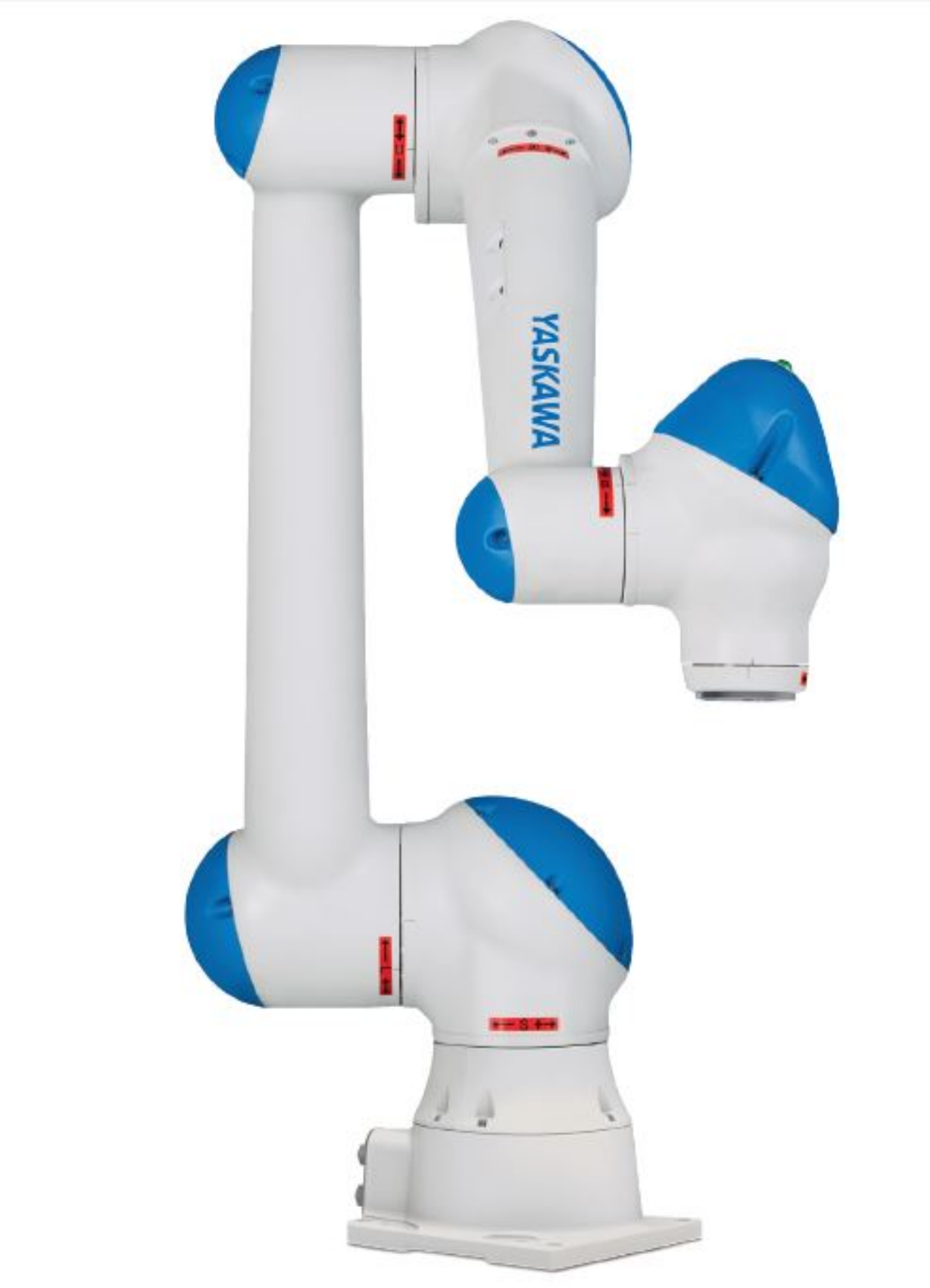}
\end{minipage}
\caption{Four 6R industrial robots with wrist offset: the GMF P150 robot (top), the Fanuc P250iB robot (top, right), the Fanuc CRX-10iA (bottom, left) and the Yaskawa Motoman HC10 (bottom, right).}
\label{fig:P250}
\end{figure}

The GMF P150 can be shown to be cuspidal in theory \cite{Wenger97}. Nevertheless, the presence of strong joint limits reduces the number of \vmc{solutions} of this robot to 2, always separated by a singularity. The robot is therefore, in practice, noncuspidal. The Fanuc P250iB robot has not yet been analyzed but, with an architecture similar to the GMF P150 robot, it is likely to be cuspidal without its joint limits. The Jaco and the CRX-10iA collaborative robots, however, are really cuspidal, because their joint ranges are very large \cite{Durgesh}. The Motoman HC10 has not been studied but, since it is similar to the Fanuc CRX-10iA, it should be cuspidal too. Difficulties in planning trajectories with these robots have been recently reported \cite{Achille}. 
\item{-} Innocenti's robots~: two imaginary cuspidal 6R robots were analyzed in \cite{innocenti1988singularity} to demonstrate the existence of cuspidal robots. One of these robots has a very simple geometry~: there are three pairs of orthogonal axes~: (1,2), (3,4), (5,6) and two pairs of parallel axes~: (2,3) and (4,5) (see Fig. \ref{fig:robotInnocenti} at left). The other robot has an arbitrary architecture (\ref{fig:robotInnocenti}, right). For each of these robots, the authors have numerically found a non-singular trajectory to switch from one \vmc{solution} to the other. 
\end{itemize}

\subsection{Rules for designing a robot}
\label{Rules}
We have seen that a cuspidal robot is more difficult to analyze, especially for the feasibility study of trajectories. Moreover, the solutions of a cuspidal robot cannot be easily identified or recognized. Indeed, the determinant of the Jacobian matrix of a cuspidal robot cannot be factored in general. This means that the solutions cannot be identified with the sign of the factors. Moreover, there is no clear geometric interpretation of the solutions (e.g. elbow up/down).

It is therefore more reasonable to try to design a noncuspidal robot. To do this, we can use the conditions stated in section \ref{sec:CondSimples}, and add a spherical wrist if we want to design a 6R robot. Applying these rules, we find most of the architectures of common industrial robots, except condition 6 of section \ref{sec:CondSimples}. This condition produces orthogonal robots, non-existent on the current market. 
This class of robots is however an interesting alternative to anthropomorphic robots. Indeed, upon choosing the geometric parameters conveniently, it was shown that a compact and regular workspace can be obtained \cite{zein2007design}, with better dynamic performance than an anthropomorphic robot \cite{Nguyen}.

\section{Cuspidality in parallel robots}
The previous sections have been devoted to the study of serial robots. The study of parallel robots, which have a much more complex architecture than serial robots, requires to extend several definitions to this class of robots. Moreover, the determination of the direct kinematic model and singularities of parallel robots is much more delicate than for their serial counterparts. Unlike 3R serial robots, providing a complete classification for a family of parallel robots is still intractable. In this section, we present some partial classification results for four families of parallel robots~: R\underline{P}R\underline{P}R robots, 3--R\underline{P}R robots, 3--\underline{PP}PS robots, and 2-U\underline{P}S--U spherical robots. Symbols U and S denote a universal joint and a spherical joint, respectively while R and P denote a revolute and prismatic joint, respectively. An underlined letter means that the joint is actuated.
\subsection{Extension of the notion of aspect for parallel robots}

The notion of aspects was extended to parallel robots with a single solution to their inverse kinematics in \cite{wenger1997definition} and then to parallel robots with several inverse kinematic solutions in \cite{chablat1998working}. In this paper, only parallel robots with only one inverse kinematic solution are analyzed. We recall below the extended definition of aspects for these parallel robots. 

The aspects ${\cal WA}_j$ of a parallel robot are defined as the largest domains of the workspace $\cal W$ such that~:
	\begin{itemize}
        \item{-} ${\cal WA}_j \in \cal W$;
        \item{-} ${\cal WA}_j$ is path-connected;
        \item{-} $\forall X \in \cal W$,  $\det(\bf{A}) \neq 0$,\\
        where ${\bf A}$ is the parallel Jacobian matrix of the robot, namely, the one associated to the inverse kinematic map ${\bf g}$ defined by $\bf q={\bf g}({\bf X})$.
	\end{itemize}

In other words, the aspects are the largest domains of the workspace that are free of parallel singularities. The set of aspects $\{{\cal WA}_j\}$ can be obtained upon subtracting the singular configurations $\cal S$ from the workspace and determining the path-connected components of the resulted singularity-free workspace: $\{ {\cal WA}_j\}= {\rm CC \{ {\cal W} \dot{-} \cal{S}}\}$ where CC = connected components and $\dot{-}$ denotes the difference between sets.

\begin{figure}
	\centering
	\includegraphics[width=0.5\linewidth]{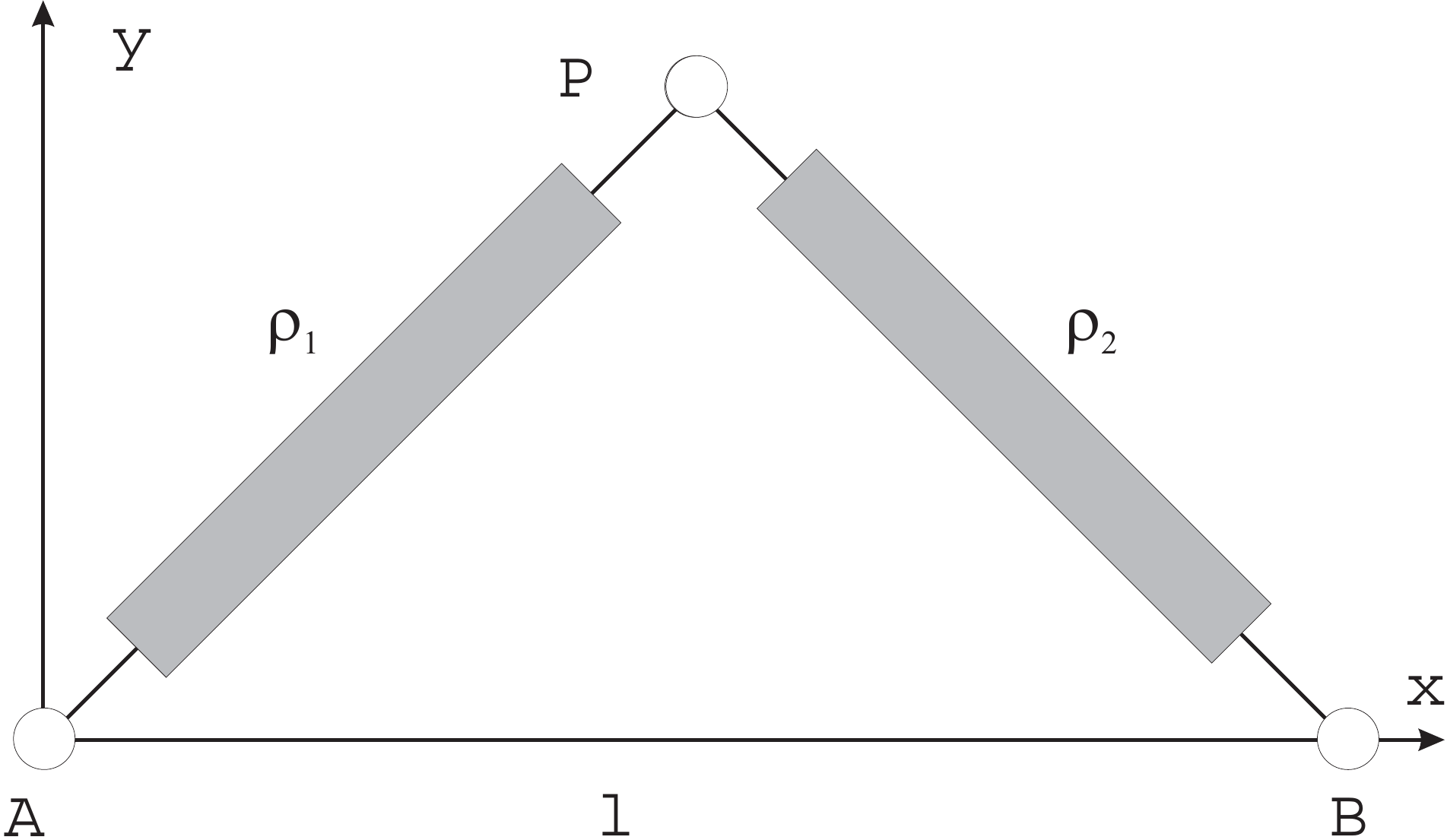}
	\caption{A planar parallel robot R\underline{P}R\underline{P}R}
	\label{fig:2-rpr}
	\includegraphics[width=0.5\linewidth]{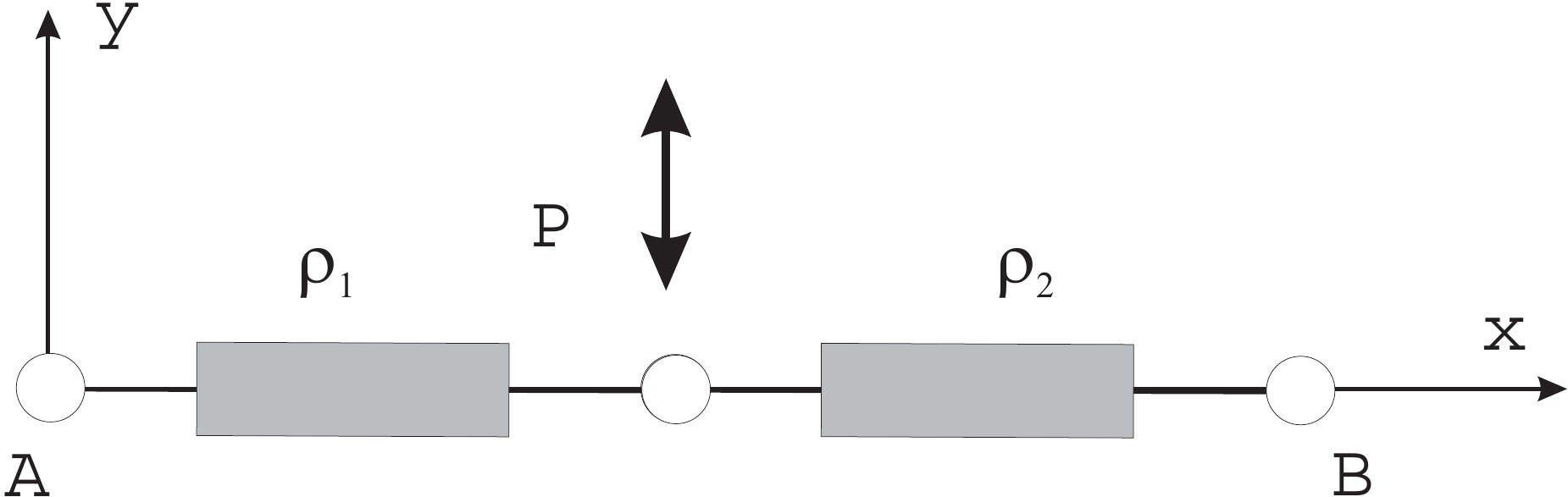}
	\caption{Singularity of the planar parallel robot R\underline{P}R\underline{P}R.}
	\label{fig:2-rpr-singularite}
\end{figure}

Figure \ref{fig:2-rpr} shows a planar parallel robot with two degrees of freedom, of type R\underline{P}R\underline{P}R.	Its input variables are the lengths of the prismatic joints $\rho_1$ and $\rho_2$, and its output variables are the two position coordinates $(x,y)$ of point $P$. Upon writing that point $P$ lies at the intersection of two circles, one centered at $A$ and of radius $\rho_1$, the other centered at $B$ and of radius $\rho_2$, we obtain the following two constraint equations:
	\begin{eqnarray}
	\rho_1^2&=& x^2+y^2 \nonumber \\
	\rho_2^2&=& (x - l)^2 + y^2 \nonumber
	\end{eqnarray}
    where $l$ denotes the distance between the two pivots $A$ and $B$. Moreover, the joint limits produce two additional constraints given by the inequalities $\rho_{i~min} \leq \rho_i \leq \rho_{i~max}$ with $i$=$(1,2)$.
    
\begin{figure}
    \centering
	\includegraphics[width=0.4\linewidth]{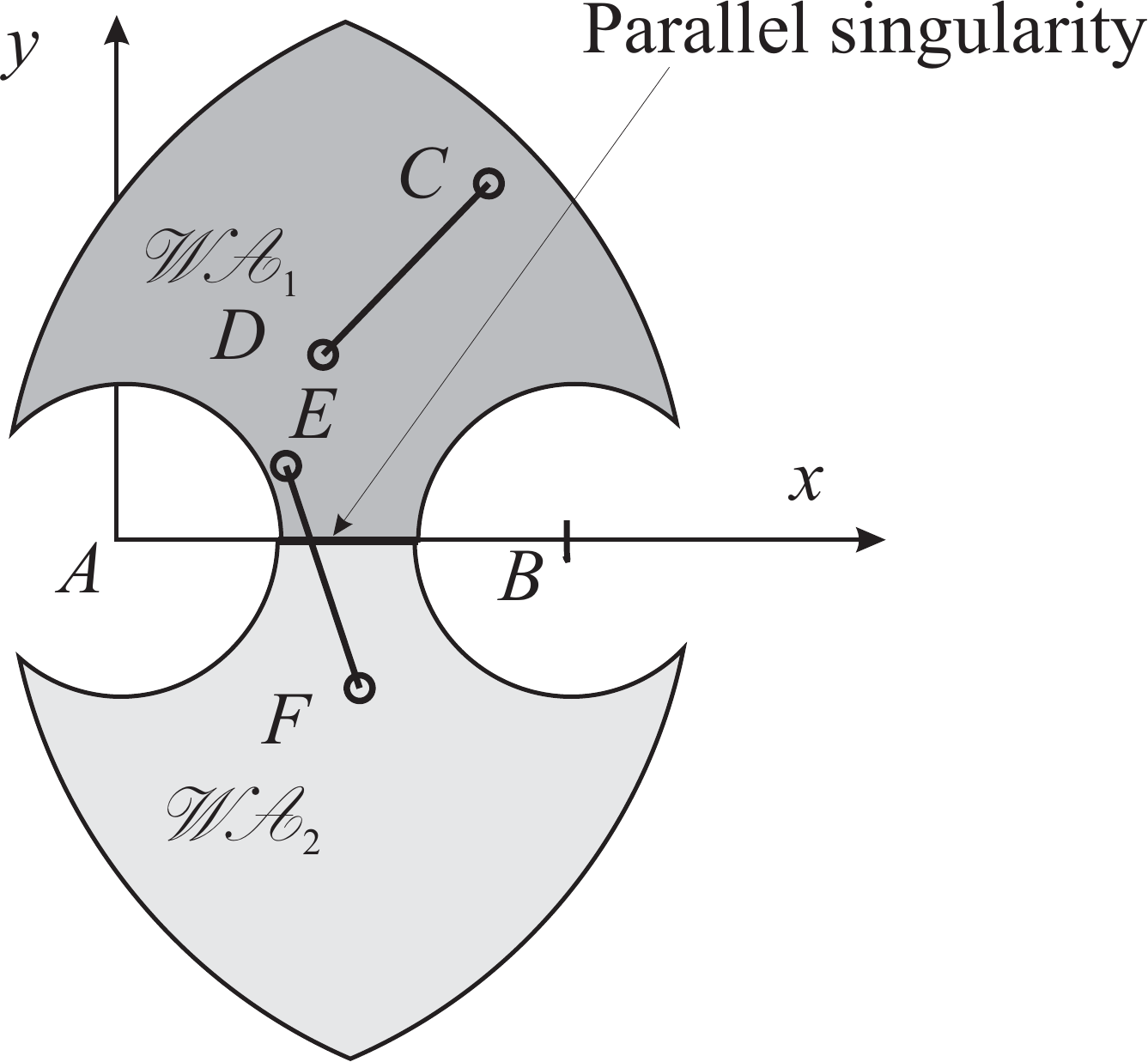}
	\caption{The two aspects of the planar parallel robot R\underline{P}R\underline{P}R when $l=5$, $\rho_{min}=2$ and $\rho_{max}=7$.}
	\label{fig:2-rpr-aspect}
\end{figure}

       Assuming that $\rho_{i~min}>0$ (which is usually the case for technological reasons), the robot has no serial singularity and has only one inverse kinematic solution.
    The parallel singularities of this robot are the configurations for which $A$, $B$ and $P$ are aligned (Fig. \ref{fig:2-rpr-singularite}).  Figure~\ref{fig:2-rpr-aspect} shows the robot workspace. We see that for the chosen dimensions, there are two aspects ${\cal WA}_1$ and ${\cal WA}_2$. Since this robot has only two assembly modes (point $P$ is above or below the $x$ axis), the aspects are uniqueness domains of the direct kinematics and thus this robot is not cuspidal.
   Note that, contrary to serial robots, the aspects of parallel robots divide the workspace and not the joint space.
    The $CD$ trajectory is feasible in the ${\cal WA}_1$ aspect while the $EF$ trajectory, straddling the two aspects ${\cal WA}_1$ and ${\cal WA}_2$, cannot be followed continuously without encountering a singular configuration.

\subsection{Cuspidal parallel robots}
\label{CuspPar}
A parallel robot is said cuspidal if it can change its assembly mode without going through a singularity.

\begin{figure}
    \centering
    \includegraphics[width=0.5\linewidth]{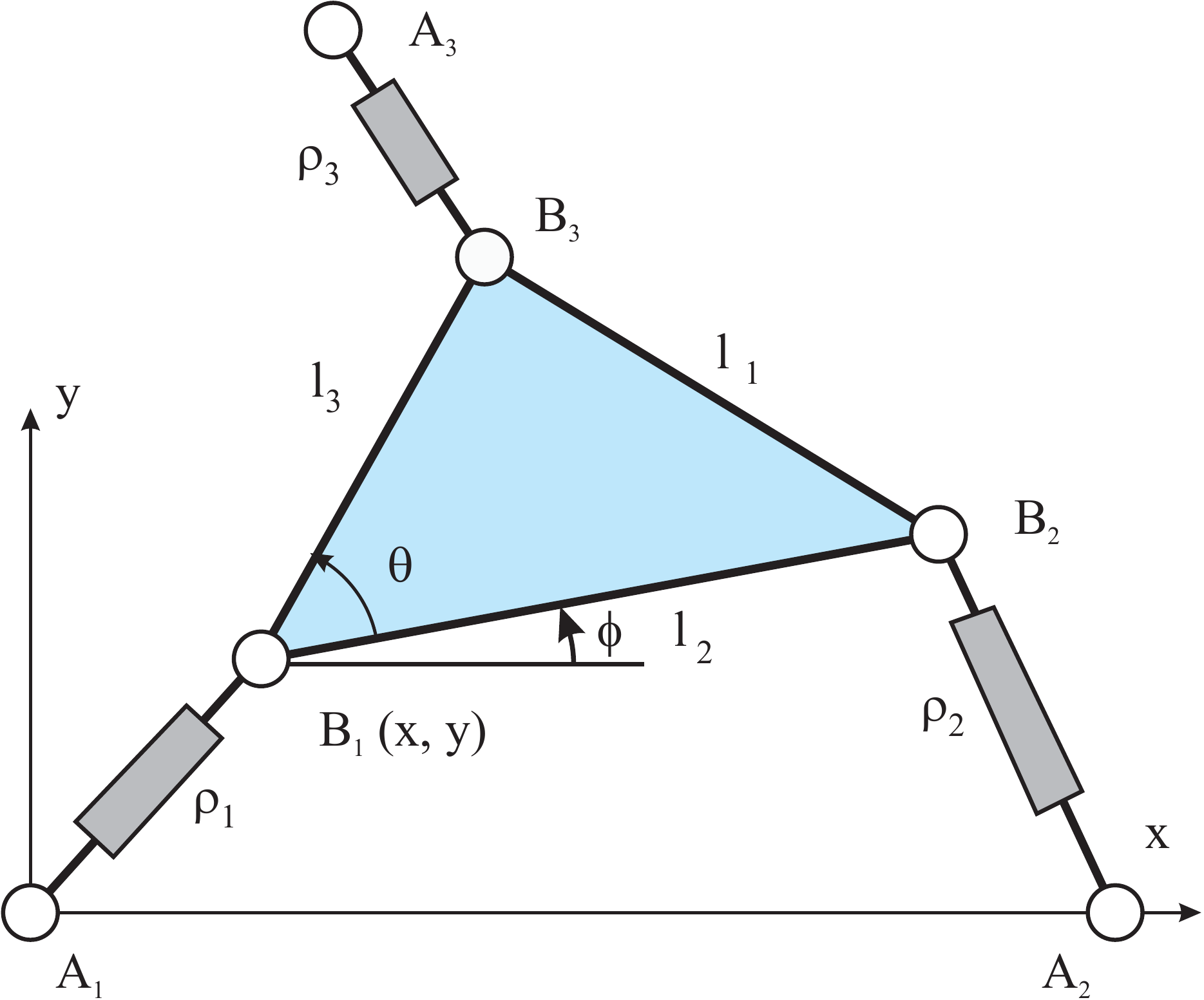}
    \caption{A 3--R\underline{P}RR parallel robot.}
    \label{fig:RPR}
\end{figure}

Figure \ref{fig:RPR} shows a 3--R\underline{P}R planar parallel robot with three degrees of freedom. Its input variables are the lengths of the prismatic joints $\rho_1$, $\rho_2$ and $\rho_3$, and its output variables are the two position coordinates $(x,y)$ of point $B_1$ and the orientation $\phi$ of the moving platform. The geometric parameters are the three sides of the moving platform $l_1$, $l_2$, $l_3$ and the position of the centers of the base revolute joints, defined by $A_1$, $A_2 $ and $A_3$. The reference frame is centered on $A_1$ and the $x$ axis passes through $A_2$. Thus, $A_1= (0, 0)$, $A_2 = (c_2, 0)$ and $A_3 = (c_3, d_3)$.

The constraint equations of this robot are given by \cite{clement1992solutions}~:
\begin{eqnarray}
\rho_1^2&=& x^2+y^2\nonumber\\
\rho_2^2&=& (x+l_2 \cos(\phi) - c_2)^2+(y+l_2 \sin(\phi))^2\nonumber\\
\rho_3^2&=& (x+l_3 \cos(\phi+\theta) - c_3)^2+ (y+l_3 \sin(\phi+\theta)-d_3)^2\nonumber
\end{eqnarray}
where $\theta$ is the angle of the moving triangle at $B_1$. We do not detail the calculation of the inverse kinematics which leads to the resolution of a polynomial of degree six, this one is described for example in \cite{merlet1997robots}.
For our study, we assume the same dimensions as in \cite{innocenti1998singularity} and \cite{merlet1997robots} (Table~\ref{tab:array_RPR}).

\begin{figure}
    \centering
    \includegraphics[width=0.35\linewidth]{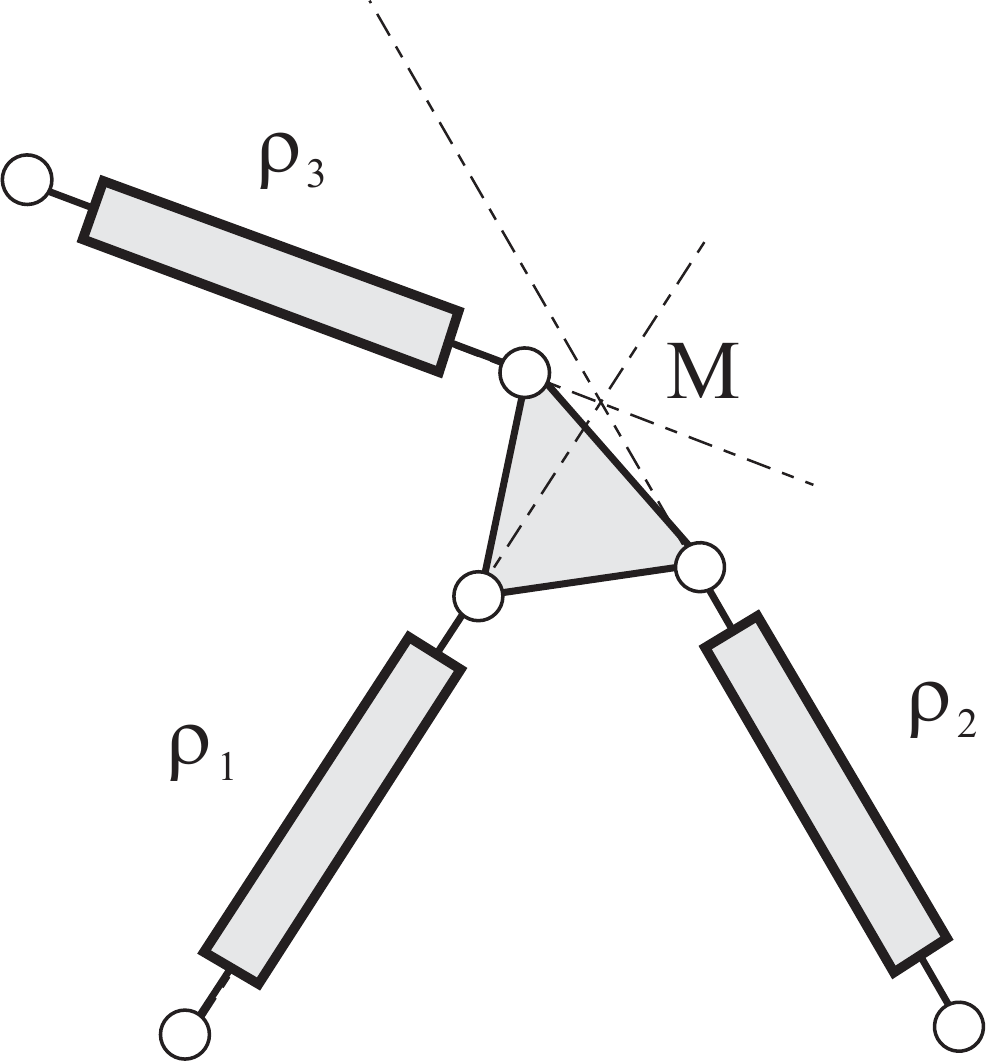}
    \caption{Parallel Singularity of the 3--R\underline{P}R parallel Robot.}
    \label{fig:singularity_RPR}
\end{figure}

\begin{table}[ht]
\centering
\caption{Dimensions of the 3--R\underline{P}R planar parallel robot studied (units do not matter).}
\begin{tabular}{|c|c|}
\hline
$A_1= (0, 0)$ & $l_1 = B_2B_3= 16.5$ \\
$A_2= (15.9, 0)$& $l_2 = B_1B_2= 17$ \\
$A_3= (0, 10)$& $l_3 = B_3B_1= 20.8$\\
\hline
\end{tabular}
\label{tab:array_RPR}
\end{table}
Assuming $\rho_i > 0, i=1, ...3$, the robot has no serial singularity. The parallel singularities of this robot are those configurations reached when the axes passing through the three prismatic joints intersect at a single point (Fig.~\ref{fig:singularity_RPR})~or are parallel \cite{gosselin1990singularity}.

The limits of the prismatic joints are those chosen by \cite{innocenti1998singularity}~:
\begin{equation}
10.0 \leq \rho_i \leq 32.0 \quad {\rm for} \quad i=(1, 2, 3)
\end{equation}
Units do not matter in this example. 

When $\rho_1 = 15.0$, $\rho_2 = 15.4$ and $\rho_3 = 12.0$, the robot admits six solutions to the direct kinematics, i.e. six assembly modes (Table~\ref{tab:table_6_MGD_3_RPR}).

\begin{figure}
 
\centering
\includegraphics[width=0.4\linewidth]{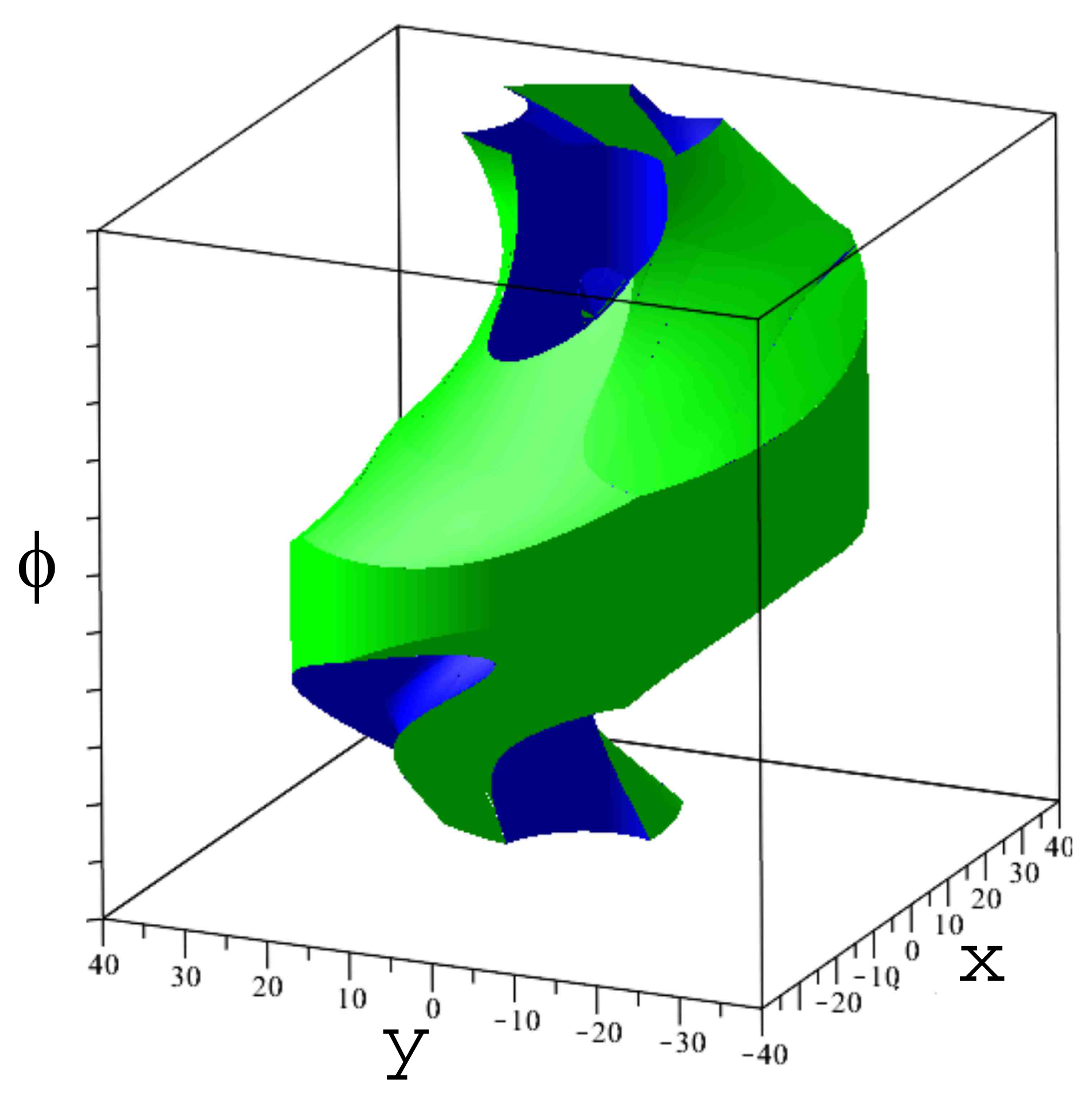}
\caption{Workspace of the 3--R\underline{P}R robot.}
\label{fig:3-RPR-workspace}
 
\end{figure}
\begin{table}[ht]
\centering
\caption{The six assembly modes of the 3--R\underline{P}R robot for $\rho_1 = 15.0$, $\rho_2 = 15.4$ and $\rho_3 = 12.0$ ($\phi$ given in radians).}
\begin{tabular}{|c|c|c|c|}
\hline
~ & $x$ & $y$ & $\phi$ \\
1 & -8.715 & 12.183 & -0.987 \\
2 & -5.495 & -13.935 & -0.047 \\
3 &-14.894 & 1.596 & 0.244 \\
4 & -13.417 & -6.660 & 0.585 \\
5 & 14.920 & -1.337 & 1.001 \\
6 & 14.673 & -3.013 & 2.133 \\
\hline
\end{tabular}
\label{tab:table_6_MGD_3_RPR}
\end{table}

\begin{figure}[ht]
    \centering
    \includegraphics[width=0.45\linewidth]{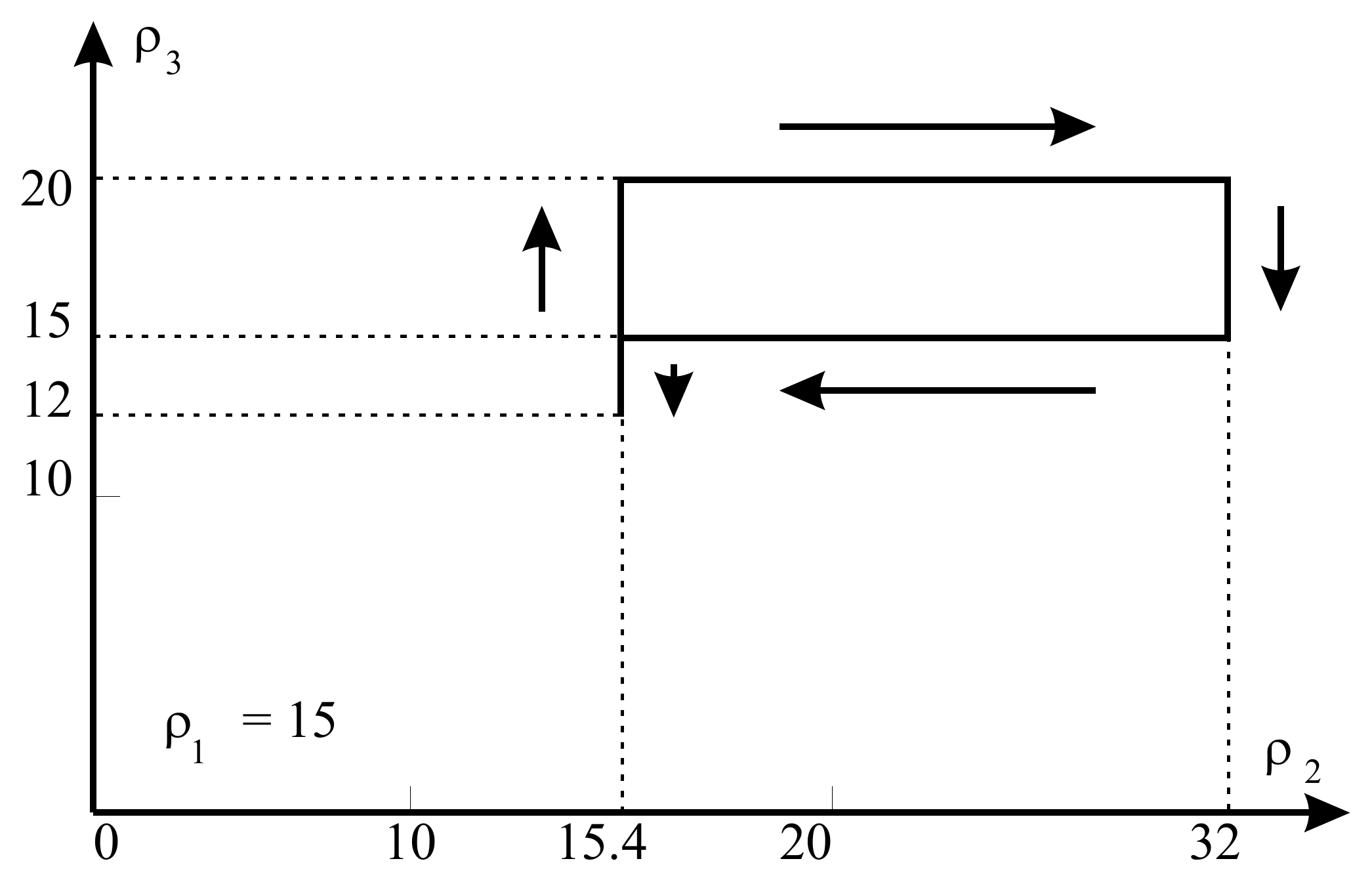}
    \caption{Example of assembly mode changing trajectory of the 3--R\underline{P}R parallel robot in joint space for $\rho_1=15$.}
    \label{fig:cycle_RPR}
    \centering
    \includegraphics[width=0.45\linewidth]{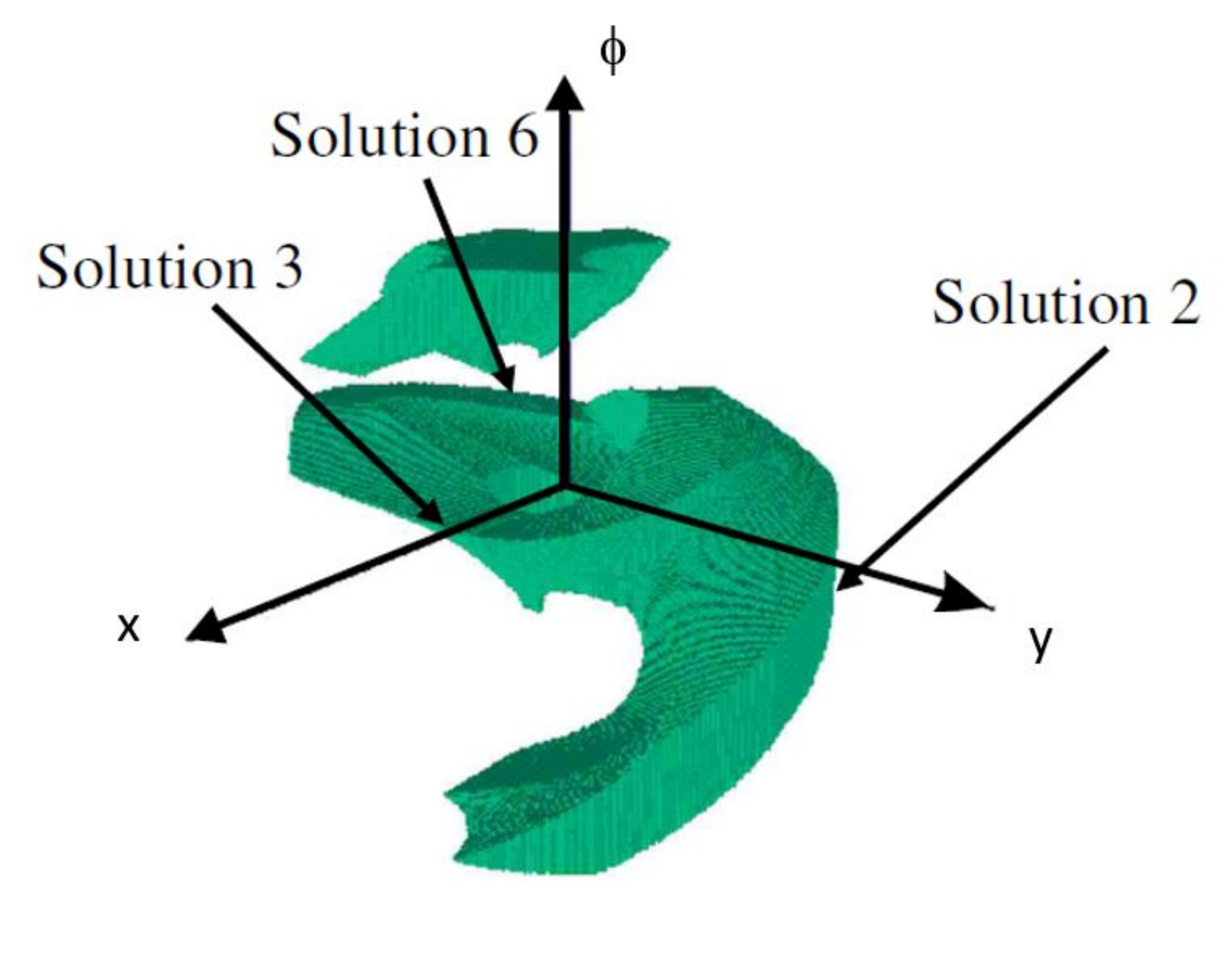}
    \caption{Assembly modes 2-3-6 in aspect ${\cal WA}_1$ of the 3--R\underline{P}R parallel robot in workspace.}
    \label{fig:solution_aspect_1}
\end{figure}

An assembly mode changing trajectory was proposed by \cite{innocenti1992direct}, thus showing that this robot is cuspidal. Figure~\ref{fig:cycle_RPR} shows this trajectory in the joint space. Its traces a rectangular loop starting and ending at the aforementioned joint configuration $\rho_1=15$, $\rho_2=15.4$ and $\rho_3=12$. 
We can show that the robot under study has two aspects \cite{zein2008non}. In fact, it was shown later that all generic 3--R\underline{P}R have always 2 aspects \cite{Manfred, SimpleProof}. Figure~\ref{fig:3-RPR-workspace} shows the workspace. The parallel singularities define a surface that splits the workspace into its two aspects. 
The assembly modes -2-, -3-, -6- of table~\ref{tab:table_6_MGD_3_RPR} are in the same aspect (${\cal WA}_1$) (Fig.~\ref{fig:solution_aspect_1}) and the assembly modes -1-, -4- and - 5- are in a second aspect (${\cal WA}_2$). This means that aspects are not uniqueness domains for the direct kinematics of this robot.

\begin{figure}
   \centering
   \includegraphics[width=10cm]{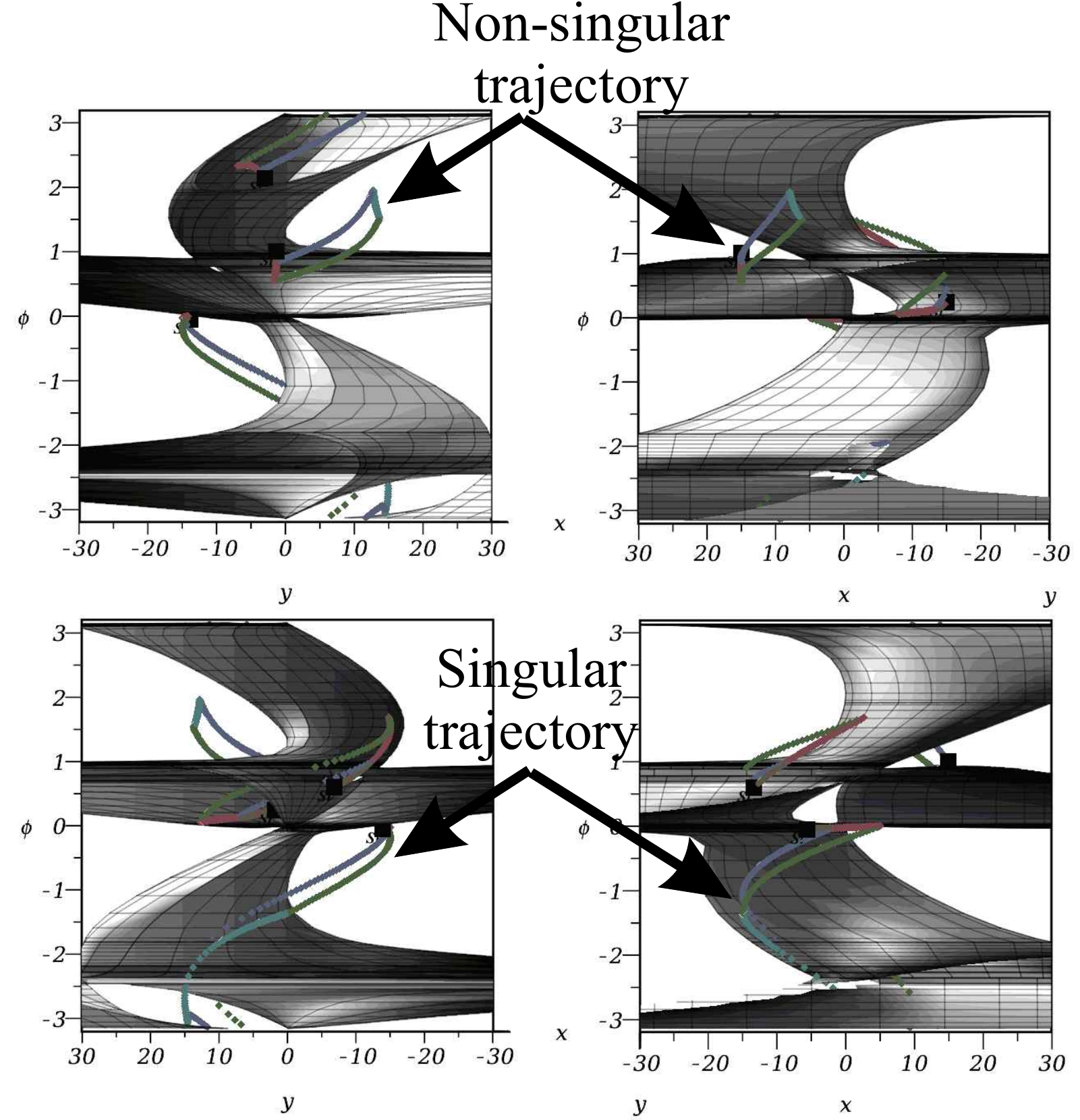}
   \caption{Assembly mode changing trajectories in the workspace for the 3--R\underline{P}R robot (projections on (x,$ \phi$) and (y, $\phi$) , $\phi$ expressed in radians).}
  \label{fig:trajectory_3RPR_cart}
\end{figure}

Figure~\ref{fig:trajectory_3RPR_cart} shows the singularity surfaces in the workspace with the non-singular assembly changing trajectory (top) and a singular assembly mode changing trajectory (bottom) . 


\begin{figure}
   \centering
   \includegraphics[width=0.5\linewidth]{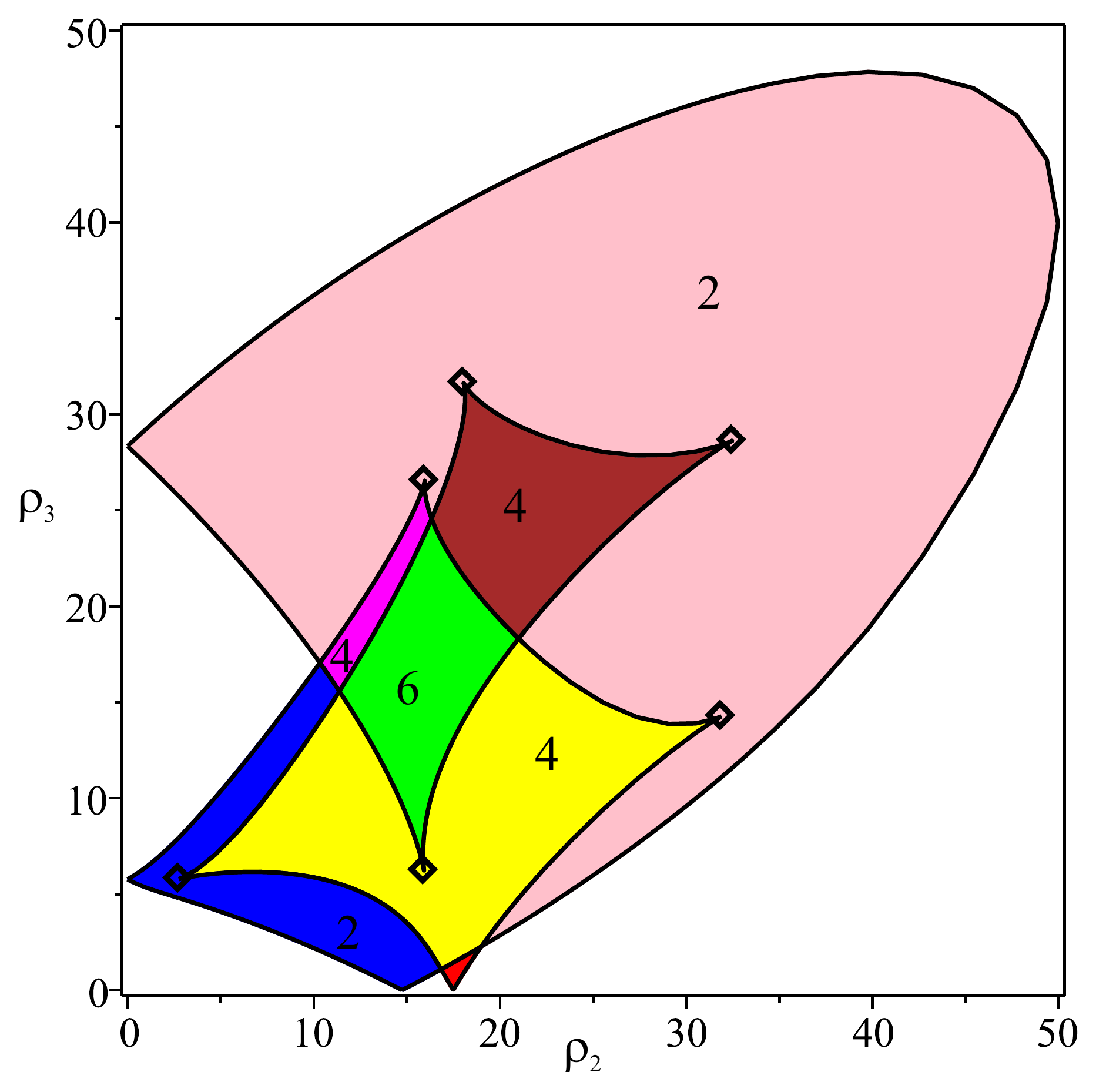}
   \caption{Section of the joint space of the 3--R\underline{P}R robot for $\rho_1=17$ with regions with 2, 4 and 6 assembly modes. There are 5 cusp points indicated by diamonds.}
  \label{fig:3RPR_coupe_art}
\end{figure}

An analysis of the joint space of this robot was presented in \cite{zein2007singular} in sections for given values of $\rho_1$. Figure~\ref{fig:3RPR_coupe_art} shows a section of the joint space in the plane ($\rho_2, \rho_3$) at $\rho_1=17$. There are five cusp points (surrounded by $\diamond$ diamonds) and the singularities divide regions with two, four or six assembly modes.  
\subsection{Identifying cuspidal parallel robots}
A robot encounters a parallel singularity when the parallel Jacobian matrix loses rank. Moreover, two assembly modes coincide on a parallel singularity.
When three assembly modes merge, the robot is said to be in a cuspidal configuration \cite{moroz2010determination}. A cuspidal configuration can be characterized by a triple root of the polynomial of the direct kinematics. \vmc{This means that a parallel robot whose direct kinematics can be solved with polynomial equations of degree lower or equal to two, cannot have cuspidal configurations}.
McAree and Daniel showed that in such configurations, the robot loses its first and second order constraints \cite{mcaree1999explanation}. In practice, this means that the robot is very unstable in a cuspidal configuration.

In \cite{moroz2010determination}, the notion of discriminant manifold and a generalization of the Jacobian criterion allow a complete and certified description of cuspidal configurations. This method has been implemented in the SIROPA library of the computer algebra software Maple \cite{chablat2019using}.
\subsection{Classification of parallel robots}
There are few studies on the classification of parallel robots. We can note the work of Merlet \cite{merlet1996direct} on planar parallel robots where the maximum number of solutions to the direct kinematics is given according to the architecture of the legs and the position of the actuated joint. For planar parallel robots, there can be two, four, or six solutions to the direct kinematics \cite{merlet1996direct}.

For 3--R\underline{P}R robots, geometric conditions exist for the direct kinematics to be solvable with second or third degree equations. Robots having this feature are called analytic \cite{KONG20011009}.

\begin{figure}
    \centering
    \includegraphics[width=0.5\linewidth]{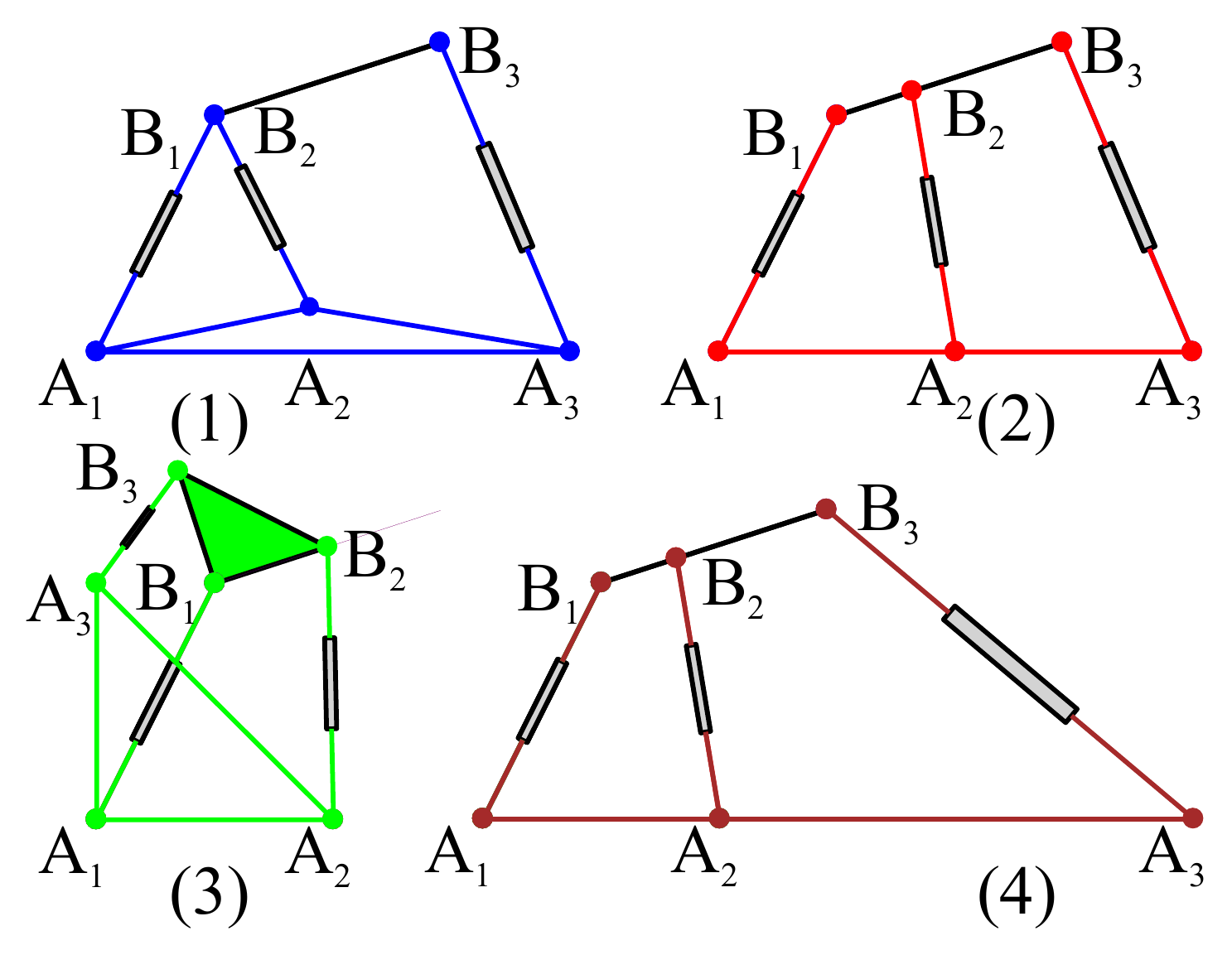}
    \caption{Examples of analytical 3--R\underline{P}R robots of types 1, 2, 3 and 4.}
    \label{fig:3-rpr-analytic}
 \end{figure}

Four types of analytic 3--R\underline{P}R robots 
were found \cite{KONG20011009}~:
\begin{itemize}
\item {- Type 1}~: Robots for which two of the base or moving platform joints coincide~(Fig. \ref{fig:3-rpr-analytic}, top left);
\item {- Type 2}~: Robots whose platform and base are aligned and arbitrary~(Fig. \ref{fig:3-rpr-analytic}, top right);
\item {- Type 3}~: Robots whose platform and base form congruent triangles
~(Fig. \ref{fig:3-rpr-analytic}, bottom left) ;
\item {- Type 4}~: Robots whose platform and base are aligned and congruent~(Fig. \ref{fig:3-rpr-analytic}, bottom right).
\end{itemize}

For types 1, 3, and 4, the direct kinematics can be solved with quadratic equations in cascade. These robots therefore cannot have cuspidal configurations. We know that the absence of cuspidal configurations in a parallel robot does not mean that this robot is noncuspdidal \cite{coste2014nonsingular} but all studied examples of type 1, 3 and 4 robots proved noncuspidal. Moreover, \cite{kong2000determination} showed that type 3 and 4 robots are noncuspidal when their platforms are similar. Whether this result can be generalized to non-similar platforms or not is still an open question. For type 2 robots, we have to solve a cubic equation and a quadratic equation in sequence \cite{KONG20011009}. These robots can therefore have cuspidal configurations. Moreover, it was found recently that all 3--R\underline{P}R robots of type 2 admit up to four assembly modes and have three aspects and these robots are thus cuspidal \cite{Jose}.

\begin{figure}
    \centering
    \includegraphics[width=0.7\linewidth]{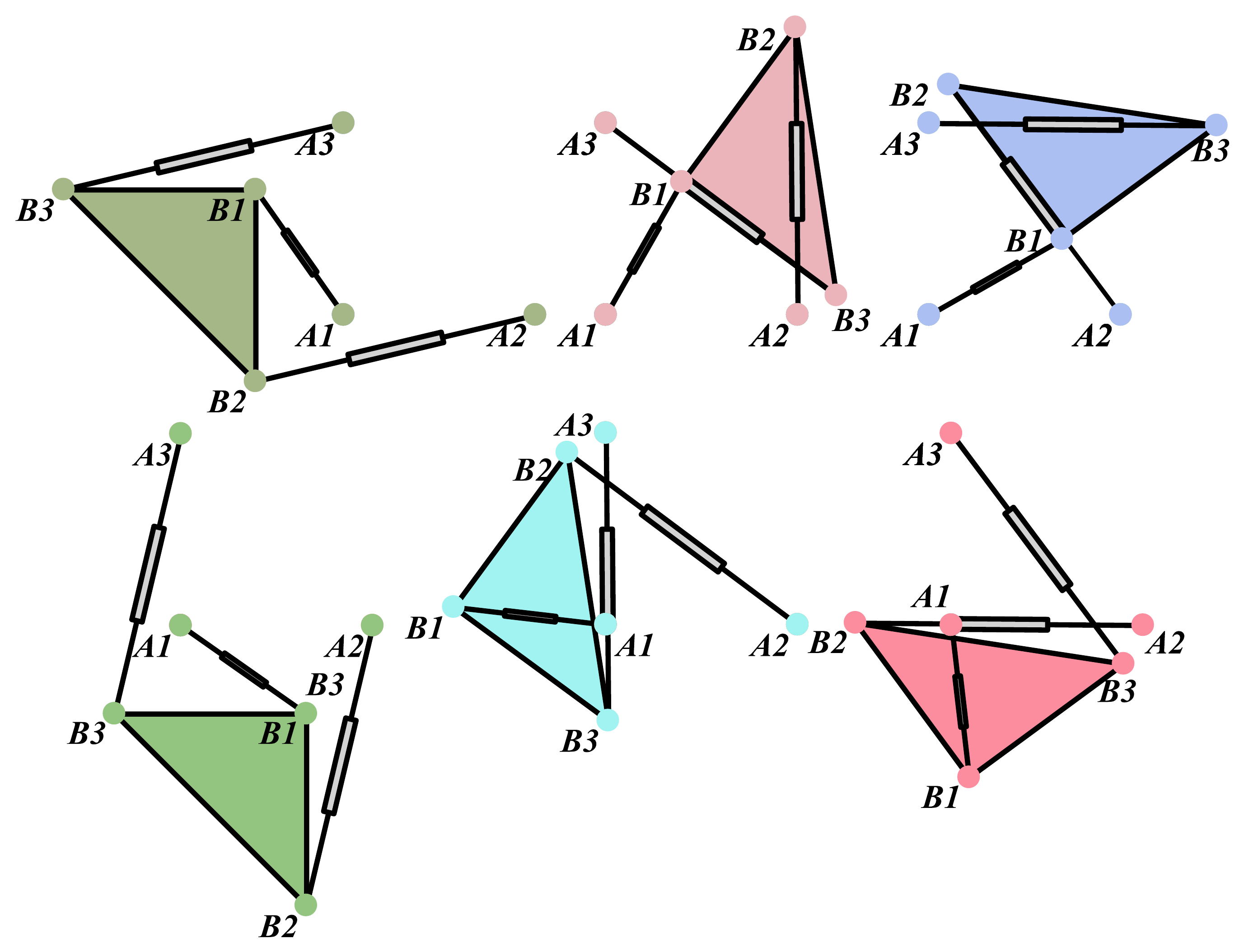}
    \caption{The six assembly modes of an analytical 3--R\underline{P}R robot with base and platform forming congruent triangles, one of which being upside down relative to the other.}
    \label{fig:3-rpr-analytic-degenerate}
\end{figure}

Another family of analytical robots was investigated in \cite{wenger2007degeneracy}. For these robots, the base and the platform form congruent triangles but one is rotated of 180 degrees about a line in plane. Their direct kinematics is solved with a cubic polynomial and a quadratic polynomial in cascade as for the type 2 robots described above. These robots have cuspidal configurations and are therefore cuspidal \cite{Analytic, coste2011singular}. In configurations where the platform is close to the base, the legs can cross (see Fig. \ref{fig:3-rpr-analytic-degenerate}). A feasible mechanical implementation of this robot would thus need to arrange the legs in different planes.

\subsection{Feasibility of trajectories for parallel robots}
The notion of t-connectivity for parallel robots is important and is analyzed directly in the workspace: parallel singularities divide the workspace into several distinct regions, which are the aspects.

A region of the workspace is said to be t-connected if any continuous trajectory in this region can be followed by the end-effector without ever encountering a parallel singularity. For a parallel robot, the t-connected regions are the aspects \cite{chablat2009characterization}.

If the aspects are the t-connected regions, two questions arise for a cuspidal parallel robot:
\begin{itemize}
    \item{-} What is the safety impact of assembly mode changes when planning paths in an aspect~? 
    \item{-} Are the t-connected regions larger than for a noncuspidal robot~?
\end{itemize}
\subsubsection{Impact of assembly mode changes}
When a parallel robot is cuspidal, it can change assembly mode without encountering a parallel singularity in a t-connected region. This means that the assembly modes at the starting and arriving point of a trajectory can be different but the control, which only knows the robot joint positions, cannot detect it. If the robot stops and loses its motion history, the current known assembly mode may be not the real one. It is therefore necessary, either to add sensors to judiciously chosen passive joints allowing the current assembly mode to be identified without ambiguity, or to always remain in a uniqueness domain of the direct kinematics.

The notion of uniqueness domains was defined for parallel robots with one or more solutions to their inverse kinematics \cite{chablat2011uniqueness}. In this paper, only the first formulation is considered. Uniqueness domains address the issue of safety by creating regions where the robot does not change assembly mode. As with serial robots, uniqueness domains require the computation of characteristic surfaces that separate solutions in the aspects.

Let ${\cal WA}_j$ be an aspect in the workspace W. We define the \emph{characteristic surfaces} of aspect ${\cal WA}_j$, denoted ${\cal S}_C({\cal WA}_j)$, as the reciprocal image in ${\cal WA}_j$ of the image in the joint space of $\overline{\cal WA}_j$, 
where $\overline{\cal WA}_j$ is the boundary of aspect ${\cal WA}_j$.

For each aspect, we have to consider the boundaries that correspond to both the singular configurations and the joint limits. 
Like for serial robots, the calculation of the characteristic surfaces is difficult to perform when the robot has more than two aspects, see \cite{chablat1998domaines} for more details.

\begin{figure}
    \centering
    \includegraphics[width=0.6\linewidth]{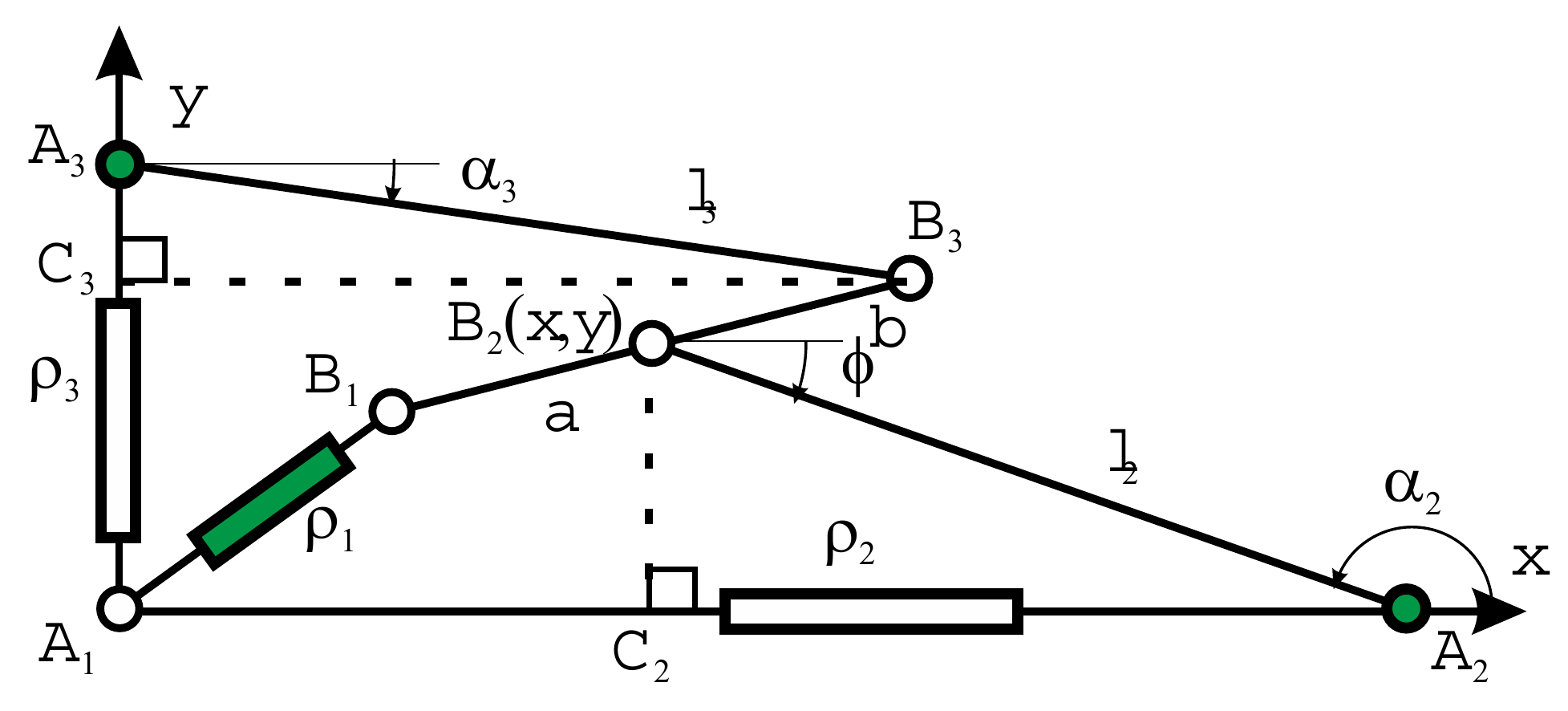}
    \caption{A \underline{P}R--2\underline{R}PR parallel robot.}
    \label{fig:robot_prp_2rpr}
\end{figure}

Let us consider a planar parallel R\underline{P}R--2\underline{R}PR robot as an illustrative example (Fig.~\ref{fig:robot_prp_2rpr}. This parallel robot, first proposed in \cite{5272470}, is composed of three legs and a moving platform. The first leg, of the R\underline{P}R type, is made up of two passive revolute joints and an actuated prismatic joint. The two other legs, of type \underline{R}PR, are made up of an actuated revolute joint, a prismatic joint and a passive revolute joint. The three actuated variables are described by $\rho_1$, $\alpha_2$ and $\alpha_3$. The moving platform has three degrees of freedom defined by its position and orientation coordinates $(x,~y,~\phi)$).

The constraint equations are as follows \cite{chablat2011uniqueness}:
\begin{eqnarray}
    \rho_2+l_2 \cos (\alpha_2) - x &=& 0 \nonumber\\
    l_2 \sin (\alpha_2) - y &=& 0\nonumber\\
    (x-a \cos(\phi))^2 + (y-a\sin(\phi))^2 - \rho_1^2 &=&0\nonumber\\
    l_3 \cos(\alpha_3) - b\cos (\phi) - x &=& 0\nonumber\\
    \rho_3 + l_3 \sin(\alpha_3) -b\sin (\phi)- y &=& 0\nonumber
    \end{eqnarray}
    
\begin{figure}
\centering
\includegraphics[width=0.4\linewidth]{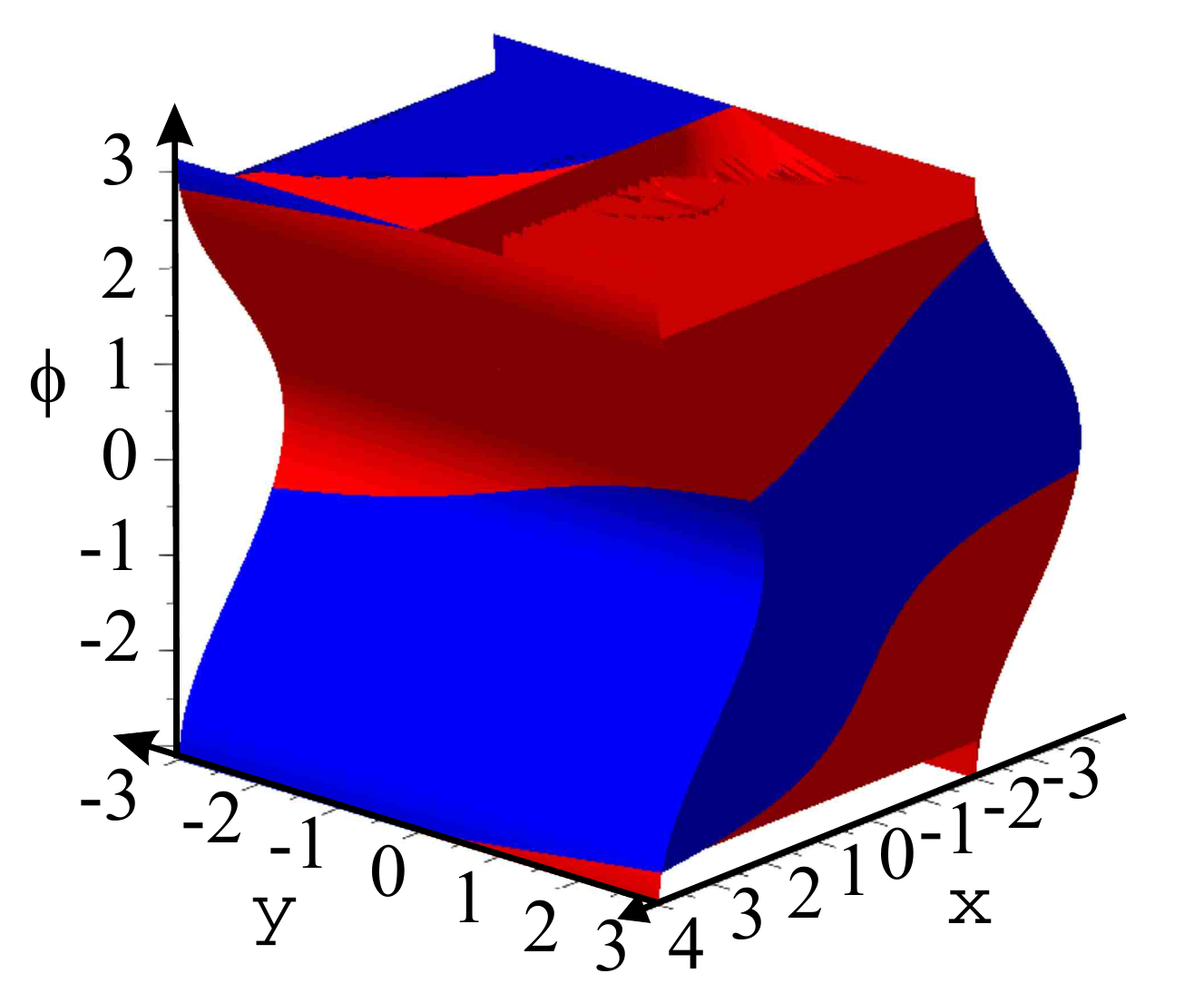}
\caption{The two aspects of the R\underline{P}R--2\underline{R}PR parallel robot.}
\label{fig:robot_prp_2rpr_aspect}
 
\end{figure}

\begin{figure}
\center  
\includegraphics[width=0.5\linewidth]{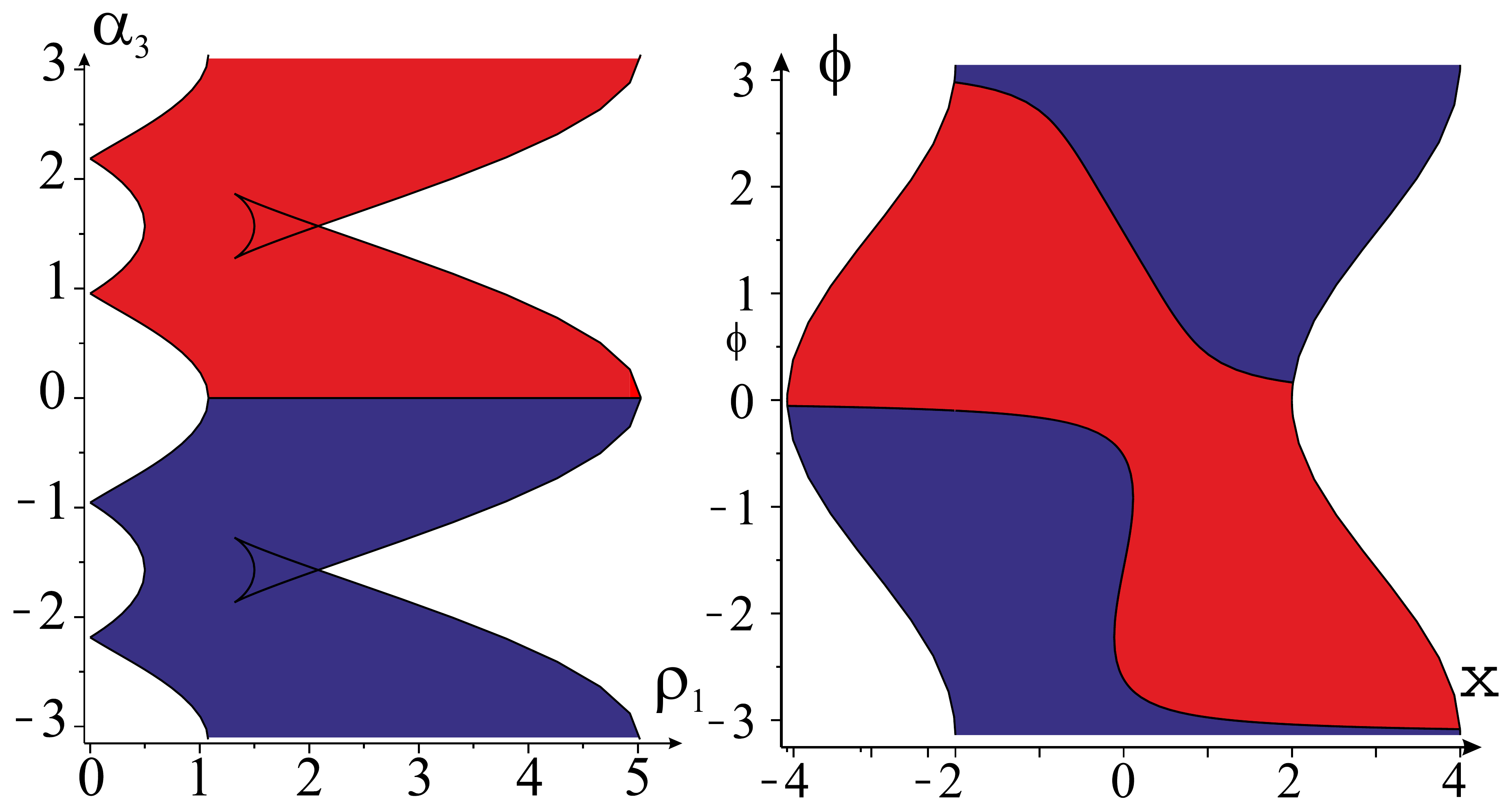}
\caption{Section of joint space at $\alpha_2 = \arcsin{1/6}$ (left) and of workspace at $y=1/2$ (right) of the  R\underline{P}R--2\underline{R}PR parallel robot.}
\label{fig:robot_prp_2rpr_cusp}
\end{figure}

The workspace of this robot is composed of two aspects, in which the number of assembly modes is two or four \cite{chablat2011uniqueness} (Fig.~\ref{fig:robot_prp_2rpr_aspect}). In a section of the joint space at $\alpha_2 = \arcsin{1/6}$, we can see four cusp points (Fig.~\ref{fig:robot_prp_2rpr_cusp}).


\begin{figure}
    \centering
    \begin{minipage}[t]{0.48\textwidth}
        \centering
        \includegraphics[width=0.7\linewidth]{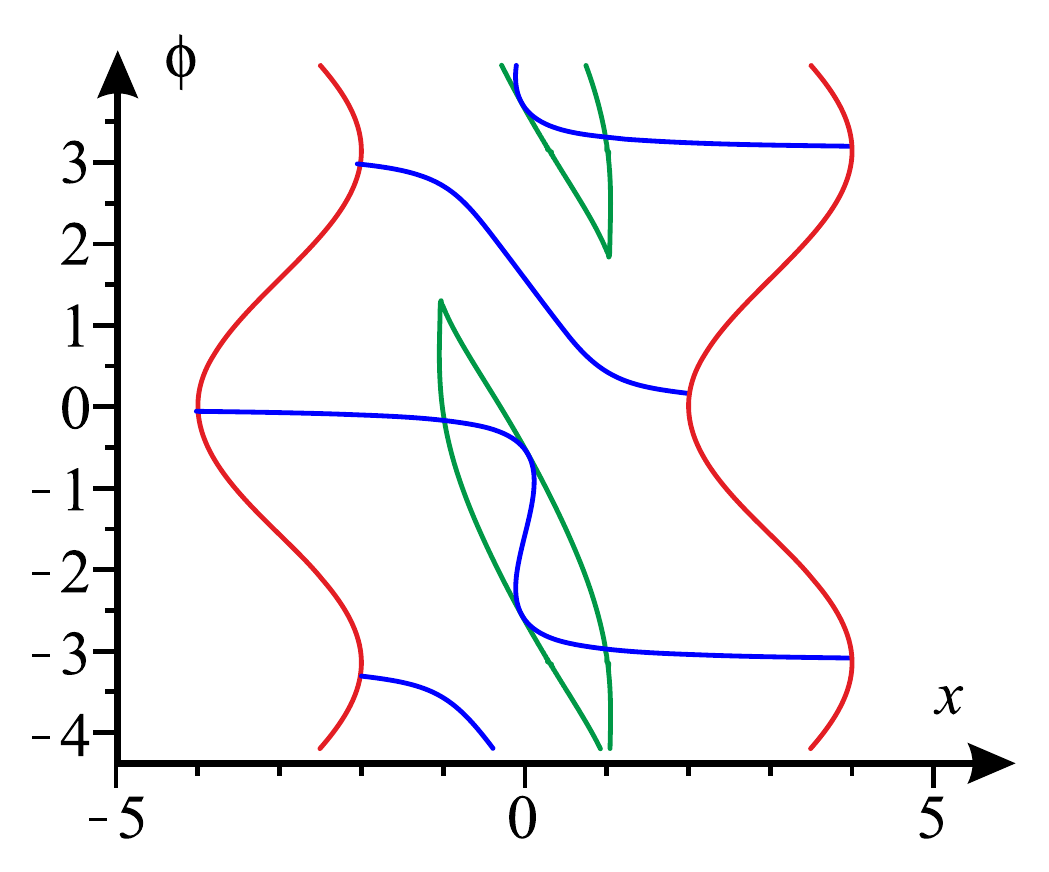}
        \caption{Singularities (in blue) and characteristic surfaces (in green) of the R\underline{P}R--2\underline{R}PR parallel robot.}
        \label{fig:robot_prp_2rpr_surface_characteristic}
    \end{minipage}
    \begin{minipage}[t]{0.48\textwidth}
        \centering
        \includegraphics[width=0.9\linewidth]{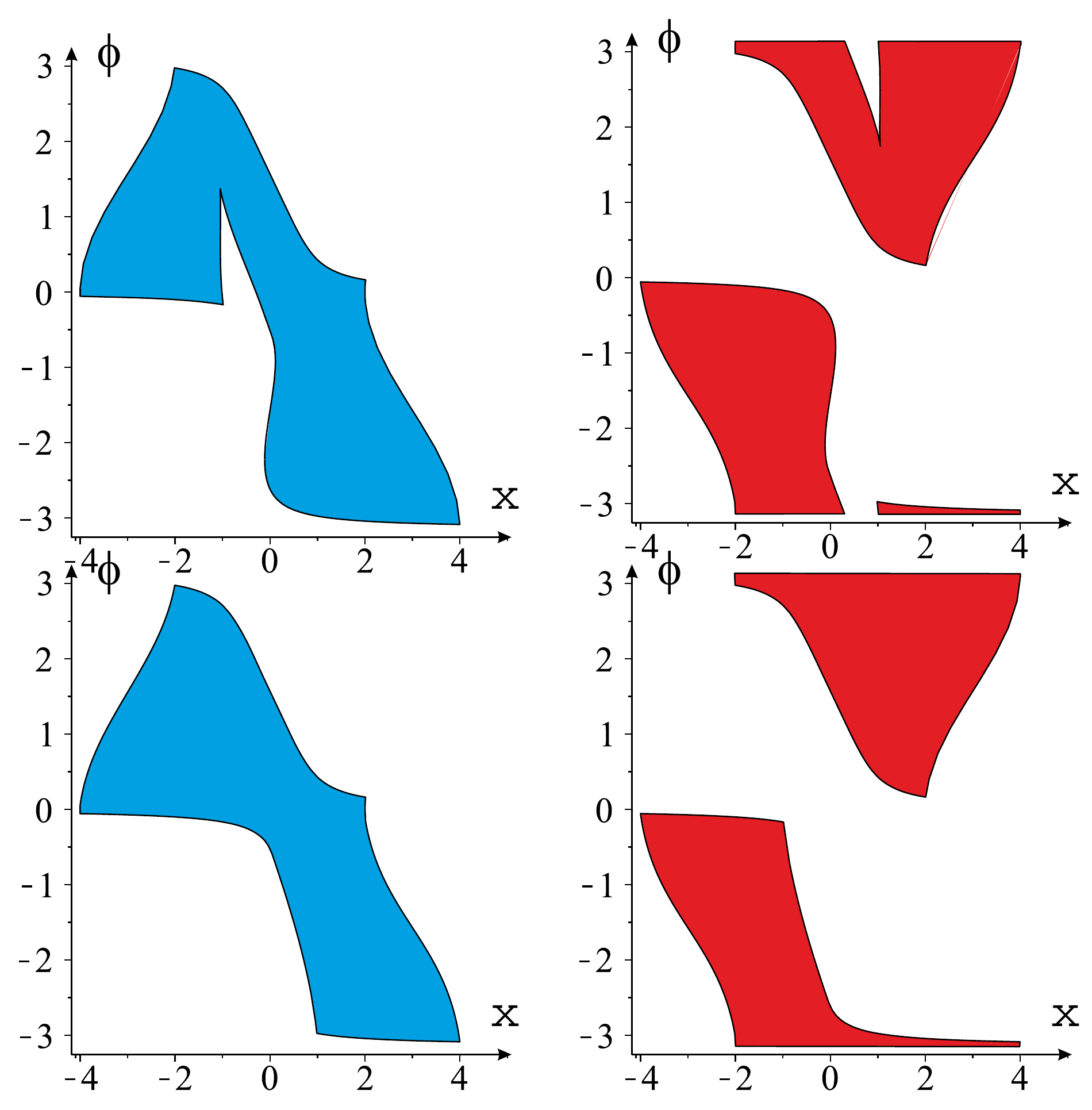}
        \caption{Uniqueness domains of the  R\underline{P}R--2\underline{R}PR parallel robot.}
        \label{fig:robot_prp_2rpr_domains_uniqueness}
    \end{minipage}
\end{figure}

The calculation of the characteristic surfaces of this robot is simple because there are only two aspects. The characteristic surfaces are tangent to the singularities (Fig. \ref{fig:robot_prp_2rpr_surface_characteristic}) and the tangency point is the image of a cusp point. 
Figure~\ref{fig:robot_prp_2rpr_domains_uniqueness} depicts the four uniqueness domains.

\begin{figure}
    \centering
    \includegraphics[width=0.4\linewidth]{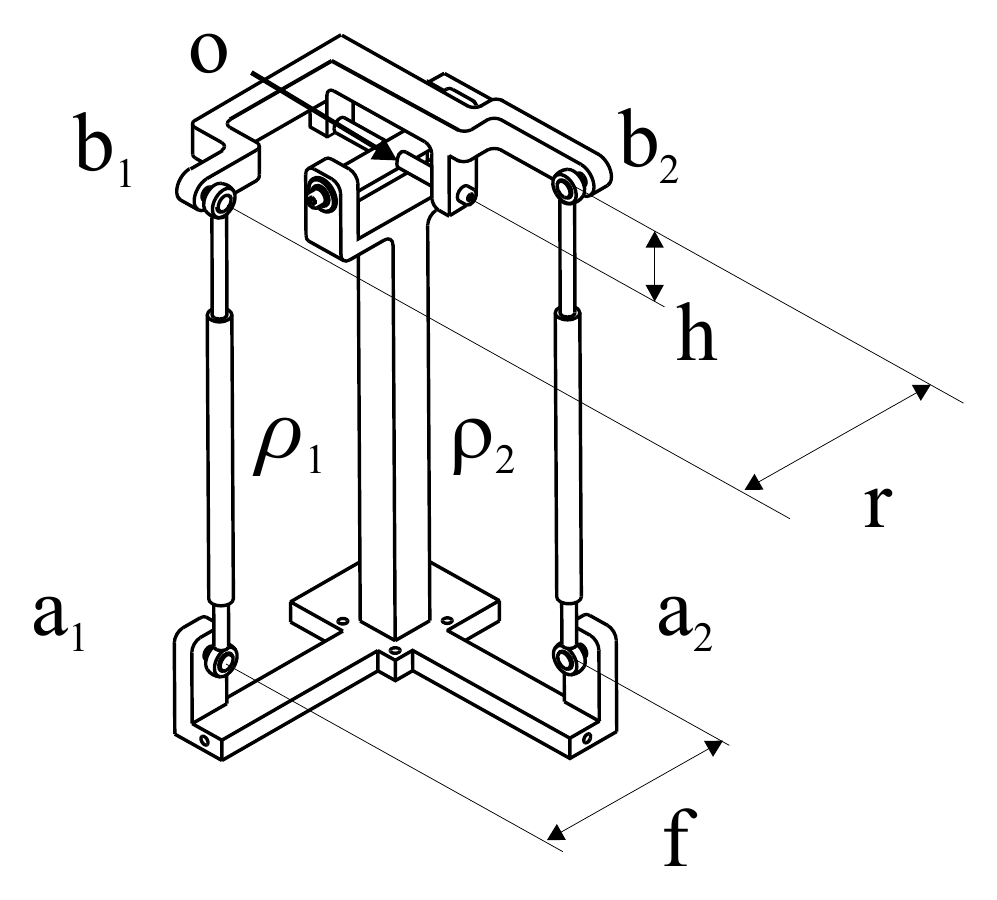}
    \caption{2--U\underline{P}S--U Spherical Parallel Robot.}
    \label{fig:spherical_robot}
\end{figure}

Let us now consider a second example: the 2--U\underline{P}S--U spherical parallel robot \cite{chablat157workspace}. This parallel robot is composed of three legs and a moving platform. The first two legs, U\underline{P}S, are composed of a universal joint, an actuated prismatic and a spherical joint. The last leg is passive and consists of a universal joint. This passive leg restricts the mobility of the moving platform to two rotations $\alpha$ and $\beta$ (Fig.~\ref{fig:spherical_robot}). 

The two prismatic joints are actuated and defined by $\rho_1$ and $\rho_2$. The constraint equations are \cite{chablat157workspace}:
\begin{eqnarray}
 -2( fh+ \cos{\alpha} r ) \sin{\beta} + 2( h\cos{\alpha} - fr ) \cos{ \beta }  +f^2+h^2+r^2+1 &=& {\rho}_1^2 \\
 2 h (f\sin{\alpha } +\cos{\alpha} ) \cos{\beta } -2 f\cos{\alpha} r+2 \sin{\alpha} r +f^2+h^2+r^2+1 &=& {\rho}_2^2 
\end{eqnarray}

\begin{figure}
    \centering
    \begin{minipage}[t]{0.45\textwidth}
        \centering
        \includegraphics[width=0.7\linewidth]{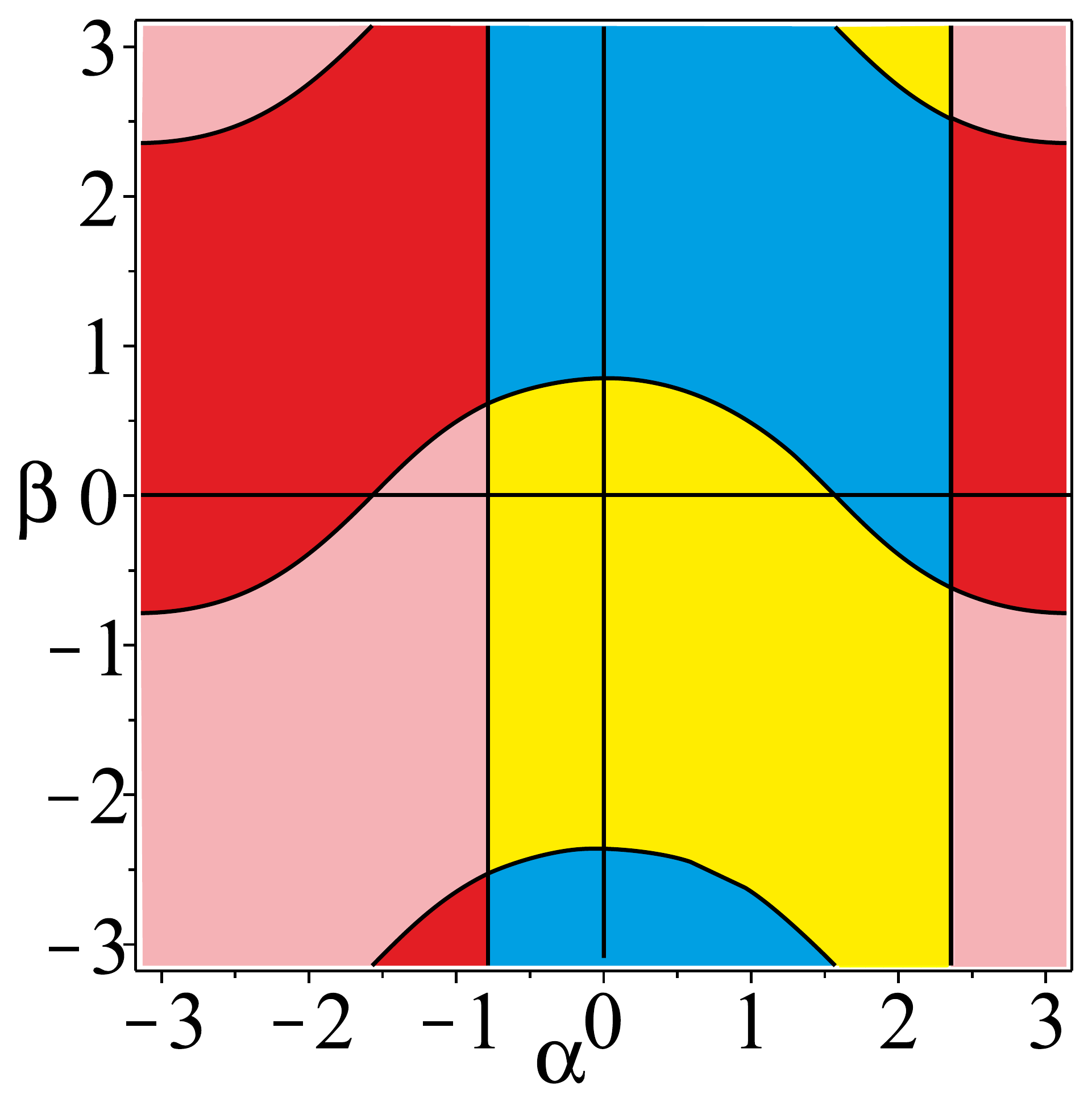}
        \caption{Workspace of the 2--U\underline{P}S--U spherical parallel robot for $h=0$, $r=1$, $f=1$ where each color is associated with an aspect (four assembly modes).}
        \label{fig:workspace_robot_spherical_1}
    \end{minipage}
    \begin{minipage}[t]{0.45\textwidth}
        \centering
        \includegraphics[width=0.7\linewidth]{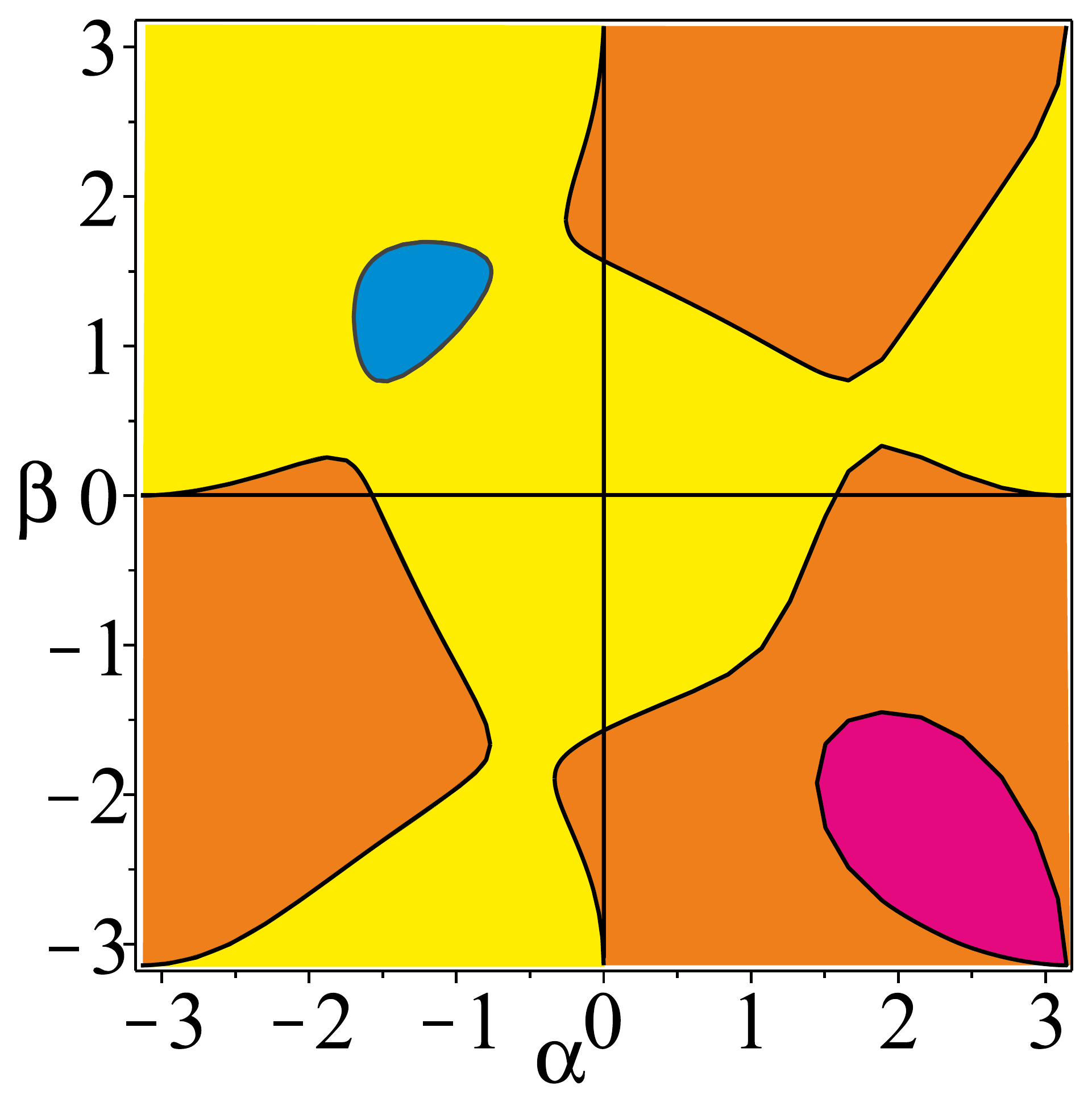}
        \caption{2--U\underline{P}S--U spherical parallel robot workspace for $h=1$, $r=1$, $f=1/2$ with three assembly modes in the aspects filled in yellow and orange.}
        \label{fig:workspace_robot_spherical_2}
    \end{minipage}
\end{figure}

Depending on the design parameters, the number of aspects as well as the number of solutions per aspect change. Figures~\ref{fig:workspace_robot_spherical_1} and \ref{fig:workspace_robot_spherical_2} show two workspace examples resulting in different cuspidality conditions. For the parameters of the robot in Fig.~\ref{fig:workspace_robot_spherical_1}, the singularity condition factors and the robot has four assembly modes and four aspects (each color is associated with an aspect). There is only one solution to the direct kinematics per aspect: the robot is therefore noncuspidal. For the parameters of the robot in Fig.~\ref{fig:workspace_robot_spherical_2}, the robot has also four aspects but it admits up to six assembly modes. Moreover, we can show that there is up to three solutions in the yellow aspect  \cite{chablat157workspace}: the robot is therefore cuspidal.

\subsubsection{Size of t-connected regions}
When a noncuspidal parallel robot admits $n$ solutions to its direct kinematics, the workspace is divided into a minimum of $n$ aspects (if the joint limits allow it). Conversely, a cuspidal robot has several solutions to its direct kinematics in the same aspect. It is then possible to have larger t-connected regions. This question was first investigated in \cite{WorkspaceAM98} using a 3--R\underline{P}R planar parallel robot.

Let us consider the two designs of the spherical parallel robot 2--U\underline{P}S--U analyzed in the example above (Figs~\ref{fig:workspace_robot_spherical_1} and \ref{fig:workspace_robot_spherical_2}). These two robots have the same number of aspects but for the cuspidal one, the aspects filled in yellow and orange are larger.

More specifically, the aspect filled in yellow around the original position ($\alpha=0$, $\beta=0$) of Fig.~\ref{fig:workspace_robot_spherical_2} is larger than that located at the same location of Fig.~\ref{fig:workspace_robot_spherical_1}. It is therefore more interesting for the user. However, to ensure safety without adding sensors, it is necessary to know if the robot changes assembly mode during these motions.

\begin{figure}
 
\centering
    \includegraphics[width=0.4\linewidth]{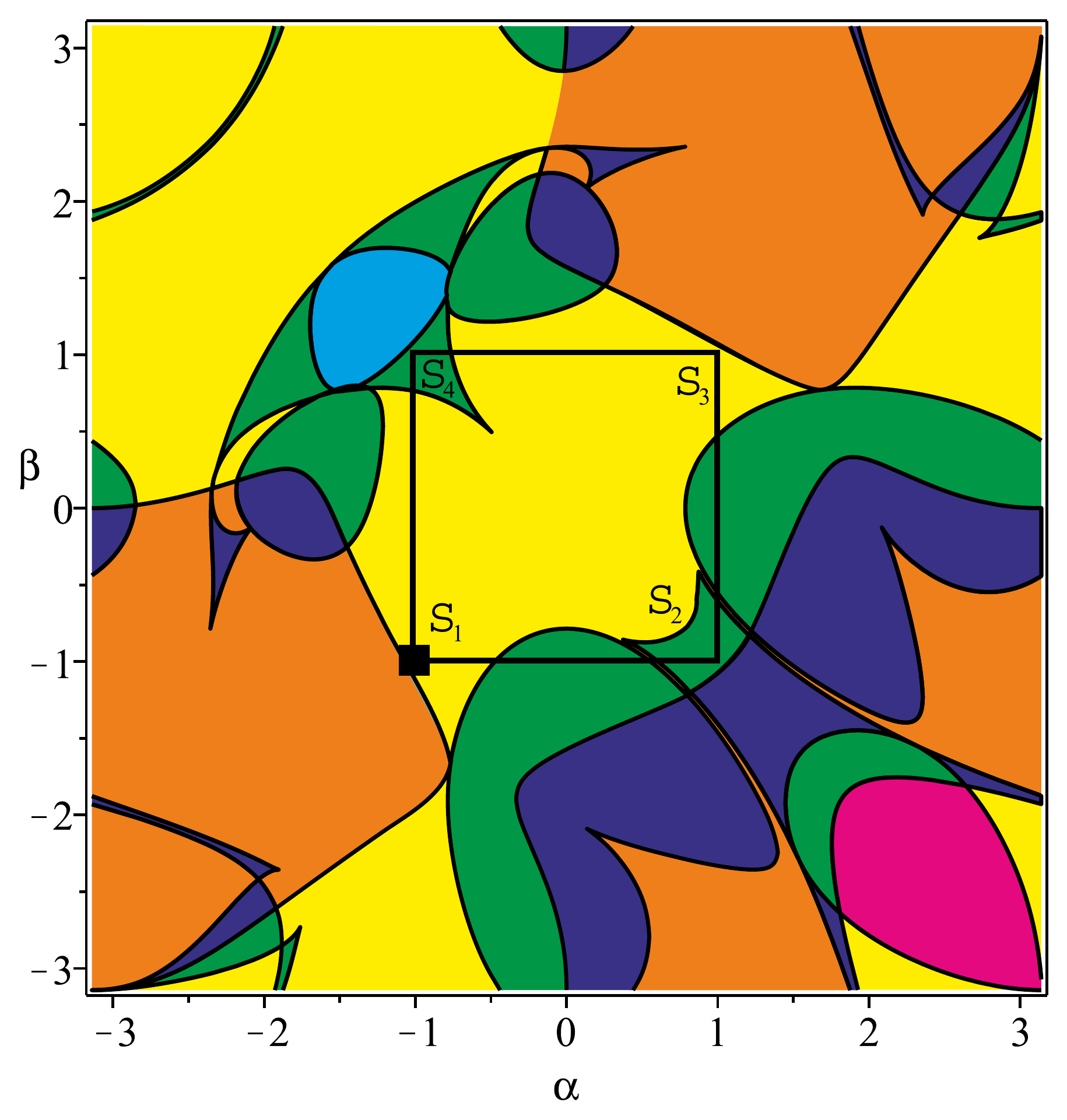}
    \caption{Characteristic surfaces and boundary of the prescribed workspace $\alpha=[-1~1]$ and $ \beta=[-1~1]$ for the 2--U\underline{P}S--U robot of Fig.~\ref{fig:workspace_robot_spherical_2}.}
    \label{fig:3-UPS-U-trajectory_cart}
\end{figure}

For this spherical parallel robot, \cite{chablat2020joint} and \cite{chablat2021workspace} have studied the position of singularities and characteristic surfaces so that the robot does not change assembly mode inside a prescribed workspace (Fig.~\ref{fig:3-UPS-U-trajectory_cart}). 

\begin{figure}
    \centering
    \includegraphics[width=0.4\linewidth]{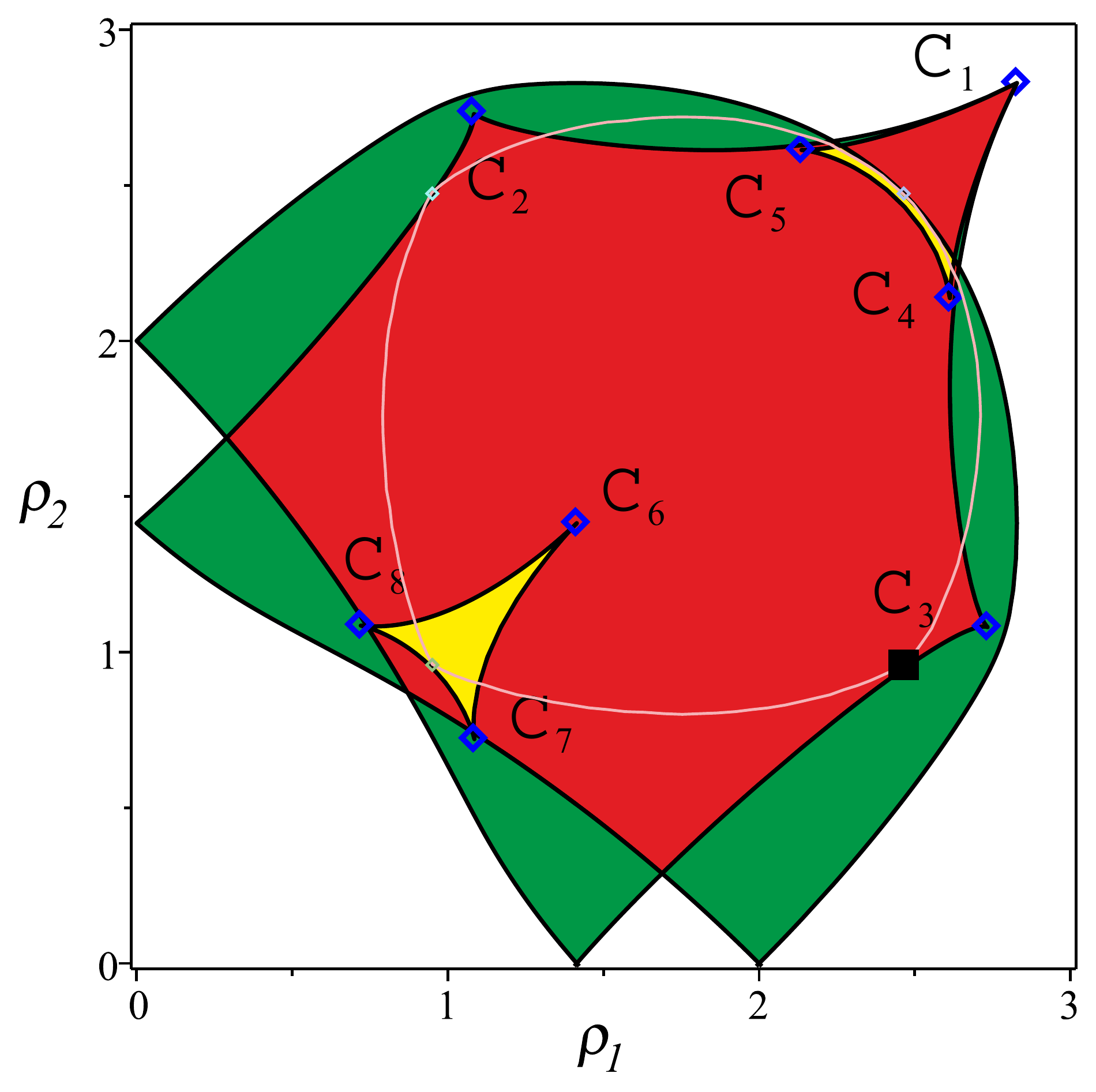}
    \caption{Joint space of the 2--U\underline{P}S--U spherical parallel robot with the regions with two (resp. four, six) assembly modes in green (resp. in red, in yellow), position of the cusp points as well as image of the regular workspace boundary.}
    \label{fig:3-UPS-U-trajectory_art}
\end{figure}

Figure~\ref{fig:3-UPS-U-trajectory_cart} shows the workspace with the characteristic surfaces. A prescribed workspace $\alpha=[-1~1]$ and $\beta=[-1~1]$ is located around the origin position ($\alpha=0$, $\beta=0 $). The image of the boundary of this prescribed workspace is shown in Fig.~\ref{fig:3-UPS-U-trajectory_art}. It has been shown that this trajectory encircles cusp points $C_4$, $C_5$ and $C_6$ but does not cause a change of assembly mode. For another robot or for another prescribed workspace, the conclusion might have been different.

\subsection{Case of robots with six degrees of freedom}
\begin{figure}
    \centering
    \includegraphics[width=0.4\linewidth]{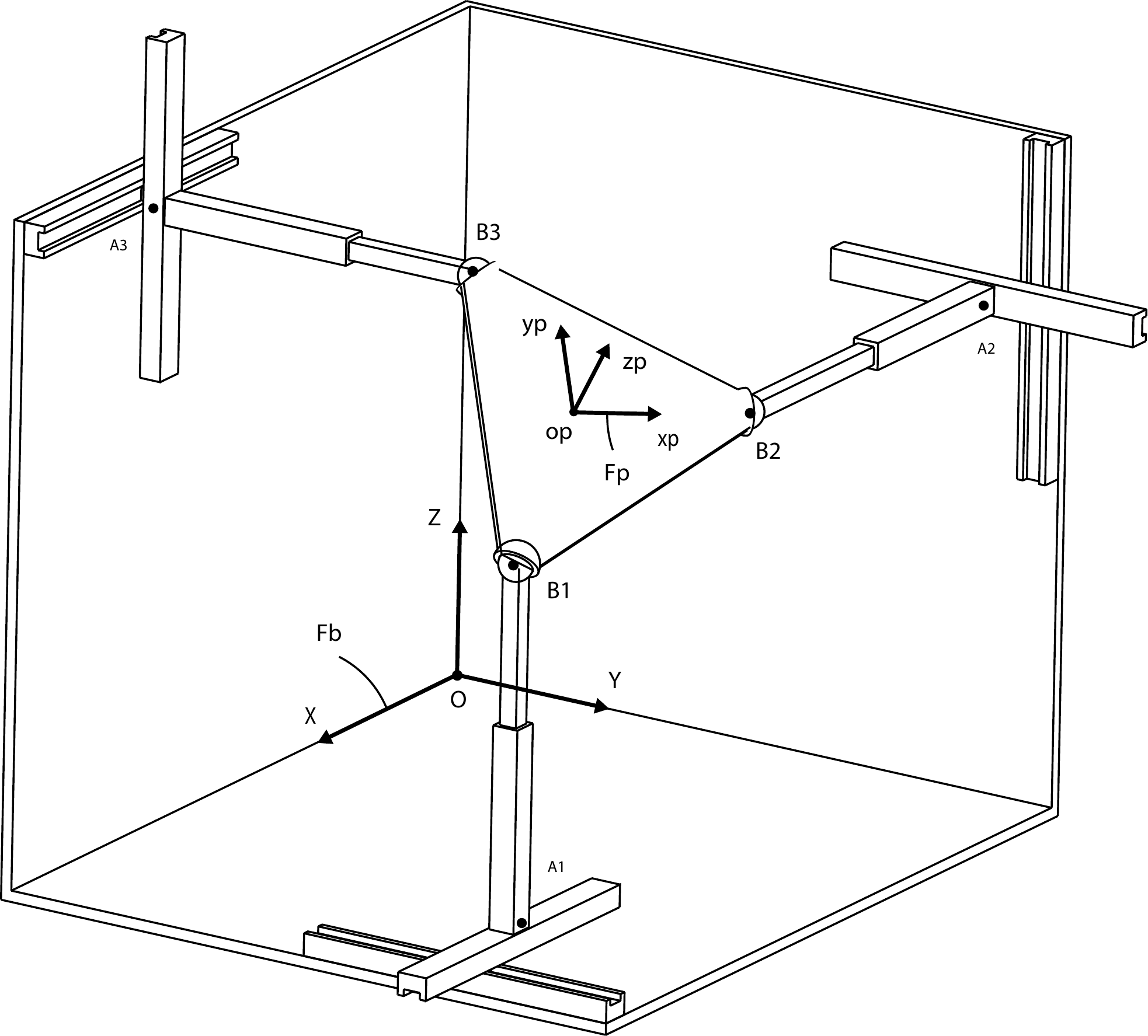}
    \caption{The 6-degree-of-freedom 3--\underline{PP}PS parallel robot.}
    \label{fig:3-UPPPS}
\end{figure}

The study of parallel robots with six degrees of freedom is challenging when the translational and rotational degrees of freedom are not decoupled, which is the most frequent situation. For such robots, the algebra involved is too complex and it is necessary to resort to numerical methods, such as, for example, a regular discretization of space or interval analysis. 
The following example is one of the few decoupled parallel robots that can be studied completely with algebraic tools: a 3-\underline{PP}PS orthogonal parallel robot (Fig. \ref{fig:3-UPPPS}).

This robot  has six degrees of freedom \cite{caro2012non} and its legs consist of three prismatic joints arranged orthogonally, two actuated joints followed by a passive one and a spherical joint . It can be shown that the singularity locus is independent of the position and is therefore a function only of the orientation. Indeed, by a judicious change of variable, it is possible to study the joint space of this robot with only three prismatic joints defined by $(x~,y~,z)$. The workspace can be described with three modified Euler angles ($\phi$~: \emph{azimuth}, $\theta$~: \emph{tilt} and $\sigma$~: \emph{torsion }) \cite{bonev2002advantages}.

The singularity condition can be factored as follows \cite{caro2012non}:
\begin{align}
    \label{eq:SingOrienWP}
 &\left( \sqrt {2}\sin \left( 3\,\phi-\sigma \right) \left( \sin \left( \frac{\theta}{2} \right) \right) ^{ 3}-\cos
 \left( \frac{3\theta}{2} \right) \cos \left( \sigma \right) \right) 
  \sin \left( \frac{\theta}{2} \right) \left( \cos \left( \frac{\theta}{2} \right) \right) ^{2} =0
\end{align}
A surface is then obtained which divides the workspace into two aspects.
    
\begin{figure}
    \centering
    \includegraphics[width=0.4\linewidth]{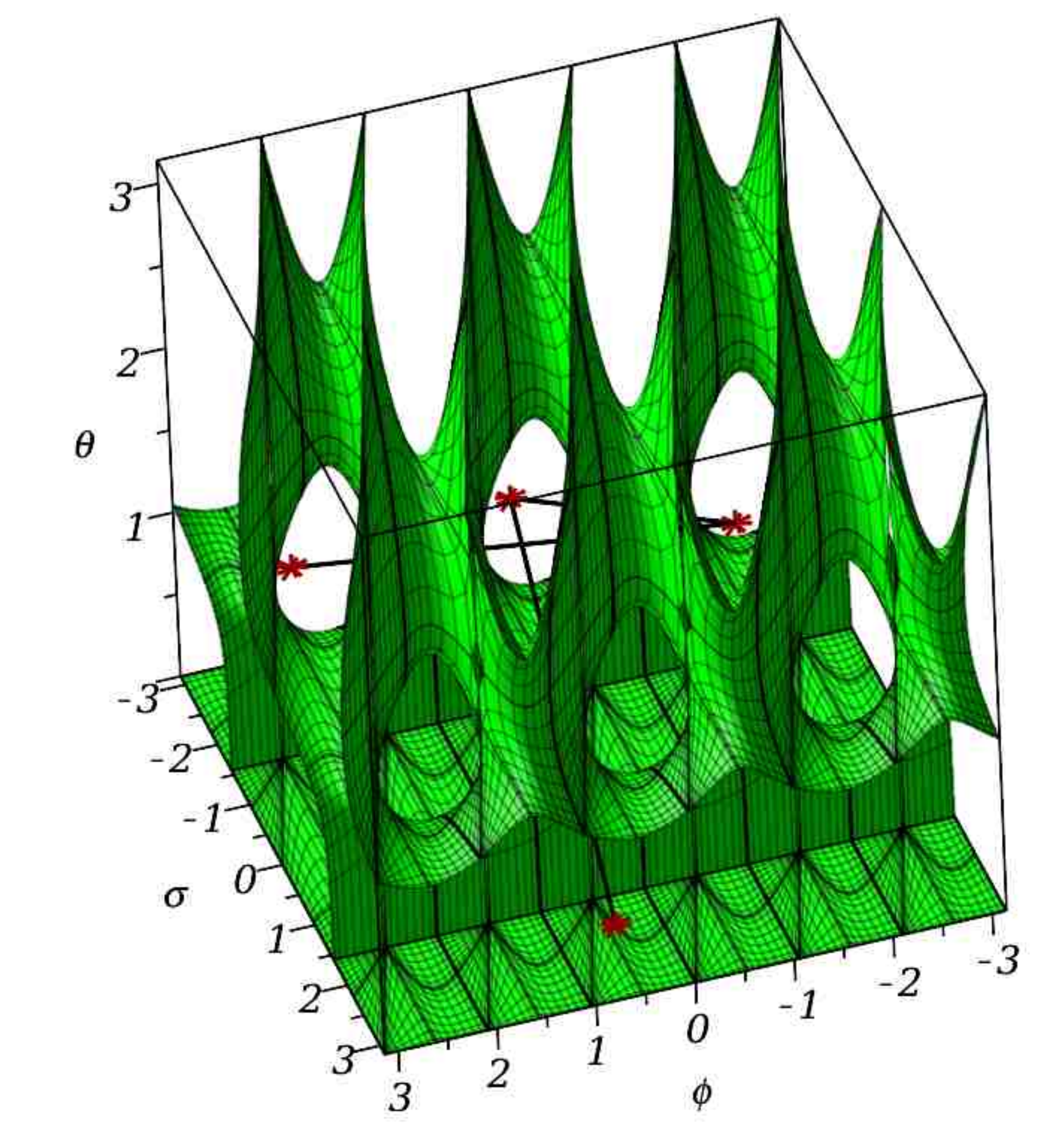}
    \includegraphics[width=0.4\linewidth]{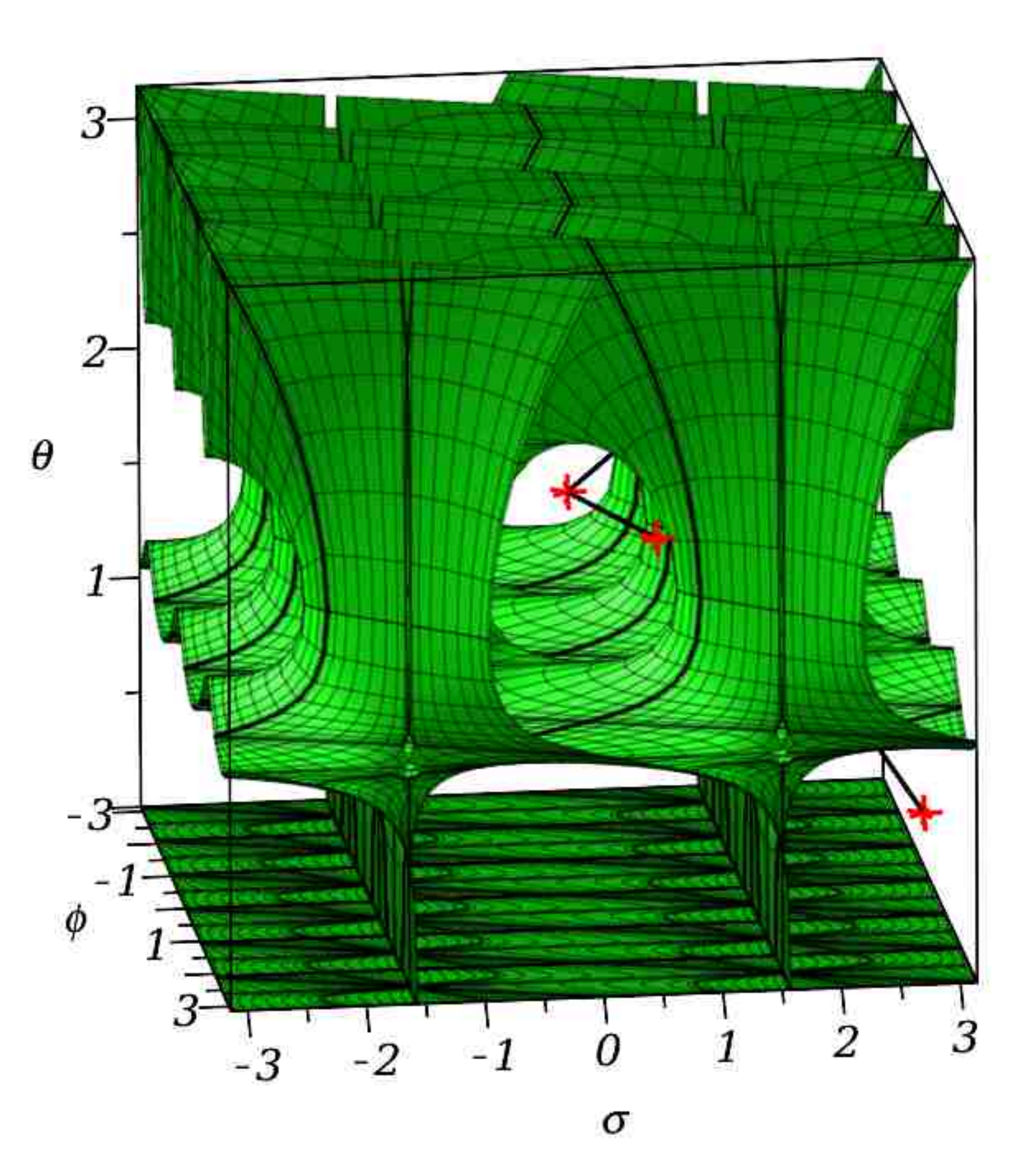}
    \caption{An assembly mode changing trajectory in workspace for the 3--\underline{PP}PS parallel robot.}
    \label{fig:3-UPPPS-traj-1}
\end{figure}

\begin{figure}
\centering
    \includegraphics[width=0.4\linewidth]{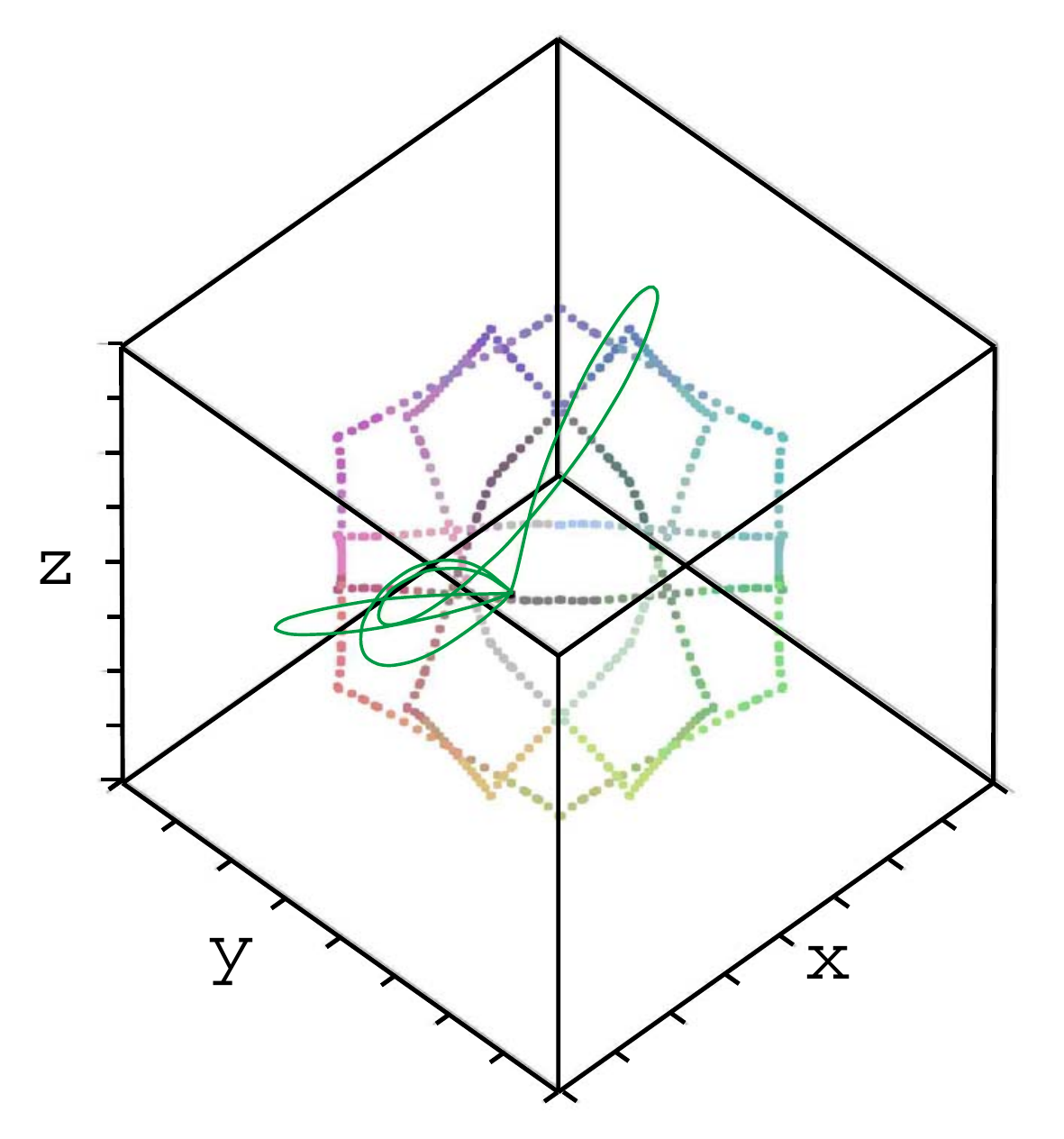}
    \caption{Cusp point curves of the 3--\underline{PP}PS parallel robot in the joint space (represented by a series of points) with the assembly mode changing trajectories in an aspect (continuous, green line).}
    \label{fig:3-UPPPS-cusp}
\end{figure}

Figure~\ref{fig:3-UPPPS-traj-1} shows singularities in the workspace as well as an example of a non-singular trajectory that connects four assembly modes for $x = 9/20$, $y = 0$ and $z = 1/40$. This trajectory encircles cusp points in the joint space. In fact, the cusp points are not isolated but form curves (Fig.~\ref{fig:3-UPPPS-cusp}). These curves can be obtained numerically by discretization but no algebraic or parametrized formulation could be obtained for this robot.
    
The possibility of changing its assembly mode without encountering a singularity allows this robot to have a very large workspace. In practice, it is necessary to take into account self-collisions in the definition of the workspace. This is a difficult problem. In \cite{merlet2006legs}, a methodology is proposed to determine the workspace that takes into account joint limits, singularities and self-collisions.

\subsection{A cuspidal parallel robot with no cusp point}\label{sec:Robotsanscusp}
For parallel robots as for serial robots, the presence of a cusp point is a sufficient condition for performing non-singular solution changes. However, this condition is not necessary as shown below.

\begin{figure}
    \centering
    \includegraphics[width=0.6\linewidth]{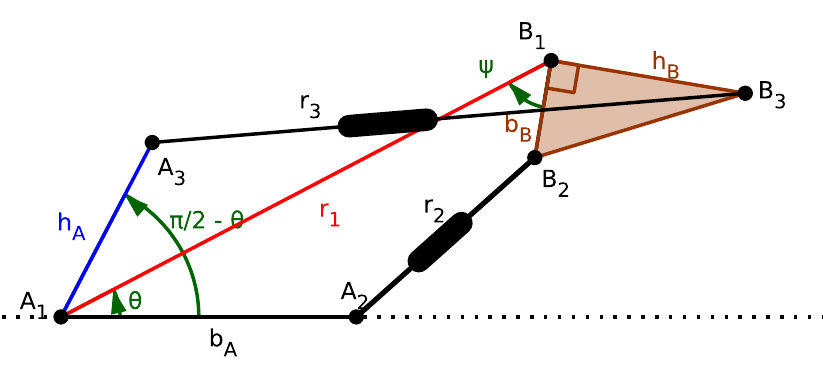}
    \caption{A 2--R\underline{P}R--RR parallel robot.}
    \label{fig:3-RPR-modified}
 \end{figure}

In \cite{coste2014nonsingular}, a change of non-singular assembly mode was pointed out for a robot without any cusp point. This robot, derived from the 3-R\underline{P}R robot, has a base whose position of point $A_3$ varies according to the orientation of the leg $(A_1B_1)$ (Fig.~\ref{fig:3-RPR-modified}). The moving platform is a right-angled triangle formed by points ($B_1, B_2, B_3$). Two of the three legs are of the R\underline{P}R type with actuated prismatic joints ($r_2, r_3$). For the other leg (in red in Fig. \ref{fig:3-RPR-modified}), the prismatic joint is replaced by a bar of fixed length $r_1$. The robot therefore has only two degrees of freedom. The output variables defining the workspace are the two angles $\theta$ and $\psi$ shown in Fig. \ref{fig:3-RPR-modified}.

The constraint equations of this robot are \cite{coste2014nonsingular}:
\begin{equation} \label{eq:robot_3-rpr-mobile}
  \begin{aligned}
    l_2 &= (r_1-b_A\cos\theta-b_B\cos\psi)^ 2 +(b_A\sin\theta+b_B\sin\psi)^ 2\;,\\
    l_3 &= (r_1-h_A\sin 2\theta -h_B\sin\psi)^ 2 +(h_A\cos 2\theta +h_B\cos\psi)^2\;.
  \end{aligned}
\end{equation}
  where $l_2=r_2^2$ and $l_3=r_3^2$.
  
 Even if this architecture was created specifically to highlight a cuspidal behavior without cusp, it is feasible because it is possible to imagine an angle multiplication mechanism allowing to relate the angle $\theta$ with $\pi/ 2 - \theta$.
   
\begin{figure}
    \centering
    \begin{minipage}[t]{0.45\textwidth}
        \centering
        \includegraphics[width=0.8\linewidth]{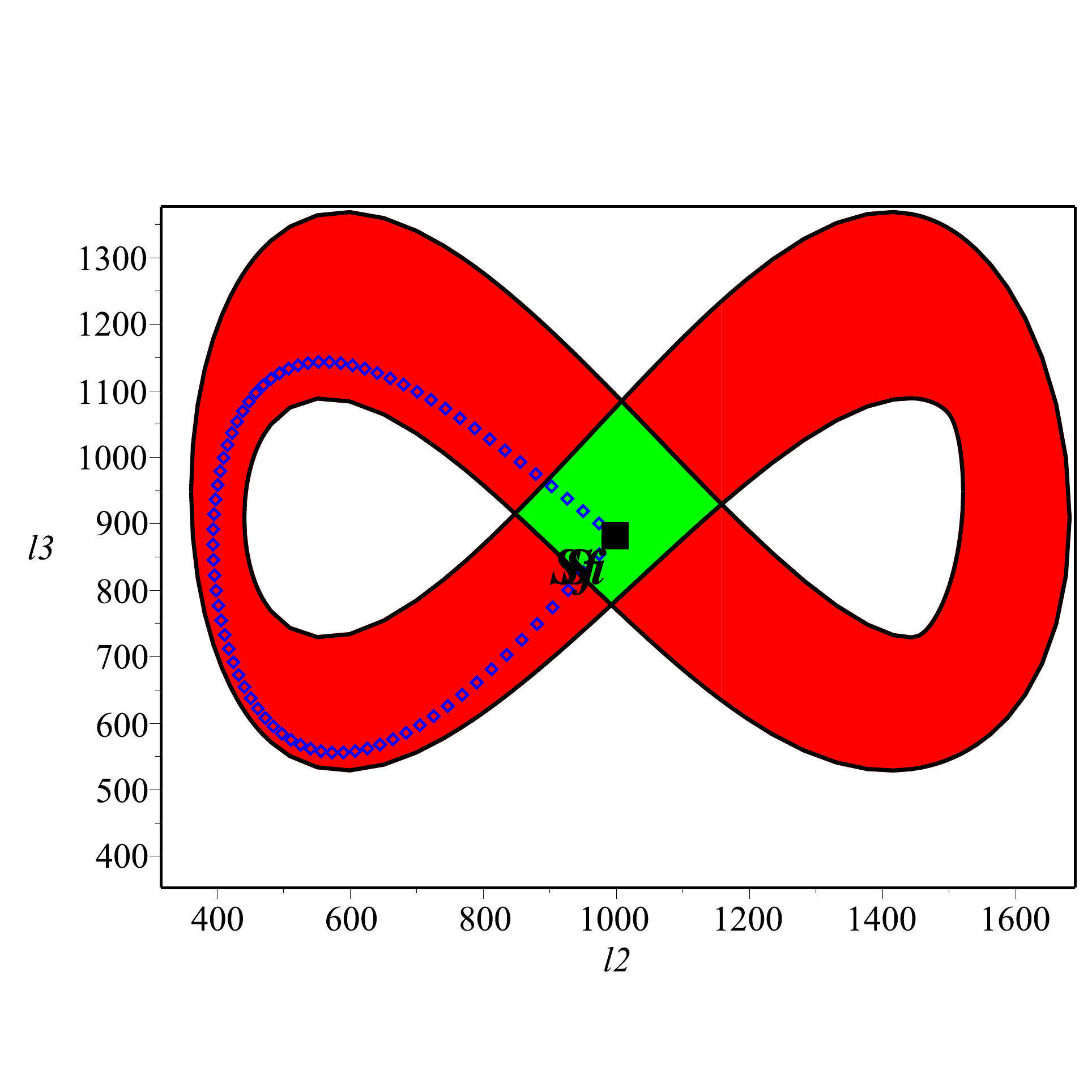}
        \caption{Workspace of the cuspidal 2--R\underline{P}R--RR parallel robot and non-singular assembly mode changing trajectory.}
        \label{fig:3-RPR-modifies-Q}
    \end{minipage}
    \begin{minipage}[t]{0.45\textwidth}
        \centering
        \includegraphics[width=0.8\linewidth]{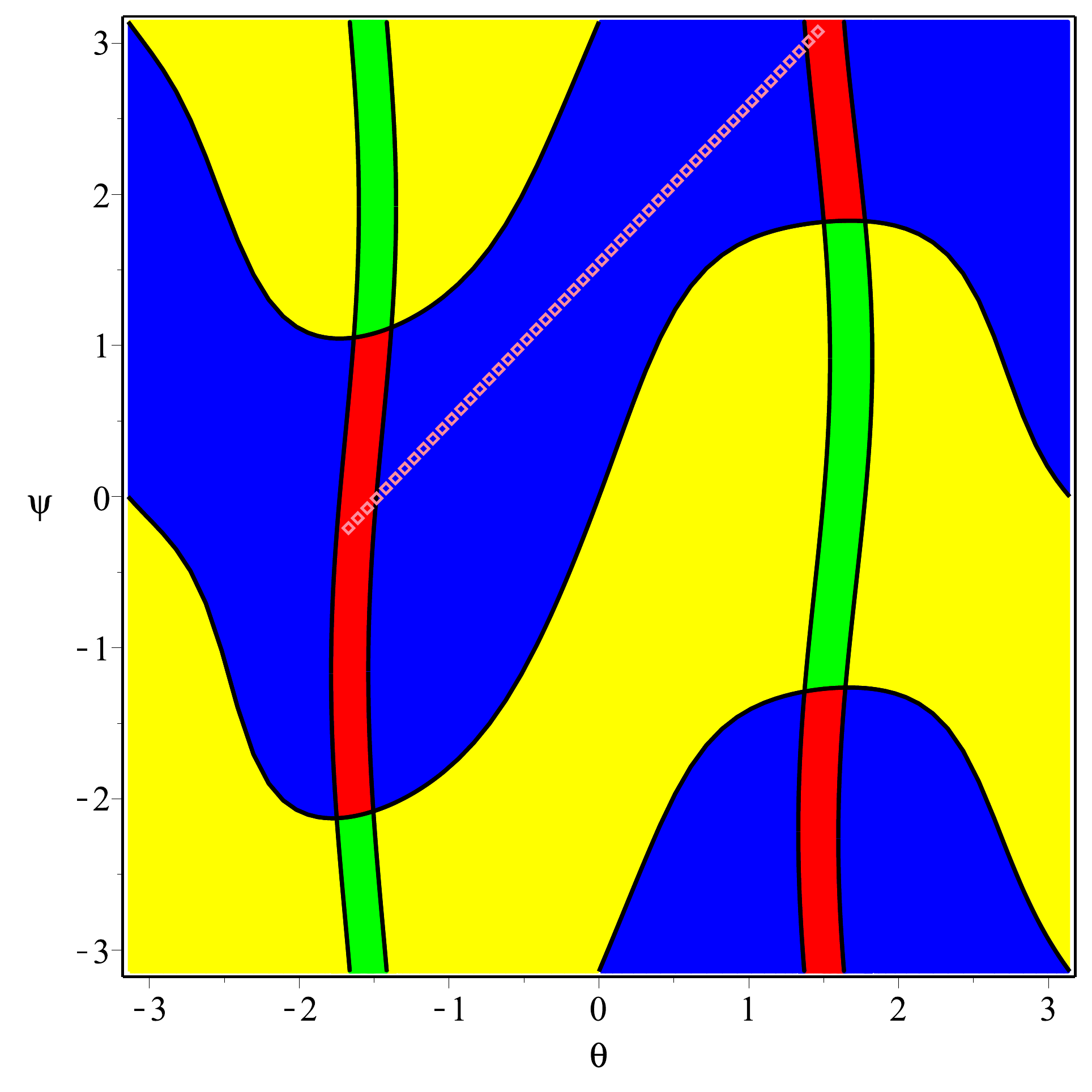}
        \caption{Joint space of the cuspidal 2--R\underline{P}R--RR parallel robot and non-singular assembly mode changing trajectory.}
        \label{fig:3-RPR-modifies-W}
    \end{minipage}
\end{figure}

 An assembly mode changing trajectory is shown in Fig.~\ref{fig:3-RPR-modifies-Q} in the workspace and in Fig.~\ref{fig:3-RPR-modifies-W} in the joint space for $l_2=1000$ and $l_3=800$ between two assembly modes: [$\theta=-1.66, \psi=-0.21$] and [$\theta=1.47, \psi=3.07$] for $r_1= 30$, $b_A= 10$, $h_A= 5$, $b_B= 1$ and $h_B=2$.
   
 We see that, contrary to cuspidal robots having cusp points, the characteristic surfaces are never tangent to the singularity surfaces but intersect them orthogonally. The only thing in common with cuspidal robots having cusp points is that the assembly mode changing trajectory passes through a region where the number of assembly modes changes from four to two.

\section{Conclusion}
The main objective of this paper was to \vmc{present cuspidal robots to uninitiated roboticists and to survey the latest results obtained in this field}. This property is little discussed in the literature and, as a consequence, it remains unknown by robot users and manufacturers. Yet, it may have great practical consequences when planning trajectories. A serial (resp. parallel) robot with at least one cusp point in its workspace (resp. in its joint space) is cuspidal but the converse is not always true.  Cuspidality is difficult to analyze because of the complexity of the equations involved. It thus requires the use of sophisticated algebraic tools. These tools are based in particular on Gröbner bases and Algebraic Cylindrical Decomposition \cite{moroz2010determination} and have been integrated into the \emph{Siropa} library available in the \emph{Maple} software \cite{chablat2019using}. The study of cuspidality is very important for the search for alternative robot architectures. We have seen that cuspidality complicates the analysis of trajectories and is definitely not desirable for a serial robot. The kinematic architectures of serial industrial robots are very little varied and obey geometric simplifications such as parallel and/or intersecting axes, which make them noncuspidal. However, if we seek to free ourselves from these rules to design a robot with an innovative architecture, we are very likely to come up with a cuspidal robot. In this paper, we have provided conditions and rules for a serial robot with three degrees of freedom to be cuspidal or not. The case of 6R robots still remains an open research topic. 

In the last part of the paper, we discussed the case of parallel robots. Few results exist for these robots, which are still rare in the industry. Their cuspidality behavior is very different from that of serial robots. Instead of impairing trajectory planning, it can even increase the size of the t-connected regions. The main difficulty of cuspidal parallel robots lies in identifying the assembly mode in which the robot finds itself at each instant, which requires the use of additional sensors, or the calculation and exploitation of uniqueness domains. These domains guarantee to always keep the same assembly mode, at the cost of a reduction in workspace.

For serial robots as for parallel robots, much work remains to be done for robots with six degrees of freedom. Visualization tools, new calculation methods that mix computer algebra with numerical methods are to be developed. Finally, we have seen in this paper the crucial role played by particular singular points, the cusp points. The impact of the cusp points on the precision and on the control remains to be studied. A cusp point is always a sufficient condition of cuspidality. On the other hand, we have seen an example of a cuspidal robot which does not have any cusp point. This robot is a very special parallel robot that was designed to exhibit this property. No other example of a serial or parallel robot could be identified that has the same feature. For 3R serial robots, it has been shown that the existence of a cusp point is also a necessary condition. It is very likely that it is the same for other families of serial and parallel robots but this question remains open.

\begin{acknowledgment}
This work was conducted in part with the financial support of the French National Research Agency (ECARP France-Austria Project ANR-19-CE48-0015).
\end{acknowledgment}

\bibliographystyle{ieeetr}
\bibliography{bibreport} 

\end{document}